\pdfoutput=1
\documentclass{article}

\usepackage[final,nonatbib]{nips_2017}

\usepackage[utf8]{inputenc} 
\usepackage[T1]{fontenc}    
\usepackage{color}
\usepackage{xcolor}

\colorlet{darkblue}{blue!60!black}
\colorlet{darkred}{red!60!black}
\usepackage[colorlinks=true,allcolors=darkblue]{hyperref}
\usepackage{booktabs}       
\usepackage{amsfonts}       
\usepackage{nicefrac}       
\usepackage{microtype}      
\usepackage{calc}
\usepackage{bm}
\usepackage{amssymb,amsmath}
\usepackage{mathtools}
\mathtoolsset{showonlyrefs}
\usepackage{wrapfig}
\usepackage{caption}

\DeclareCaptionFormat{cont}{#1 (continued)#2#3\par}

\usepackage{everypage}
\usepackage{environ}
\usepackage{pdflscape}
\newcounter{abspage}

\makeatletter
\newcommand{\newSFPage}[1]{\global\expandafter\let\csname SFPage@#1\endcsname\null}

\NewEnviron{SidewaysFigure}{\begin{figure}[p]
\protected@write\@auxout{\let\theabspage=\relax}
  {\string\newSFPage{\theabspage}}%
\ifdim\textwidth=\textheight
  \rotatebox{90}{\parbox[c][\textwidth][c]{\linewidth}{\BODY}}%
\else
  \rotatebox{90}{\parbox[c][\textwidth][c]{\textheight}{\BODY}}%
\fi
\end{figure}}
\NewEnviron{SidewaysContFigure}{\begin{figure}[p]\ContinuedFloat
\protected@write\@auxout{\let\theabspage=\relax}
  {\string\newSFPage{\theabspage}}%
\ifdim\textwidth=\textheight
  \rotatebox{90}{\parbox[c][\textwidth][c]{\linewidth}{\BODY}}%
\else
  \rotatebox{90}{\parbox[c][\textwidth][c]{\textheight}{\BODY}}%
\fi
\end{figure}}

\AddEverypageHook{
  \ifdim\textwidth=\textheight
    \stepcounter{abspage}
  \else
    \@ifundefined{SFPage@\theabspage}{}{\global\pdfpageattr{/Rotate 0}}%
    \stepcounter{abspage}%
    \@ifundefined{SFPage@\theabspage}{}{\global\pdfpageattr{/Rotate 90}}%
  \fi}
\makeatother

\newcommand\spacerule{\addlinespace[0.33em]}

\newcommand\wrt{w.r.t.\ }
\newcommand\tr{\mathrm{T}}
\newcommand\bs[1]{\bm{#1}}
\newcommand\maxop{{\max}_\Omega}
\newcommand\projop{\Pi_\Omega}
\newcommand\projsimplex{P_{\Delta^d}}
\newcommand\proxtv{P_{\text{TV}}}
\newcommand\proxoscar{P_{\text{OSC}}}
\newcommand\partialfrac[2]{\frac{\partial #1}{\partial #2}}
\newcommand\lang[1]{\makebox[\widthof{\scriptsize WMT16}][c]{#1}}

\DeclareMathOperator*{\dom}{dom}
\DeclareMathOperator*{\diag}{diag}
\DeclareMathOperator*{\argmax}{arg\,max}
\DeclareMathOperator*{\argmin}{arg\,min}
\DeclareMathOperator*{\sign}{sign}
\DeclareMathOperator*{\logsumexp}{log\,sum\,exp}

\newtheorem{proposition}{Proposition} 
\newtheorem{lemma}{Lemma} 

\title{A Regularized Framework for\\Sparse and Structured Neural Attention}

\author{
  Vlad Niculae\thanks{Work performed during an internship at NTT Commmunication
  Science Laboratories, Kyoto, Japan.} \\
  Cornell University \\
  Ithaca, NY \\
  \texttt{vlad@cs.cornell.edu} \\
  \And
  Mathieu Blondel\\
  NTT Communication Science Laboratories \\
  Kyoto, Japan\\
  \texttt{mathieu@mblondel.org} \\
}

\begin{document}

\maketitle

\begin{abstract}
Modern neural networks are often augmented with an attention mechanism, which
tells the network where to focus within the input.  We propose in this paper a
new framework for sparse and structured attention, building upon a smoothed max
operator. We show that the gradient of this operator defines a mapping from real
values to probabilities, suitable as an attention mechanism. Our framework
includes softmax and a slight generalization of the recently-proposed sparsemax
as special cases. However, we also show how our framework can incorporate modern
structured penalties, resulting in more interpretable attention mechanisms, that
focus on entire segments or groups of an input.  We derive efficient algorithms
to compute the forward and backward passes of our attention mechanisms, enabling
their use in a neural network trained with backpropagation.  To showcase their
potential as a drop-in replacement for existing ones, we evaluate our attention
mechanisms on three large-scale tasks: textual entailment, machine translation,
and sentence summarization.  Our attention mechanisms improve interpretability
without sacrificing performance; notably, on textual entailment and
summarization, we outperform the standard attention mechanisms based on softmax
and sparsemax.
\end{abstract}

\section{Introduction}

Modern neural network architectures are commonly augmented with an attention
mechanism, which tells the network where to look within the input in order to
make the next prediction.  Attention-augmented architectures have been
successfully applied to machine translation \cite{nmt_bahdanau,nmt_luong},
speech recognition \cite{attention_speech}, image caption generation
\cite{show_attent_tell}, textual entailment \cite{rockt,sparsemax}, and sentence
summarization \cite{rush_summary}, to name but a few examples. At the heart of
attention mechanisms is a mapping function that converts real values to
probabilities, encoding the relative importance of elements in the input.  For
the case of sequence-to-sequence prediction, at each time step of generating the
output sequence, attention probabilities are produced, conditioned on the
current state of a decoder network. They are then used to aggregate an input
representation (a variable-length list of vectors) into a single vector, which
is relevant for the current time step. That vector is finally fed into the
decoder network to produce the next element in the output sequence. This process
is repeated until the end-of-sequence symbol is generated. Importantly, such
architectures can be trained end-to-end using backpropagation.

Alongside empirical successes, neural attention---while not necessarily
correlated with human attention---is increasingly crucial in bringing more
{\bf interpretability} to neural networks by helping explain how individual
input elements contribute to the model's decisions.  However, the most
commonly used attention mechanism, {\em softmax}, yields dense attention weights:
all elements in the input always make at least a small
contribution to the decision. To overcome this limitation, {\em sparsemax} was
recently proposed \cite{sparsemax}, using the Euclidean projection onto the
simplex as a sparse alternative to softmax.  Compared to softmax, sparsemax
outputs more interpretable attention weights, as illustrated in \cite{sparsemax}
on the task of textual entailment.  The principle of parsimony, which states
that simple explanations should be preferred over complex ones, is not, however,
limited to sparsity: it remains open whether new attention
mechanisms can be designed to benefit from more structural prior knowledge.

\textbf{Our contributions.} The success of sparsemax motivates us to explore new
attention mechanisms that can both output sparse weights and take advantage of
structural properties of the input through the use of modern sparsity-inducing
penalties. To do so, we make the following contributions:

1) We propose a new \textbf{general framework} that builds upon a $\max$ operator,
regularized with a strongly convex function. We show that this operator is
differentiable, and that its gradient defines a mapping from real values to
probabilities, suitable as an attention mechanism. Our framework
\textbf{includes as special cases} both \textbf{softmax} and a slight generalization of
\textbf{sparsemax}. (\S \ref{sec:proposed_framework})

\begin{figure}[t]
    \includegraphics[width=\textwidth]{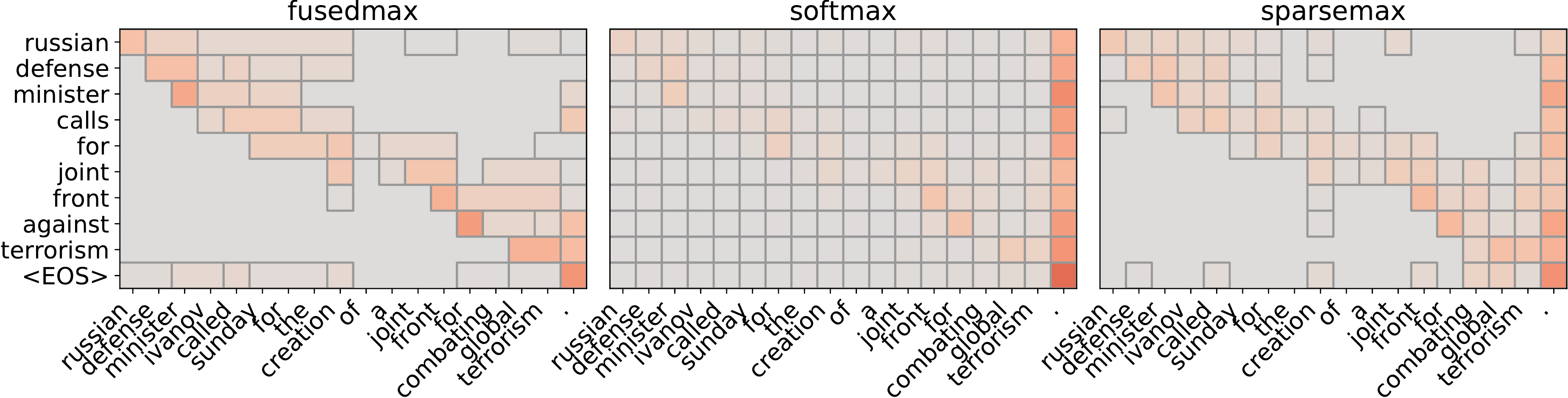} \caption{Attention
        weights produced by the proposed {\em fusedmax}, compared to {\em
        softmax} and {\em sparsemax}, on sentence summarization.  The input
        sentence to be summarized (taken from \cite{rush_summary}) is along the
        $x$-axis.  From top to bottom, each row shows where the attention is
        distributed when producing each word in the summary. All rows sum to 1,
        the grey background corresponds to exactly 0 (never achieved by
        softmax), and adjacent positions with exactly equal weight are not
        separated by borders. Fusedmax pays attention to contiguous segments of
        text with equal weight; such segments never occur with softmax and
        sparsemax.  In addition to enhancing interpretability, we show in
    \S\ref{sec:exp_summarization} that fusedmax outperforms both softmax and
sparsemax on this task in terms of ROUGE scores.\label{fig:intro_summ}}
\end{figure}

2) We show how to incorporate the fused lasso \cite{fused_lasso} in this
framework, to derive a new attention mechanism, named \emph{fusedmax}, which
encourages the network to pay attention to \textbf{contiguous segments of text}
when making a decision. This idea is illustrated in Figure~\ref{fig:intro_summ}
on sentence summarization. For cases when the contiguity assumption is too
strict, we show how to incorporate an OSCAR penalty \cite{oscar} to derive a
new attention mechanism, named \emph{oscarmax}, that encourages the network to
pay equal attention to \textbf{possibly non-contiguous groups of words}.
(\S\ref{sec:structured_attention})

3) In order to use attention mechanisms defined under our framework
in an autodiff toolkit,
two problems must be addressed: 
evaluating the attention itself and computing its
Jacobian. However, our attention mechanisms require solving a convex
optimization problem and do not generally enjoy a simple analytical expression,
unlike softmax. Computing the Jacobian of the solution of an optimization
problem is called argmin/argmax differentiation and is currently an area of
active research (cf. \cite{optnet} and references therein).
One of \textbf{our key algorithmic contributions is to show how to compute
this Jacobian} under our general framework, as well as
for fused lasso and OSCAR. (\S\ref{sec:structured_attention})

4) To showcase the potential of our new attention mechanisms as a
\textbf{drop-in replacement} for existing ones, we show empirically that our new
attention mechanisms enhance interpretability while achieving comparable or
better accuracy on three diverse and challenging tasks: \textbf{textual
entailment, machine translation, and sentence summarization}.
(\S\ref{sec:experiments}) 

\textcolor{darkred}{\textbf{Errata.} The NeurIPS 2017 version of this paper
contained an error in Proposition~\ref{prop:structured_forward}: the composition
of $\projsimplex$ and $\proxoscar$ is not equal to oscarmax in general, but only
an approximation of it.}

\textbf{Notation.}
We denote the set
$\{1,\dots,d\}$ by $[d]$.
We denote the $(d-1)$-dimensional probability simplex by $\Delta^d \coloneqq \{
\bs{x} \in \mathbb{R}^d \colon \|\bs{x}\|_1 = 1, \bs{x} \ge 0 \}$
and the Euclidean projection onto it by $\projsimplex(\bs{x}) \coloneqq
\argmin_{\bs{y} \in \Delta^d} \|\bs{y} - \bs{x}\|^2$.  
Given a function $f \colon \mathbb{R}^d \to \mathbb{R} \cup \{\infty\}$,
its convex conjugate is defined by
$
f^*(\bs{x}) \coloneqq 
\sup_{\bs{y} \in \dom f} ~ \bs{y}^\tr \bs{x} - f(\bs{y}).
$
Given a norm $\|\cdot\|$, its dual is defined by
$
\|\bs{x}\|_* \coloneqq \sup_{\|\bs{y}\| \le 1} ~ \bs{y}^\tr \bs{x}.
$
We denote the subdifferential of a function $f$ at $\bs{y}$ by $\partial
f(\bs{y})$. Elements of the subdifferential are called subgradients and
when $f$ is differentiable, $\partial f(\bs{y})$ contains a single element,
the gradient of $f$ at $\bs{y}$, denoted by $\nabla f(\bs{y})$.
We denote the Jacobian of a function $g \colon \mathbb{R}^d \to
\mathbb{R}^d$ at $\bs{y}$ by $J_g(\bs{y}) \in \mathbb{R}^{d \times d}$ and the
Hessian of a function $f \colon \mathbb{R}^d \to \mathbb{R}$ at $\bs{y}$ by
$H_f(\bs{y}) \in \mathbb{R}^{d \times d}$.

\section{Proposed regularized attention framework}
\label{sec:proposed_framework}

\subsection{The max operator and its subgradient mapping}

To motivate our proposal, we first show in this section that the subgradients
of the maximum operator define a mapping from $\mathbb{R}^d$ to $\Delta^d$, but
that this mapping is highly unsuitable as an attention mechanism.  The maximum
operator is a function from $\mathbb{R}^d$ to $\mathbb{R}$ and can be defined by
\begin{equation}
    \max(\bs{x}) \coloneqq \max_{i \in [d]} x_i = 
    \sup_{\bs{y} \in \Delta^d} \bs{y}^\tr \bs{x}.
\end{equation}
The equality on the r.h.s comes from the fact that the supremum of a linear form
over the simplex is always achieved at one of the vertices, i.e.,
one of the standard basis vectors $\{\bs{e}_i\}_{i=1}^d$.
Moreover, it is not hard to check that any solution $\bs{y}^\star$ of that
supremum is precisely a subgradient of $\max(\bs{x})$:
$\partial \max(\bs{x}) =
\{ \bs{e}_{i^\star} \colon i^\star \in \argmax_{i \in [d]} x_i \}$.
We can see these subgradients as a mapping $\Pi \colon \mathbb{R}^d \to
\Delta^d$ that \textbf{puts all the probability mass onto a single element}:
$\Pi(\bs{x}) = \bs{e}_i$ for any $\bs{e}_i \in \partial \max(\bs{x})$.
However, this behavior is undesirable, as the
resulting mapping is a discontinuous function (a Heaviside step function when
$\bs{x} = [t,0]$), which is not amenable to optimization by gradient
descent. 

\subsection{A regularized max operator and its gradient mapping}

\begin{figure}[t]
    \centering
    \includegraphics[scale=0.281]{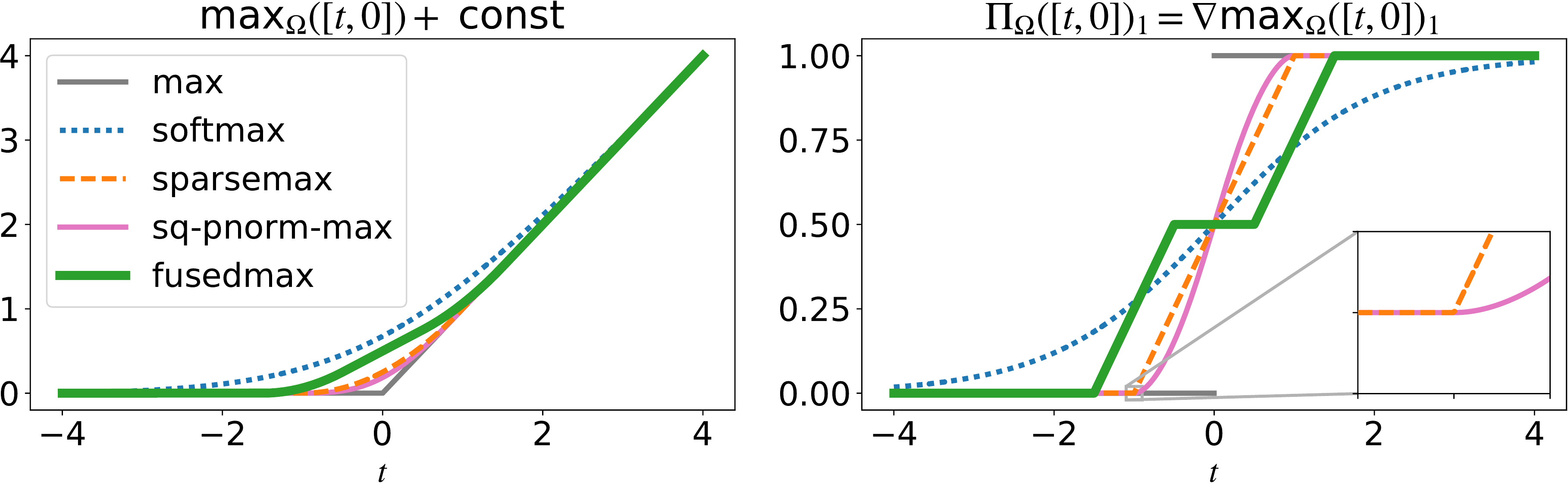}
    \caption{The proposed $\maxop(\bs{x})$ operator up to a constant (left) and
        the proposed $\projop(\bs{x})$ mapping (right), illustrated with $\bs{x}
        = [t, 0]$ and $\gamma=1$.  In this case, $\maxop(\bs{x})$ is a ReLu-like
        function and $\projop(\bs{x})$ is a sigmoid-like function.  Our
        framework recovers \emph{softmax} (negative entropy) and
        \emph{sparsemax} (squared $2$-norm) as special cases. We also introduce
        three new attention mechanisms: \emph{sq-pnorm-max} (squared $p$-norm,
        here illustrated with $p=1.5$), \emph{fusedmax} (squared $2$-norm +
        fused lasso), and \emph{oscarmax} (squared 2-norm + OSCAR; not pictured
        since it is equivalent to fusedmax in 2-d). Except for softmax, which
    never exactly reaches 0, all mappings shown on the right encourage sparse
outputs.  }
\label{fig:max_ops}
\end{figure}

These shortcomings encourage us to consider a regularization of the maximum
operator. Inspired by the seminal work of Nesterov \cite{nesterov_smooth}, we
apply a smoothing technique. The conjugate of $\max(\bs{x})$ is
\begin{equation}
    {\max}^*(\bs{y}) =
\begin{cases} 
0, & \mbox{if }  \bs{y} \in \Delta^d  \\ 
\infty,  & \mbox{o.w. } 
\end{cases}.
\end{equation}
For a proof, see for instance \cite[Appendix B]{smooth_and_strong}.
We now add regularization to the conjugate
\begin{equation}
\maxop^*(\bs{y}) \coloneqq
\begin{cases} 
\gamma \Omega(\bs{y}), & \mbox{if }  \bs{y} \in \Delta^d  \\ 
\infty,  & \mbox{o.w. } 
\end{cases},
\end{equation}
where we assume that $\Omega \colon \mathbb{R}^d \to \mathbb{R}$ is
\textbf{$\beta$-strongly convex} \wrt some norm $\|\cdot\|$ and $\gamma > 0$
controls the regularization strength. To define a
smoothed $\max$ operator, we take the conjugate once again
\begin{equation}
\maxop(\bs{x}) =
\maxop^{**}(\bs{x}) =
\sup_{\bs{y} \in \mathbb{R}^d} \bs{y}^\tr \bs{x} - \maxop^*(\bs{y})
= \sup_{\bs{y} \in \Delta^d} \bs{y}^\tr \bs{x} - \gamma \Omega(\bs{y}).
\label{eq:regularized_max}
\end{equation}
Our main proposal is a mapping $\projop \colon \mathbb{R}^d \to
\Delta^d$, defined as the \textit{argument} that achieves this supremum.
\begin{equation}
\boxed{
\projop(\bs{x}) \coloneqq 
\argmax_{\bs{y} \in \Delta^d} \bs{y}^\tr \bs{x} - \gamma \Omega(\bs{y}) =
\nabla \maxop(\bs{x})
}
\end{equation}
The r.h.s. holds by combining that i) $\maxop(\bs{x}) = (\bs{y}^\star)^\tr
\bs{x} - \maxop^*(\bs{y}^\star) \Leftrightarrow \bs{y}^\star \in \partial
\maxop(\bs{x})$ and ii) $\partial \maxop(\bs{x}) = \{ \nabla \maxop(\bs{x}) \}$,
since \eqref{eq:regularized_max} has a unique solution. Therefore, $\projop$
is a \textbf{gradient mapping}. 
We illustrate $\maxop$ and $\projop$ for various choices of $\Omega$ in Figure
\ref{fig:max_ops} (2-d) and in Appendix \ref{appendix:3d_plots} (3-d).

\textbf{Importance of strong convexity.}
Our $\beta$-strong convexity assumption on $\Omega$ plays a crucial role and
should not be underestimated. Recall that a function $f
\colon \mathbb{R}^d \to \mathbb{R}$ is $\beta$-strongly convex \wrt a norm
$\|\cdot\|$ if and only if its conjugate $f^*$ is $\frac{1}{\beta}$-smooth
\wrt the dual norm $\|\cdot\|_*$ \cite[Corollary 3.5.11]{zalinescu}
\cite[Theorem 3]{kakade_jmlr}.  This is sufficient to ensure that $\maxop$ is
\textbf{$\frac{1}{\gamma \beta}$-smooth}, or, in other words, that it is
differentiable everywhere and its gradient, $\projop$, is $\frac{1}{\gamma
\beta}$-Lipschitz continuous \wrt $\|\cdot\|_*$.

\textbf{Training by backpropagation.} In order to use $\projop$ in a neural
network trained by backpropagation, two problems must be addressed for any
regularizer $\Omega$. The first is the \textbf{forward} computation: how to
evaluate $\projop(\bs{x})$, i.e., how to solve the optimization problem in
\eqref{eq:regularized_max}.  The second is the \textbf{backward} computation:
how to evaluate the Jacobian of $\projop(\bs{x})$, or, equivalently, the Hessian
of $\maxop(\bs{x})$.  One of our key contributions, presented in
\S\ref{sec:structured_attention}, is to show how to solve these two problems for
general differentiable $\Omega$, as well as for two structured regularizers:
fused lasso and OSCAR.

\subsection{Recovering softmax and sparsemax as special cases}

Before deriving new attention mechanisms using our framework, we now show how we
can recover softmax and sparsemax, using a specific regularizer $\Omega$.

\textbf{Softmax.} We choose $\Omega(\bs{y}) = \sum_{i=1}^d y_i \log y_i$, 
the negative entropy.  The conjugate of the negative entropy restricted to the
simplex is the $\logsumexp$ \cite[Example 3.25]{boyd_book}.  Moreover, if
$f(\bs{x})=\gamma g(\bs{x})$ for $\gamma > 0$, then $f^*(\bs{y}) =
\gamma g^*(\bs{y} / \gamma)$.  We therefore get a closed-form expression:
$
\maxop(\bs{x}) 
= \gamma ~ \logsumexp(\bs{x} / \gamma)
\coloneqq \gamma \log \sum_{i=1}^d e^{x_i / \gamma}.
$
Since the negative entropy is $1$-strongly convex \wrt $\|\cdot\|_1$ over
$\Delta^d$, we get that $\maxop$ is $\frac{1}{\gamma}$-smooth \wrt
$\|\cdot\|_\infty$. We obtain the classical softmax, with temperature parameter
$\gamma$, by taking the gradient of $\maxop(\bs{x})$,
\begin{equation}
\projop(\bs{x})
= \frac{e^{\bs{x}/\gamma}}{\sum_{i=1}^d e^{x_i / \gamma}},
\tag*{(softmax)}
\end{equation}
where $e^{\bs{x}/\gamma}$ is evaluated element-wise.  Note that some authors
also call $\maxop$ a ``soft max.'' Although $\projop$ is really a soft {\em
arg\,max}, we opt to follow the more popular terminology.  When
$\bs{x} = [t, 0]$, it can be checked that $\maxop(\bs{x})$ reduces to the
softplus \cite{softplus} and $\projop(\bs{x})_1$ to a sigmoid.

\textbf{Sparsemax.} We choose $\Omega(\bs{y}) = \frac{1}{2} \|\bs{y}\|^2_2$,
also known as Moreau-Yosida regularization in proximal operator theory
\cite{nesterov_smooth,proximal_algorithms}. 
Since $\frac{1}{2} \|\bs{y}\|^2_2$ is $1$-strongly convex \wrt $\|\cdot\|_2$,
we get that $\maxop$ is $\frac{1}{\gamma}$-smooth \wrt $\|\cdot\|_2$.  In
addition, it is easy to verify that
\begin{equation}
\projop(\bs{x}) = 
\projsimplex(\bs{x} / \gamma) =
\argmin_{\bs{y} \in \Delta^d} \|\bs{y} - \bs{x} / \gamma\|^2.
\tag*{(sparsemax)}
\end{equation}
This mapping was introduced {\em as is} in \cite{sparsemax} with $\gamma=1$ and
was named sparsemax, due to the fact that it is a sparse alternative to softmax.
Our derivation thus gives us a slight generalization, where $\gamma$ controls
the sparsity (the smaller, the sparser) and could be tuned; in our experiments,
however, we follow the literature and set $\gamma=1$.
The Euclidean projection onto the simplex, $\projsimplex$, can be computed
exactly \cite{michelot,duchi} (we discuss the complexity in
Appendix~\ref{appendix:complex}). Following \cite{sparsemax}, the Jacobian of
$\projop$ is
\begin{equation}
J_{\projop}(\bs{x}) =
\frac{1}{\gamma} J_{\projsimplex}(\bs{x} / \gamma) =
\frac{1}{\gamma} \left(
\diag(\bs{s}) - \bs{s} \bs{s}^\tr / \|\bs{s}\|_1 \right),
\label{eq:Jacobian_sparsemax}
\end{equation}
where $\bs{s} \in \{0,1\}^d$ indicates the nonzero elements of 
$\projop(\bs{x})$. Since $\projop$ is
Lipschitz continuous, Rademacher's theorem implies that $\projop$ is
differentiable almost everywhere. For points where $\projop$ is not
differentiable (where $\maxop$ is not twice differentiable), we can take
an arbitrary matrix in the set of Clarke's generalized Jacobians
\cite{clarke_book}, the convex hull of Jacobians of the form
$\lim\limits_{\bs{x}_t \to \bs{x}} J_{\projop}(\bs{x}_t)$ \cite{sparsemax}.

\section{Deriving new sparse and structured attention mechanisms}
\label{sec:structured_attention}

\subsection{Differentiable regularizer $\Omega$}

Before tackling more structured regularizers, we address in this section the
case of general differentiable regularizer $\Omega$. Because
$\projop(\bs{x})$ involves maximizing \eqref{eq:regularized_max}, a concave
function over the simplex, it can be computed globally using any off-the-shelf
projected gradient solver. Therefore, the main challenge is how to compute the
Jacobian of $\projop$. This is what we address in the next proposition.

\begin{proposition}{Jacobian of $\projop$ for any differentiable $\Omega$
    (backward computation)}

Assume that $\Omega$ is differentiable over $\Delta^d$ and that $\projop(\bs{x})
= \argmax_{\bs{y} \in \Delta^d} \bs{y}^\tr \bs{x} - \gamma \Omega(\bs{y}) =
\bs{y}^\star$ has been computed. Then the Jacobian of $\projop$ at $\bs{x}$,
denoted $J_{\projop}$, can be obtained by solving the system
\begin{equation}
(I + A(B - I)) ~ J_{\projop} = A,
\end{equation}
where we defined the shorthands 
$
A \coloneqq J_{\projsimplex}(\bs{y}^\star -
\gamma \nabla \Omega(\bs{y}^\star) + \bs{x}) 
\quad \text{and} \quad
B \coloneqq \gamma H_\Omega(\bs{y}^\star).
$
\label{prop:Jacobian_diff_regul}
\end{proposition}
The proof is given in Appendix \ref{appendix:proof_Jacobian_diff_regul}.
Unlike recent work tackling argmin differentiation through matrix differential
calculus on the Karush--Kuhn--Tucker (KKT) conditions \cite{optnet}, our
proof technique relies on differentiating the
fixed point iteration $\bs{y}^* = \projsimplex(\bs{y}^\star - \nabla
f(\bs{y}^\star))$.
To compute $J_{\projop} \bs{v}$ for
an arbitrary $\bs{v} \in \mathbb{R}^d$, as required by backpropagation,
we may directly solve
$(I + A(B - I)) ~ (J_{\projop} \bs{v}) = A \bs{v}$. We show in
Appendix~\ref{appendix:complex} how this system can be solved
efficiently thanks to the structure of $A$.

\textbf{Squared $p$-norms.} As a useful example of a differentiable function over
the simplex, we consider squared $p$-norms: $\Omega(\bs{y}) =
\frac{1}{2} \|\bs{y}\|_p^2 = \left( \sum_{i=1}^d y_i^p \right)^{2/p}$, where
$\bs{y} \in \Delta^d$ and $p
\in (1,2]$. For this choice of $p$, it is known that the squared p-norm is
strongly convex \wrt $\|\cdot\|_p$ \cite{ball_1994}. This implies that
$\maxop$ is $\frac{1}{\gamma(p-1)}$ smooth \wrt $\|.\|_q$, where $\frac{1}{p} +
\frac{1}{q} = 1$. We call the induced mapping function
\emph{sq-pnorm-max}: 
\begin{equation}
\projop(\bs{x}) = \argmin_{\bs{y} \in \Delta^d} 
\frac{\gamma}{2} \|\bs{y}\|_p^2 - \bs{y}^\tr \bs{x}.
\tag*{(sq-pnorm-max)}
\end{equation}%
The gradient and Hessian needed for Proposition
\ref{prop:Jacobian_diff_regul} can be computed by
$\nabla \Omega(\bs{y}) =
\frac{\bs{y}^{p-1}}{\|\bs{y}\|^{p-2}_p}$ and
\begin{equation}
H_\Omega(\bs{y}) = \diag(\bs{d}) + \bs{u} \bs{u}^\tr,
\quad \text{where} \quad
\bs{d} =  \frac{(p-1)}{\|\bs{y}\|_p^{p-2}} ~ \bs{y}^{p-2} 
\quad \text{and} \quad
\bs{u} = \sqrt{\frac{(2-p)}{\|\bs{y}\|_p^{2p-2}}} ~ \bs{y}^{p-1},
\end{equation}
with the exponentiation performed element-wise.  sq-pnorm-max recovers
sparsemax with $p=2$ and, like sparsemax, encourages sparse outputs.  However,
as can be seen in the zoomed box in Figure \ref{fig:max_ops} (right), the
transition between $\bs{y}^\star = [0,1]$ and $\bs{y}^\star = [1,0]$ can be
smoother when $1 < p < 2$.
Throughout our experiments, we use $p=1.5$.

\subsection{Structured regularizers: fused lasso and OSCAR}
\label{sec:structured_attention_fused_oscar}

\textbf{Fusedmax.} For cases when the input is sequential and the order is
meaningful, as is the case for many natural languages, we propose {\em
fusedmax}, an attention mechanism based on {\em fused lasso} \cite{fused_lasso},
also known as 1-d total variation (TV). Fusedmax encourages paying attention to
\textbf{contiguous segments}, with equal weights within each one. It is
expressed under our framework by choosing 
$\Omega(\bs{y}) =
\frac{1}{2} \|\bs{y}\|_2^2 + \lambda
\sum_{i=1}^{d-1} |y_{i+1} - y_i|$, i.e., the sum of a strongly convex term and
of a 1-d TV penalty. It is easy to verify that this choice yields the mapping
\begin{equation}
\projop(\bs{x}) = \argmin_{\textcolor{darkred}{\bs{y} \in \Delta^d}} 
\frac{1}{2} \|\bs{y} - \bs{x} / \gamma\|^2 + 
\lambda \sum_{i=1}^{d-1} |y_{i+1} - y_i|.
\tag*{(fusedmax)}
\end{equation}

\textbf{Oscarmax.} For situations where the contiguity assumption may
be too strict, we propose {\em oscarmax},
based on the OSCAR penalty \cite{oscar}, to encourage attention weights to
\textbf{merge into clusters with the same value}, regardless of position in the
sequence.  This is accomplished by replacing the 1-d TV penalty in fusedmax
with an $\infty$-norm penalty on each pair of attention weights, i.e., 
$\Omega(\bs{y}) = 
\frac{1}{2} \|\bs{y}\|_2^2 + \lambda \sum_{i < j}
\max(|y_i|, |y_j|)$. This results in the mapping
\begin{equation}
\projop(\bs{x}) = \argmin_{\textcolor{darkred}{\bs{y} \in \Delta^d}} 
\frac{1}{2} \|\bs{y} - \bs{x} / \gamma\|^2 + 
\lambda \sum_{i < j} \max(|y_i|, |y_j|).
\tag*{(oscarmax)}
\end{equation}
\textbf{Forward computation.} Due to the $\textcolor{darkred}{\bs{y} \in \Delta^d}$
constraint, computing fusedmax/oscarmax does not seem trivial on first sight.
Fortunately, the next proposition shows how to compute fusedmax exactly without
any iterative method.
\begin{proposition}{Computing fusedmax (forward computation)}
\begin{equation}
    \text{fusedmax: } \projop(\bs{x}) =
\projsimplex\left(\proxtv\left(\bs{x} / \gamma\right)\right),  \quad
\proxtv(\bs{x}) \coloneqq \displaystyle{\argmin_{%
\textcolor{darkred}{\bs{y} \in \mathbb{R}^d}}}%
~ \frac{1}{2} \|\bs{y} - \bs{x}\|^2 + %
\lambda \sum_{i=1}^{d-1} |y_{i+1} - y_i|. 
\end{equation}
\label{prop:structured_forward}
\end{proposition}%
Here, $\proxtv$ indicates the proximal operator of 1-d TV and can be computed
\textbf{exactly} by \cite{tv1d_prox}.  To remind the reader, $\projsimplex$
denotes the Euclidean projection onto the simplex and can be computed exactly
using \cite{michelot,duchi}. Proposition \ref{prop:structured_forward} shows
that we can compute fusedmax using the composition of two
functions, for which exact non-iterative algorithms exist.  This is a surprising
result, since the proximal operator of the sum of two functions is not, in
general, the composition of the proximal operators of each function. The proof
follows by showing that the indicator function of $\Delta^d$ satisfies the
conditions of \cite[Corollary 4]{decomposing_prox}.
For oscarmax, we use a similar decomposition, although it is no longer exact,
since the indicator function of $\Delta^d$ does not satisfy the conditions of
\cite[Corollary 5]{decomposing_prox}.
\begin{equation}
\text{oscarmax: }\projop(\bs{x}) \approx
\projsimplex\left(\proxoscar\left(\bs{x}/ \gamma\right)\right), \quad
\proxoscar(\bs{x}) \coloneqq \displaystyle \argmin_{%
\textcolor{darkred}{\bs{y}\in \mathbb{R}^d}}%
\hspace{0.5mm} \frac{1}{2} \|\bs{y} - \bs{x}\|^2 + 
\lambda \sum_{i < j} \max(|y_i|, |y_j|).  
\end{equation}
Here, $\proxoscar$ indicates the proximal operator of OSCAR, and can be computed
exactly by \cite{oscar_prox}.  

\textbf{Groups induced by $\proxtv$ and $\proxoscar$.} Let $\bs{z}^\star$ be the
optimal solution of $\proxtv(\bs{x})$ or $\proxoscar(\bs{x})$. For $\proxtv$, we
denote the group of {\bf adjacent elements with the same value} as $z^\star_i$
by $G_i^\star$, $\forall i \in [d]$. Formally, $G_i^\star = [a,b]\cap\mathbb{N}$
with $a \leq i \leq b$ where $a$ and $b$ are the minimal and maximal indices
such that $z^\star_i = z^\star_j$ for all $j \in G^\star_i$.  For $\proxoscar$,
we define $G_i^\star$ as the indices of {\bf elements with the same absolute
value} as $z^\star_i$, more formally $G_i^\star = \{j \in [d] \colon |z_i^\star|
= |z_j^\star| \}$. Because $\projsimplex(\bs{z}^\star) = \max(\bs{z}^\star -
\theta, 0)$ for some $\theta \in \mathbb{R}$, fusedmax/oscarmax either shift a
group's common value or set all its elements to zero.

$\lambda$ controls the trade-off between no fusion (sparsemax) and all elements
fused into a single trivial group.  While tuning $\lambda$ may improve
performance, we observe that $\lambda=0.1$ (fusedmax) and $\lambda=0.01$
(oscarmax) are sensible defaults that work well across all tasks and report
all our results using them.

\textbf{Backward computation.} We already know that the Jacobian of
$\projsimplex$ is the same as that of sparsemax with $\gamma=1$.  Then, if we
know how to compute the Jacobians of $\proxtv$ and $\proxoscar$, we can obtain
the Jacobians of fusedmax and of the approximate oscarmax by mere
application of the chain rule.  However, although $\proxtv$ and $\proxoscar$ can
be computed exactly, they lack analytical expressions.  We next show that we can
nonetheless compute their Jacobians efficiently, without needing to solve a
system. 

\begin{minipage}[c]{\textwidth}
\begin{proposition}{Jacobians of $\proxtv$ and $\proxoscar$ (backward
    computation)}

\vspace{0.3cm}
Assume $\bs{z}^\star = \proxtv(\bs{x}) \text{ or } \proxoscar(\bs{x})$ has
been computed. Define the groups derived from $\bs{z}^\star$ as above.
\begin{equation}
\text{Then, } 
[J_{\proxtv}(\bs{x})]_{i,j} = 
\begin{cases}
\frac{1}{|G_i^\star|} & \mbox{if } j \in G_i^\star,\\
0 & \mbox{o.w. }
\end{cases}
\hspace{.3em}\text{and}\hspace{.3em}
[J_{\proxoscar}(\bs{x})]_{i,j} =
\begin{cases}
\frac{\sign(z^\star_i z^\star_j)}{|G_i^\star|} & \mbox{if } j \in G_i^\star \mbox{ and } z^\star_i \neq 0, \\
0 & \mbox{o.w.}
\end{cases}.
\end{equation}
\vspace{-0.3cm}
\label{prop:structured_backward}
\end{proposition}
\end{minipage}

The proof is given in Appendix \ref{appendix:proof_structured_backward}.
Clearly, the structure of these Jacobians permits efficient Jacobian-vector
products; we discuss the computational complexity and implementation details in
Appendix~\ref{appendix:complex}.  Note that $\proxtv$ and $\proxoscar$ are
differentiable everywhere except at points where groups change. For these
points, the same remark as for sparsemax applies, and we can use Clarke's
Jacobian.

\section{Experimental results}
\label{sec:experiments}

We showcase the performance of our attention mechanisms on three challenging
natural language tasks: textual entailment, machine translation,
and sentence summarization. We rely on available,
well-established neural architectures, so as to demonstrate simple drop-in
replacement of softmax with structured sparse attention; quite 
likely, newer task-specific models could lead to further improvement.

\subsection{Textual entailment (a.k.a. natural language inference) experiments}

Textual entailment is the task of deciding, given a text T and an hypothesis H,
whether a human reading T is likely to infer that H is true \cite{dagan_nle}.
We use the Stanford Natural Language Inference (SNLI) dataset \cite{snli}, a
collection of 570,000 English sentence pairs. Each pair consists of a sentence
and an hypothesis, manually labeled with one of the labels \textsc{entailment},
\textsc{contradiction}, or \textsc{neutral}.  

\begin{wraptable}{r}{4cm}
\vspace{-10pt}
\caption{Textual entailment test accuracy on SNLI \cite{snli}.}
    \small
    \centering
    \begin{tabular}{r c c}
        \toprule
        attention & accuracy \\
        \midrule
        softmax & 81.66 \\
        sparsemax & 82.39 \\
        \spacerule
        fusedmax & \textbf{82.41} \\
        oscarmax & 81.76 \\
        \bottomrule
    \end{tabular}
\label{table:snli}
\vspace{-10pt}
\end{wraptable}

We use a variant of the  neural attention--based classifier proposed for this
dataset by \cite{rockt} and follow the same methodology as \cite{sparsemax} in
terms of implementation, hyperparameters, and grid search. We employ the
CPU implementation provided in \cite{sparsemax} and simply replace sparsemax
with fusedmax/oscarmax; we observe that training time per epoch is essentially
the same for each of the four attention mechanisms (timings and more
experimental details in Appendix \ref{appendix:textual_entailment}).

\begin{figure}[t]
    \includegraphics[width=\textwidth]{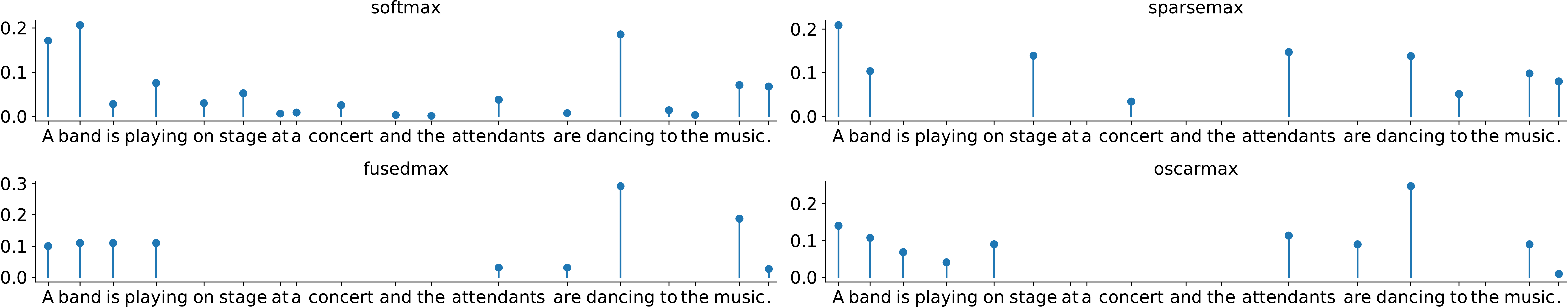}
    \caption{Attention weights when considering the contradicted hypothesis ``No
    one is dancing.''}
    \label{fig:snli}
\end{figure}

Table \ref{table:snli} shows that, for this task, fusedmax reaches the highest
accuracy, and oscarmax slightly outperforms softmax.  Furthermore, fusedmax
results in the most interpretable feature groupings: Figure \ref{fig:snli} shows
the weights of the neural network's attention to the text, when considering the
hypothesis ``No one is dancing.'' In this case, all four models correctly
predicted that the text ``A band is playing on stage at a concert and the
attendants are dancing to the music,'' denoted along the $x$-axis,
\textbf{contradicts} the hypothesis, although the attention weights differ.
Notably, fusedmax identifies the meaningful segment ``band is playing''.

\subsection{Machine translation experiments}
\label{sec:exp_mt}

Sequence-to-sequence neural machine translation (NMT) has recently become a
strong contender in machine translation \cite{nmt_bahdanau,nmt_luong}. In NMT, 
attention weights can be seen as an {\em alignment} between source and
translated words.  To demonstrate the potential of our new attention mechanisms
for NMT, we ran experiments on 10 language pairs.  We build on OpenNMT-py
\cite{opennmt}, based on PyTorch \cite{pytorch}, with all default
hyperparameters (detailed in Appendix~\ref{appendix:machine_translation}),
simply replacing softmax with the proposed $\Pi_\Omega$.

OpenNMT-py with softmax attention is optimized for the GPU. Since sparsemax,
fusedmax, and oscarmax rely on sorting operations, we implement their
computations on the CPU for simplicity, keeping the rest of the pipeline on the
GPU. However, we observe that, even with this context switching, the number of
tokens processed per second was within \nicefrac{3}{4} of the softmax pipeline.
For sq-pnorm-max, we observe that the projected gradient solver used in
the forward pass, unlike the linear system solver used in the backward pass,
could become a computational bottleneck. To mitigate this effect, we set the tolerance of the
solver's stopping criterion to $10^{-2}$.

\begin{figure}[h]
    \centering
    \includegraphics[width=0.9\textwidth]{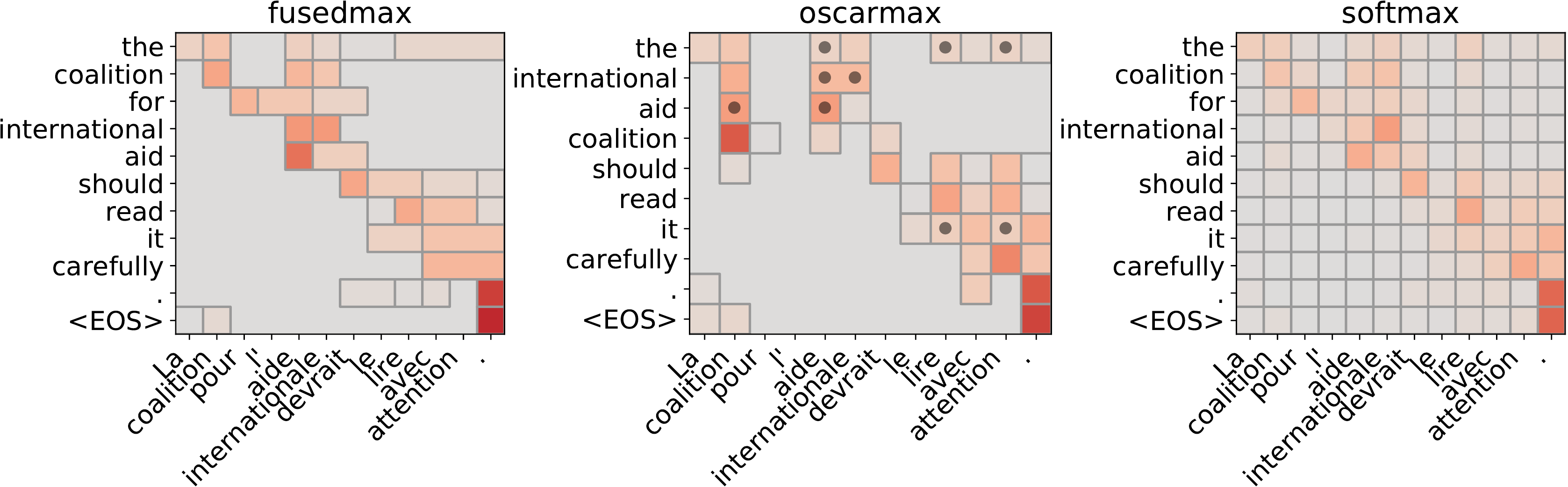}
    \caption{Attention weights for French to English translation, using the
        conventions of Figure~\ref{fig:intro_summ}.  Within a row, weights
        grouped by oscarmax under the same cluster are denoted
        by ``\textbullet''. Here, oscarmax finds a slightly more natural
        English translation. More visulizations are given in
        Appendix~\ref{appendix:machine_translation}.
}
    \label{fig:nmt}
\end{figure}

Quantitatively, we find that all compared attention mechanisms are always within
1 BLEU score point of the best mechanism (for detailed results, cf.  Appendix
\ref{appendix:machine_translation}). This suggests that structured sparsity does
not restrict accuracy. However, as illustrated in Figure \ref{fig:nmt}, fusedmax
and oscarmax often lead to more interpretable attention alignments, as well as
to qualitatively different translations.

\subsection{Sentence summarization experiments}
\label{sec:exp_summarization}

Attention mechanisms were recently explored for sentence summarization in
\cite{rush_summary}. To generate sentence-summary pairs at low cost, the authors
proposed to use the title of a news article as a noisy summary of the article's
leading sentence. They collected 4 million such pairs from the Gigaword dataset
and showed that this seemingly simplistic approach leads to models that
generalize surprisingly well. We follow their experimental setup and are able
to reproduce comparable results to theirs with OpenNMT when using softmax
attention. The models we use are the same as in \S\ref{sec:exp_mt}.

Our evaluation follows \cite{rush_summary}: we use the
standard DUC 2004 dataset (500 news articles each paired with 4 different
human-generated summaries) and a randomly held-out subset of Gigaword, released
by \cite{rush_summary}. We report results on {\footnotesize ROUGE-1},
{\footnotesize ROUGE-2}, and {\footnotesize ROUGE-L}.  Our results, in
Table~\ref{table:summarization_results}, indicate that fusedmax is the best
under nearly all metrics, always outperforming softmax.
In addition to Figure~\ref{fig:intro_summ}, we
exemplify our enhanced interpretability and provide more detailed results in
Appendix~\ref{appendix:summarization}.

\begin{table}[t]
    \caption{Sentence summarization results, following the same experimental
        setting as in \cite{rush_summary}.}
    \small
    \centering
    \begin{tabular}{r c c c c c c}
        \toprule
                   & \multicolumn{3}{c}{DUC 2004} & \multicolumn{3}{c}{Gigaword}\\
        \cmidrule(lr){2-4} \cmidrule(lr){5-7} 
        attention  & ROUGE-1 & ROUGE-2 & ROUGE-L & ROUGE-1 & ROUGE-2 & ROUGE-L \\
        \midrule
        softmax    &   27.16& 9.48 &  24.47& 35.13& 17.15 & 32.92\\ 
        sparsemax  &   27.69& 9.55 &  24.96& 36.04& \textbf{17.78} & 33.64\\ 
        \spacerule
        fusedmax   &  \textbf{28.42}& \textbf{9.96}&  \textbf{25.55}&
        \textbf{36.09}& 17.62 & \textbf{33.69}\\
        oscarmax   &   27.84& 9.46 &  25.14& 35.36& 17.23 & 33.03\\
      sq-pnorm-max &   27.94& 9.28 &  25.08& 35.94& 17.75 & 33.66\\
        \bottomrule
    \end{tabular}
    \label{table:summarization_results}
\end{table}

\section{Related work}

\textbf{Smoothed max operators.} Replacing the $\max$ operator
by a differentiable approximation based on the $\logsumexp$ has been exploited
in numerous works. Regularizing the $\max$ operator with a squared $2$-norm is
less frequent, but has been used to obtain a smoothed multiclass hinge loss
\cite{prox_sdca} or smoothed linear programming relaxations for maximum
a-posteriori inference \cite{smooth_and_strong}. Our work differs from these in
mainly two aspects. First, we are less interested in the $\max$ operator itself
than in its gradient, which we use as a mapping from $\mathbb{R}^d$ to
$\Delta^d$. Second, since we use this mapping in neural networks trained with
backpropagation, we study and compute the mapping's Jacobian (the Hessian of a
regularized $\max$ operator), in contrast with previous works.

\textbf{Interpretability, structure and sparsity in neural networks.} Providing
interpretations alongside predictions is important for accountability, error
analysis and exploratory analysis, among other reasons.  Toward this goal,
several recent works have been relying on visualizing hidden layer activations
\cite{ozan_thesis,jiwei} and the potential for interpretability provided by
attention mechanisms has been noted in multiple works
\cite{nmt_bahdanau,rockt,rush_summary}.  Our work aims to fulfill this potential by
providing a unified framework upon which new interpretable attention
mechanisms can be designed, using well-studied tools from the field of structured
sparse regularization.

Selecting contiguous text segments for model interpretations
is explored in \cite{taolei}, where an {\em explanation generator} network is
proposed for justifying predictions using a fused lasso penalty. However, this
network is not an attention mechanism and has its own parameters to be learned.
Furthemore, \cite{taolei} sidesteps the need to backpropagate through the fused
lasso, unlike our work, by using a stochastic training approach.  In constrast,
our attention mechanisms are deterministic and {\bf drop-in replacements} for
existing ones. 
As a consequence, our mechanisms can be coupled with recent research
that builds on top of softmax attention, for example in order to incorporate
soft prior knowledge about NMT alignment into attention through penalties on the
attention weights \cite{cohn}.

A different approach to incorporating structure into attention uses the
posterior marginal probabilities from a conditional random field as attention weights
\cite{structured_attn}.  While this approach takes into account structural
correlations, the marginal probabilities are generally dense and different from
each other. Our proposed mechanisms produce sparse and clustered attention
weights, a visible benefit in interpretability.
The idea of deriving constrained alternatives to softmax has been independently
explored for differentiable easy-first decoding \cite{andre-easy}.
Finally, sparsity-inducing penalties have been used to
obtain convex relaxations of neural networks \cite{convex_nn_bengio} or to compress
models \cite{sparse_cnn,structured_dnn,group_sparse_dnn}. These works differ
from ours, in that sparsity is enforced on the network \textbf{parameters}, while our
approach can produce sparse and structured \textbf{outputs} from neural attention layers.

\section{Conclusion and future directions}

We proposed in this paper a unified regularized framework upon which new
attention mechanisms can be designed. To enable such mechanisms to be used in a
neural network trained by backpropagation, we demonstrated how to carry out
forward and backward computations for general differentiable regularizers. We
further developed two new structured attention mechanisms, \emph{fusedmax} and
\emph{oscarmax}, and demonstrated that they enhance interpretability while
achieving comparable or better accuracy on three diverse and challenging tasks:
textual entailment, machine translation, and summarization. 

The usefulness of a differentiable mapping from real values to the simplex or to
$[0,1]$  with sparse or structured outputs goes beyond attention
mechanisms.  We expect that our framework will be useful to sample from
categorical distributions using the Gumbel trick
\cite{gumbel_softmax,concrete_distribution}, as well as for conditional computation
\cite{conditional_bengio} or differentiable neural
computers \cite{NTM,diff_neural_computer}.  We plan to explore these
in future work.

\section*{Acknowledgements}

We are grateful to
Andr\'{e} Martins,
Takuma Otsuka,
Fabian Pedregosa,
Antoine Rolet,
Jun Suzuki,
and
Justine Zhang
for helpful discussions.
We thank the anonymous reviewers for the valuable feedback.


\clearpage
\appendix

\begin{center}
    {\Huge \bf Supplementary material}
\end{center}

\section{Proofs}

\subsection{Proof of Proposition \ref{prop:Jacobian_diff_regul}}
\label{appendix:proof_Jacobian_diff_regul}

Recall that
\begin{equation}
\projop(\bs{x}) = 
\argmin_{\bs{y} \in \Delta^d} f(\bs{y}),
\quad \text{where} \quad 
f(\bs{y}) \coloneqq
\gamma \Omega(\bs{y}) - \bs{y}^\tr \bs{x}.
\end{equation}

At an optimal solution, we have the fixed point iteration 
\cite[\S4.2]{proximal_algorithms}
\begin{equation}
\bs{y}^* = \projsimplex(\bs{y}^\star - \nabla f(\bs{y}^\star)).
\label{eq:fixed_point}
\end{equation}
Seeing $\bs{y}^\star$ as a function of $\bs{x}$, and $\projsimplex$ and $\nabla
f$ as functions of their inputs, we can apply the chain rule to
\eqref{eq:fixed_point} to obtain
\begin{equation}
    J_{\projop}(\bs{x}) = J_{\projsimplex} \left(\bs{y}^\star - \nabla
    f(\bs{y}^\star)\right) \left(J_{\projop}(\bs{x}) - J_{\nabla
f \circ \bs{y}^\star}(\bs{x})\right).
\label{eq:fixed_point_chain_rule}
\end{equation}
Applying the chain rule once again to $\nabla f(\bs{y}^\star) = \gamma \nabla
\Omega(\bs{y}^\star) - \bs{x}$, we obtain
\begin{equation}
\begin{aligned}
J_{\nabla f \circ \bs{y}^\star}(\bs{x}) &=
\gamma J_{\nabla \Omega}(\bs{y}^\star) J_{\projop}(\bs{x}) - I \\
&= \gamma H_\Omega(\bs{y}^\star) J_{\projop}(\bs{x}) - I. 
\end{aligned}
\end{equation}
Plugging 
this into \eqref{eq:fixed_point_chain_rule} and re-arranging, we obtain
\begin{equation}
(I + A(B - I)) ~ J_{\projop} = A,
\end{equation}
where we defined the shorthands 
\begin{equation}
J_{\projop} \coloneqq J_{\projop}(\bs{x}), \quad
A \coloneqq J_{\projsimplex}(\bs{y}^\star -
\gamma \nabla \Omega(\bs{y}^\star) + \bs{x}) 
\quad \text{and} \quad
B \coloneqq \gamma H_\Omega(\bs{y}^\star).
\end{equation}

\subsection{Proof of Proposition \ref{prop:structured_backward}}
\label{appendix:proof_structured_backward}

{\bf Proof outline.} Let $\bs{z}^\star = \proxtv(\bs{x})$ or
$\proxoscar(\bs{x})$.  We use the optimality conditions of $\proxtv$,
respectively $\proxoscar$
in order to express $\bs{z}^\star$ as an explicit
function of $\bs{x}$.  Then, obtaining the Jacobians of $\proxtv(\bs{x})$ and
$\proxoscar(\bs{x})$ follows by application of the chain rule to the two
expressions. We discuss the proof for points where $\proxtv$ and $\proxoscar$
are differentiable; on the (zero-measure) set of nondifferentiable points (i.e.\
where the group structure changes) we may take one of Clarke's generalized
gradients \cite{clarke_book}.

\textbf{Jacobian of $\proxtv$.}

\begin{lemma}
    Let $\bm{z}^\star = \proxtv(\bm{x}) \in \mathbb{R}^d$ and $G^\star_i$ be the
    set of indices around $i$ with the same value at the optimum, as defined in
    \S\ref{sec:structured_attention_fused_oscar}. Then, we have

    \begin{equation}
        z^\star_i = \frac{\sum_{j \in G^\star_i} x_j + \lambda(s_{a_i} -
        s_{b_i})}{|G^\star_i|},
        \label{eq:fused_fwd_expr}
    \end{equation}
    where $a_i = \min G^\star_i, b_i = \max G^\star_i$ are the boundaries
    of segment $G^\star_i$, and
    \begin{equation}
        s_{a_i} = \begin{cases}
            0 & \mbox{if } a = 1, \\
            \operatorname{sign}(z^\star_{{a_i}-1} - z^\star_i) & \mbox{if }a > 1 \\
        \end{cases} \qquad \text{and} \qquad s_{b_i} = \begin{cases}
            0 & \mbox{if } b = d, \\
            \operatorname{sign}(z^\star_i - z^\star_{b_i + 1}) & \mbox{if } b < d \\
        \end{cases}.
    \end{equation}
    \label{lemma:fused_fwd}
\end{lemma}

To prove Lemma~\ref{lemma:fused_fwd}, we make use of the
optimality conditions of the fused lasso proximity operator
\cite[Equation 27]{pathwise}, which state that $\bs{z}^\star$ satisfies
\begin{equation}
    z^\star_j - x_j + \lambda(t_j - t_{j + 1}) = 0, \quad \text{where} \quad
    t_j \in \begin{cases}
        \{0\} & \mbox{if } i \in \{1, d\}, \\
        \{\text{sign}(z^\star_j - z^\star_{j-1})\} & \mbox{if } z^\star_j \neq z^\star_{j-1}, \\
        [-1, 1] & \text{o.w}. \\
\end{cases}
\qquad \forall j \in [d].
\label{eq:subgrad}
\end{equation}
The optimality conditions \eqref{eq:subgrad} form a system with unknowns
$z^\star_j, t_j$ for $j \in [d]$. To express $\bs{z}^\star$ as a function of
$\bs{x}$, we shall now proceed to eliminate the unknowns $t_j$.

Let us focus on a particular segment $G^\star_i$. For readability, we drop the
segment index $i$ and use the shorthands $z \coloneqq z^\star_i, a \coloneqq a_i,$
and $b \coloneqq b_i$. By definition, $a$ and $b$ satisfy\vspace{0.15cm}
\begin{equation}
    z^\star_j = z \quad \forall a \leq j \leq b, \qquad 
    z^\star_{a - 1} \neq z ~ \text{if } a > 1, \qquad 
    z^\star_{b + 1} \neq z ~ \text{if } b < d. 
\end{equation}

It immediately follows from the definition of $t_j$ in \eqref{eq:subgrad} that 
\begin{equation}
t_a = \begin{cases}
    0 & \mbox{if } a = 1, \\
    \operatorname{sign}(z- z^\star_{a-1}) & \mbox{if }a > 1 \\
\end{cases}\quad\mbox{and}\quad
t_{b+1} = \begin{cases}
    0 & \mbox{if }b = d, \\
    \operatorname{sign}(z^\star_{b+1} - z) & \mbox{if }b < d \\
\end{cases}. 
\end{equation}
In other words, the unknowns $t_a$ and $t_b$ are already uniquely determined.
To emphasize that they are known, we introduce $s_a \coloneqq t_a$
and $s_b \coloneqq t_{b+1}$, leaving $t_j$ only unknown for $a < j \leq b$.

By rearranging the optimality conditions \eqref{eq:subgrad} we obtain the
recursion
\begin{equation}
    \lambda t_j = x_j - z + \lambda t_{j + 1} \quad \forall a \leq j \leq b.
    \label{eq:recur}
\end{equation}
We start with the first equation in the segment (at $j=a$), and unroll the
recursion until reaching the stopping condition $j=b$.
\begin{equation}
    \begin{aligned}
        \lambda s_a &= x_a - z + \lambda t_{a+1} \\
                    &= x_a - z + x_{a+1} - z + \dots + x_b - z + \lambda s_b \\
                    &= \sum_{k=a}^{b} x_k - (b - a + 1)z + \lambda s_b
\end{aligned}
\end{equation}
Rearranging the terms, we obtain the expression
\begin{equation}
    z = \frac{\sum_{k=a}^{b} x_k + \lambda(s_b - s_a)}{b - a + 1}.
\end{equation}
Applying this calculation to each segment in $\bm{z}^\star$ yields the desired
result. \hfill $\square$

The proof of Proposition~\ref{prop:structured_backward} follows by applying
the chain rule to \eqref{eq:fused_fwd_expr}, noting that the groups $G^*_i$
are constant within a neighborhood of $\bm{x}$ (observation also used for OSCAR
in \cite{oscar}).  Therefore, for $\proxtv$,
\begin{equation}
    \partialfrac{z^\star_i}{x_j} = \frac{1}{|G^\star_i|}
    \left(\sum_{k \in G_i^\star} \partialfrac{x_k}{x_j} +
        \lambda \left( \partialfrac{s_b}{x_j} 
        -\partialfrac{s_a}{x_j}\right)  \right).
\end{equation}
Since $s_b$ and $s_a$ are either constant or sign functions w.r.t.\ $\bm{x}$,
their partial derivatives are $0$, and thus 
\begin{equation}
    \partialfrac{z^\star_i}{x_j} = \begin{cases}
        \frac{1}{|G^\star_i|} & \mbox{if } j \in G^\star_i, \\
        0 & \text{o.w.} \\ 
    \end{cases}.
\end{equation}

\textbf{Jacobian of $\proxoscar$.}

\begin{lemma}(\cite[Theorem 1]{oscar_prox}, \cite[Proposition 3]{oscar_proof})
    Let $\bm{z}^\star = \proxoscar(\bm{x}) \in \mathbb{R}^d$ and $G^\star_i$ be the set of
    indices around $i$ with the same value at the optimum: 
    $G^\star_i = \{ j \in [d]: |z^\star_i| = |z^\star_j| \}.$ Then, we have

    \begin{equation}
        z^\star_i = \sign(x_i) \max \left( \frac{\sum_{j \in G^\star_i} |x_j|}{|G^\star_i|} -
        w_i, 0 \right),
        \label{eq:oscar_fwd_expr}
    \end{equation}
    \begin{equation}
    \text{where~}
    w_i = \lambda\left(d - \frac{u_i + v_i}{2}\right), \quad
    u_i = \left| \{j \in [d]: |z^\star_j| < |z^\star_i| \} \right|, \quad
    v_i = u_i + |G^\star_i|.
    \end{equation}
    \label{lemma:oscar_fwd}
\end{lemma}

Lemma~\ref{lemma:oscar_fwd} is a simple reformulation of Theorem 1, part
\textit{ii} from \cite{oscar_prox}.  With the same observation that the induced
groups do not change within a neighborhood of $\bm{x}$, we may differentiate
\eqref{eq:oscar_fwd_expr} to obtain
\begin{equation}
    \partialfrac{z^\star_i}{x_j} = \begin{cases}
        0 & \mbox{if } z^\star_i = 0, \\

        \displaystyle\frac{\sign(x_i)}{|G^\star_i|}
    \sum_{k \in G_i^\star} \partialfrac{|x_k|}{x_k} \partialfrac{x_k}{x_j} -
    \partialfrac{w_i}{x_j} & \text{o.w.}
\end{cases}.
\end{equation}

Noting that $\partialfrac{w_i}{x_j}=0$, as $w_i$ is derived only from group
indices and the term $\partialfrac{|x_k|}{x_k} \partialfrac{x_k}{x_j}$ either
vanishes (when $k \neq j$) or else equals $\sign(x_j)$ with $x_j \neq 0$, we 
substitute $\sign(z_j^\star)$ for $\sign(x_j)$ \cite{oscar_prox}
to get
\begin{equation}
\partialfrac{z^\star_i}{x_j} = \begin{cases}
    \displaystyle\frac{\sign(z^\star_i z^\star_j)}{|G_i^\star|}
        & \mbox{if } j \in G_i^\star \mbox{ and } z_i^\star \neq 0, \\
    0
        & \mbox{o.w. }
\end{cases}.
\end{equation}

\section{Computational complexity and implementation details}
\label{appendix:complex}

\subsection{Sparsemax}
Computing the forward and backward pass of sparsemax is a compositional
building block in fusedmax, oscarmax, as well as in the general case; for this
reason, we discuss it before the others.

{\bf Forward pass.} The problem is exactly the Euclidean projection on the
simplex, which can be computed exactly in worst-case $\mathcal{O}(d
\log d)$ due to the required sort \cite{sparsemax,michelot,duchi}, or in
expected $\mathcal{O}(d)$ time using a pivot algorithm similar to median
finding \cite{duchi}. Our implementation is based on sorting.

{\bf Backward pass.} From \cite{sparsemax} we have that the result of a
Jacobian-vector product $J_{\projop} v$ has the same sparsity pattern as
$\bs{y}^\star$.  If we denote the number of nonzero elements of $\bs{x}$ by
$\text{nnz}(\bs{x})$, we can see that $\hat{v}$ in \cite[eq. 14]{sparsemax},
and thus the Jacobian-vector product itself, can be computed in
$\mathcal{O}(\text{nnz}(\bs{y}^\star))$.

\subsection{Fusedmax}
We implement fusedmax as the composition of the fused lasso proximal operator
with sparsemax.

{\bf Forward pass.} We need to solve the proximal operator of fused lasso.
The algorithm we use is $\mathcal{O}(d^2)$ in the worst case, but has
strong performance on realistic benchmarks, close to $\mathcal{O}(d)$
\cite{tv1d_prox}.

{\bf Backward pass.} Due to
the structure of the Jacobian and the locality of fused groups, Jacobian-vector
products $J_{\projop} v$ can be computed
in $\mathcal{O}(d)$ using a simple algorithm that iterates over the output
$\bs{y}^\star$ and the vector $\bs{v}$ simultaneously,
averaging the elements of $\bs{v}$ whose indices map to fused elements of
$\bs{y}^\star$. Since only consecutive elements can be fused, this amounts to
resetting to a new group as soon as we encounter an index $i$ such that
$y^\star_i \neq y^\star_{i-1}$.

\subsection{Oscarmax}

We implement oscarmax as the composition of the OSCAR proximal operator
with sparsemax.

{\bf Forward pass.} The proximal operator of the OSCAR penalty can be computed
in $\mathcal{O}(d \log d)$ as a particular case of the ordered weighted
$\ell_1$ (OWL) proximal operator, using an algorithm involving a sort
followed by isotonic regression \cite{owl}.

{\bf Backward pass.} The algorithm is similar in spirit to fusedmax, but
because groups can reach across non-adjacent indices, a single pass is not
sufficient. With no other information other than $\bf{y}^\star$,
the backward pass can be computed in $\mathcal{O}(d \log d)$ using a stable
sort followed by a linear-time pass for finding groups. Further optimization is
possible if group indices may be saved from the forward pass.

\subsection{General case and sq-pnorm-max}

{\bf Forward pass.} For general $\projop$ we may use any projected gradient
solver; we choose FISTA \cite{fista}. Each iteration requires a projection onto
the simplex; in the case of sq-pnorm-max, this dominates every
iteration, leading to a complexity of $\mathcal{O}(td \log d)$ where $t$ is the
number of iterations performed.

{\bf Backward pass.} To compute Jacobian-vector products we solve the
linear system from Proposition~\ref{prop:Jacobian_diff_regul}:
$(I + A(B - I)) ~ (J_{\projop}\bs{v}) = A\bs{v}$.
This is a $d \times d$ system, which at first sight suggests a complexity of
$\mathcal{O}(d^3)$. However, we can use the structure of $A$ to solve it more
efficiently.

The matrix $A$ is defined as $A \coloneqq J_{\projsimplex}(\bs{y}^\star - \nabla
f(\bs{y}^\star))$.
As a sparsemax Jacobian, $A$ is row-
and column-sparse, and uniquely defined by its sparsity pattern. By splitting the
system into equations corresponding to zero and nonzero rows of $A$, we obtain
that the solution $J_{\projop}\bs{v}$ must have the same sparsity pattern as the
row-sparsity of $A$, therefore we only need to solve a subset of the system.
From the fixed-point iteration 
$ \bs{y}^* = \projsimplex(\bs{y}^\star - \nabla f(\bs{y}^\star))$,
we have that the row-sparsity of $A$ is the same as
the sparsity of the forward pass solution $\bs{y}^*$. The
backward pass complexity is thus $\mathcal{O}(\text{nnz}(\bs{y}^*)^3)$.

\section{Additional experimental results}

\subsection{Visualizing attention mappings in 3-d}
\label{appendix:3d_plots}

\begin{figure}[h]
    \centering \includegraphics[width=\textwidth]{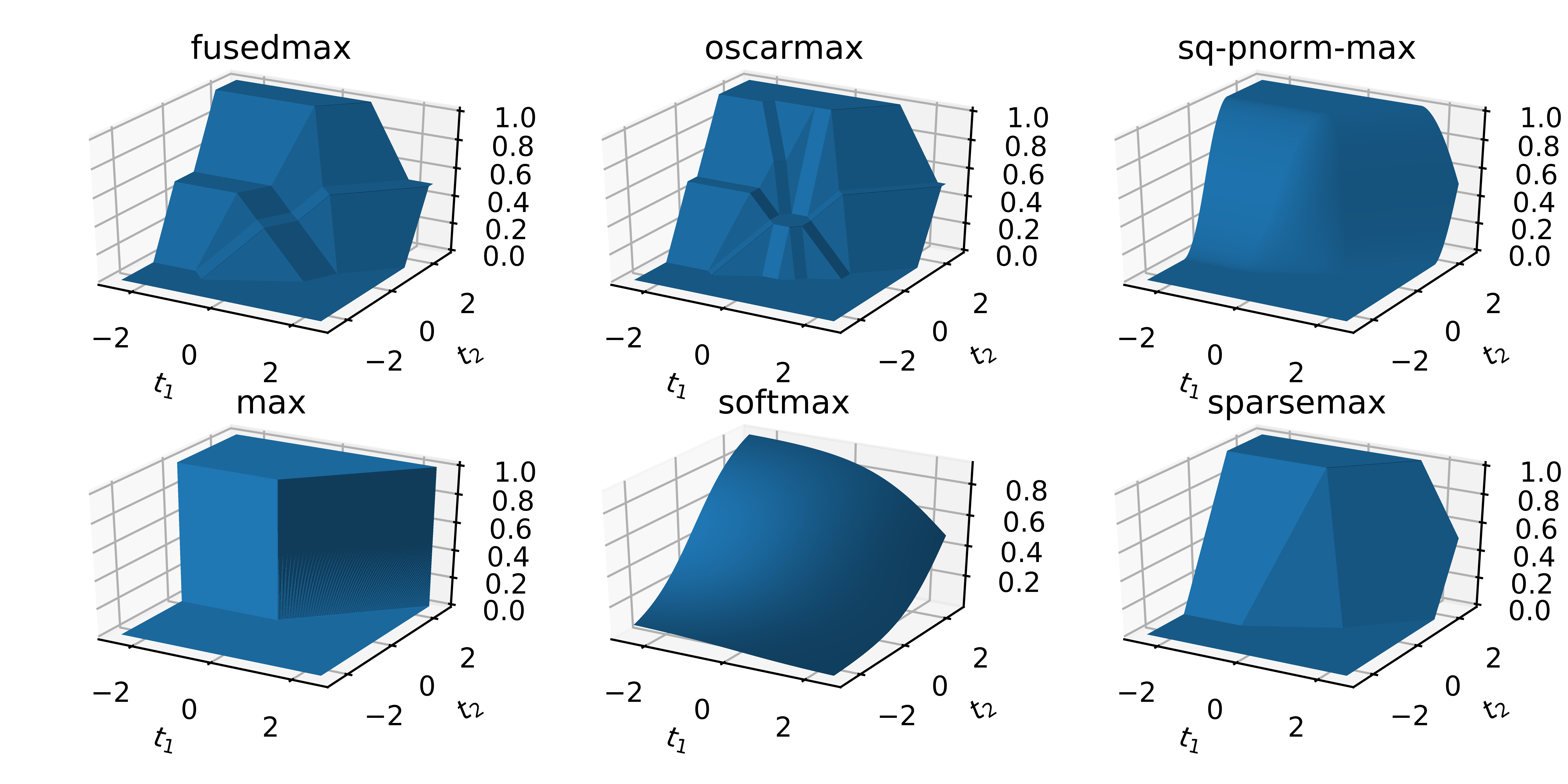}
    \caption{3-d visualization of $\projop([t_1, t_2, 0])_2$ for several
        proposed and existing mappings $\projop$. \mbox{sq-pnorm-max} with $p=1.5$
        resembles sparsemax but with smoother transitions. The proposed
        structured attention mechanisms, fusedmax and oscarmax, exhibit plateaus and
ridges in areas where weights become fused together. We set $\gamma=1$ and
$\lambda=0.2$.}
\end{figure}

\subsection{Textual entailment results}
\label{appendix:textual_entailment}

{\bf Experimental setup.} We build upon the implementation from
\cite{sparsemax}, which is a slight variation of the attention model from
\cite{rockt}, using GRUs instead of LSTMs.  The GRUs encoding the premise and
hypothesis have separate parameters, but the hypothesis GRU is initialized with
the last state of the premise GRU.  We use the same settings and methodology as
\cite{sparsemax}: we use fixed 300-dimensional GloVe vectors, we train for 200
epochs using ADAM with learning rate $3 \cdot 10^{-4}$, we use a drop-out
probability of 0.1, and we choose an $l_2$ regularization coefficient from $\{0,
10^{-4}, 3 \cdot 10^{-4}, 10^{-3}\}$. Experiments are performed on machines
with 2$\times$Xeon X5675 3.06GHz CPUs and 96GB RAM.

\begin{figure}[h]
    \centering
    \includegraphics[width=.65\textwidth]{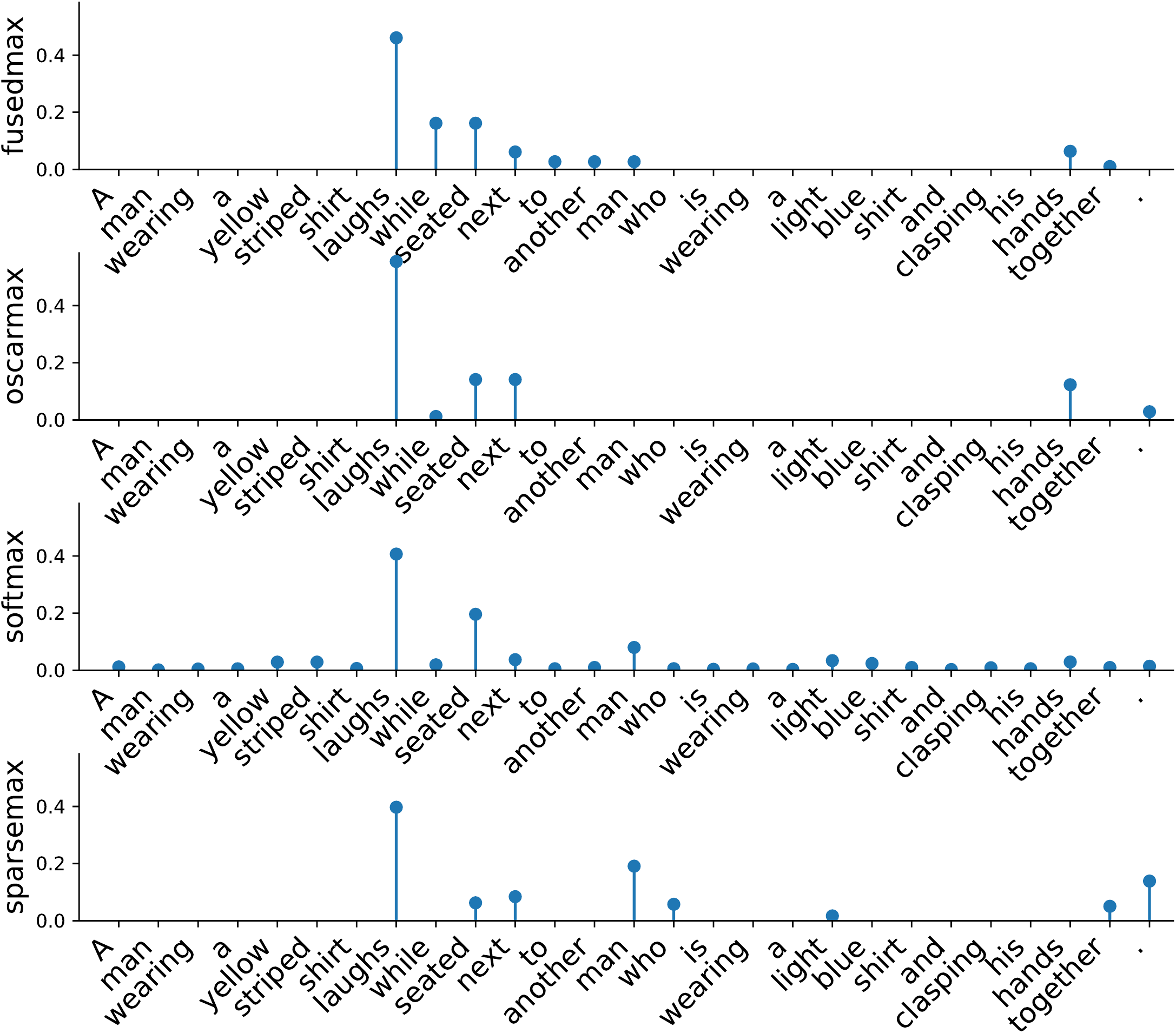}
    \\[0.6cm]
    \includegraphics[width=.38\textwidth]{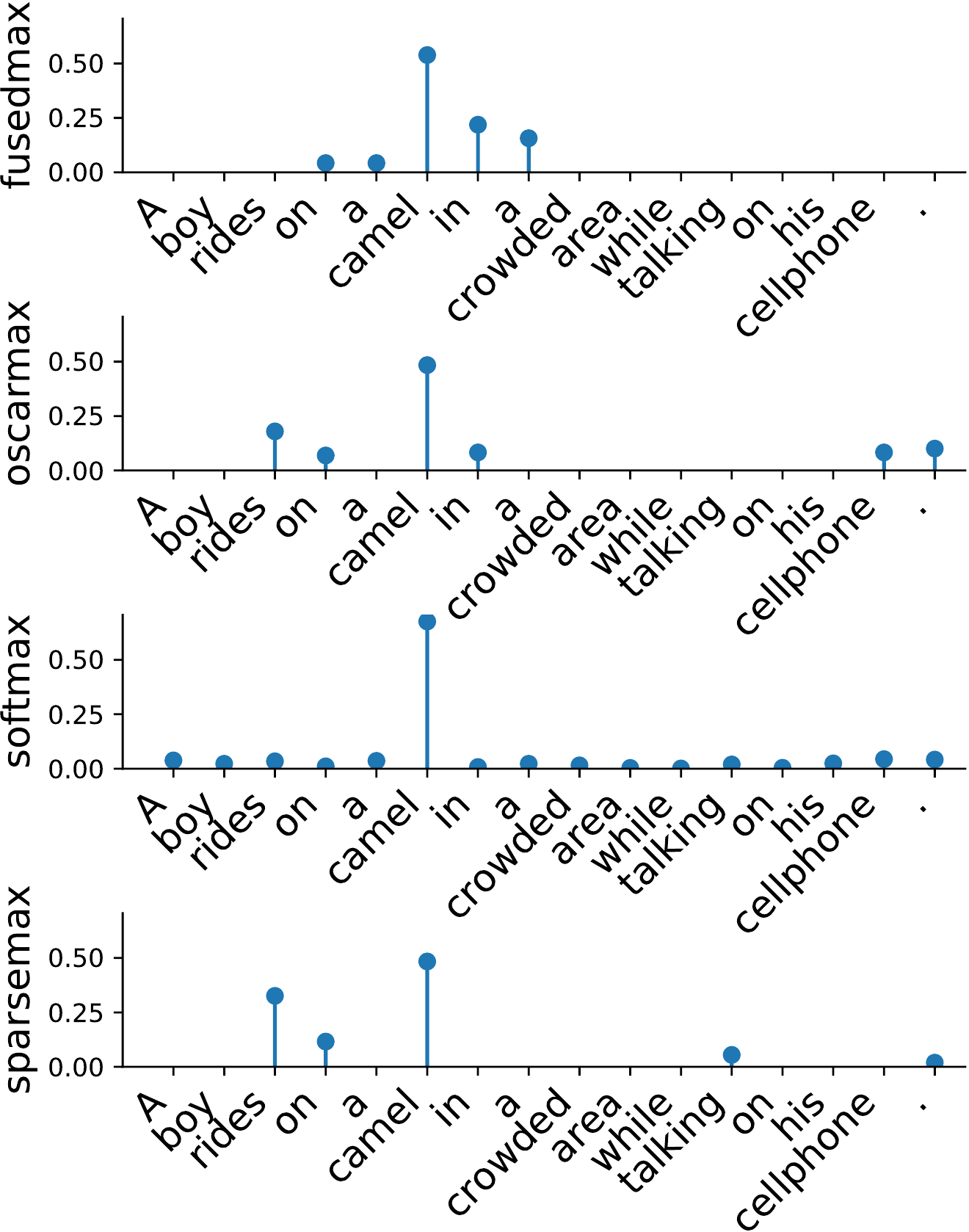}\hspace{1cm}
    \includegraphics[width=.26\textwidth]{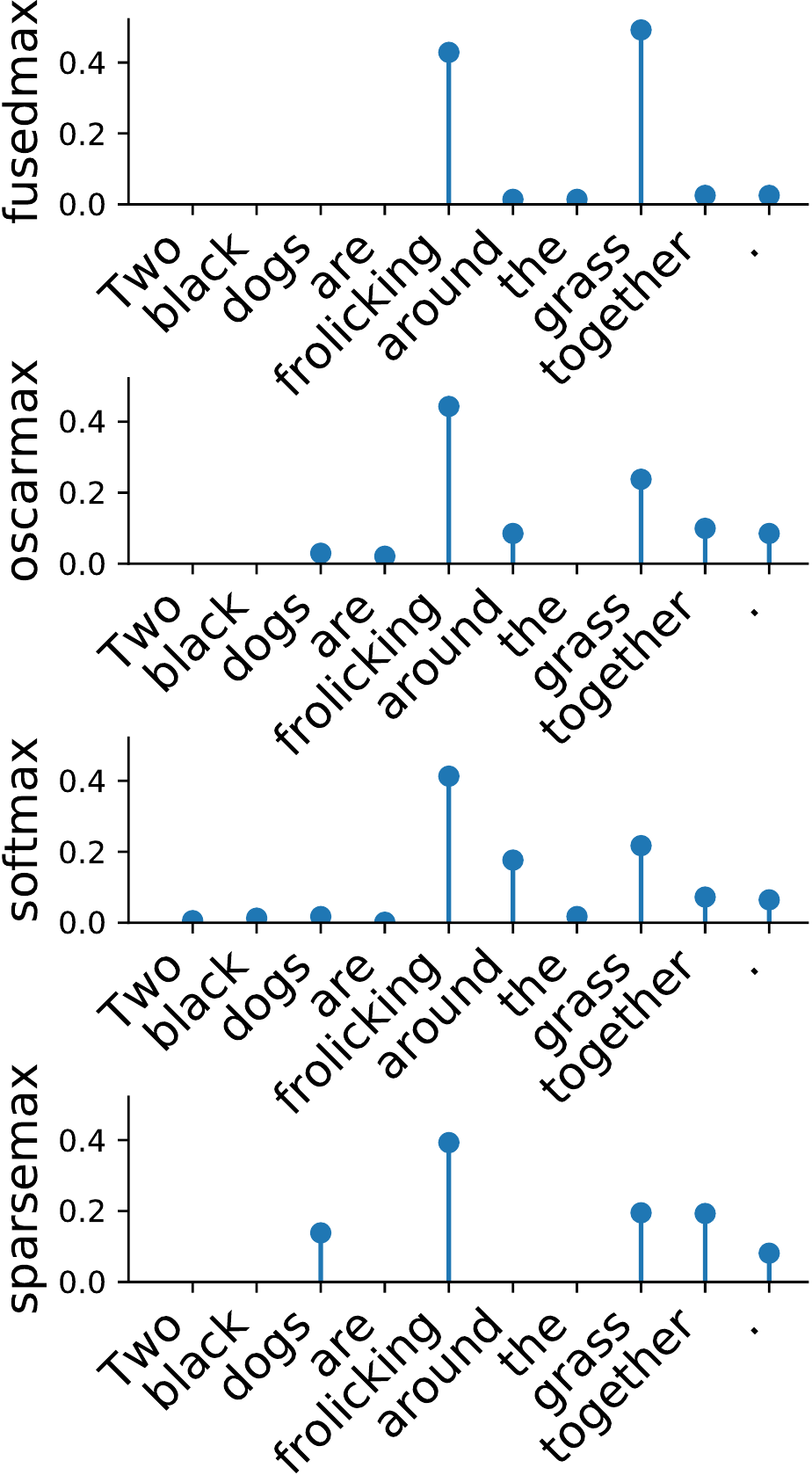}
    \caption{Attention weights on several examples also used in
        \cite{rockt,sparsemax}. The hypotheses considered are
        ``Two mimes sit in complete silence.'' (top), 
        ``A boy is riding an animal.'' (left), and ``Two dogs swim in the
        lake.'' (right). All attention mechanisms result in correct
        classifications (top: contradiction; left: entailment; right:
        contradiction). As can be seen, fusedmax prefers contiguous support
    segments even when not all weights are tied.\label{fig:supp_snli}}
\end{figure}

{\bf Dataset and preprocessing.} We use the SNLI v1 dataset \cite{snli}. We
apply the minimal preprocessing from \cite{sparsemax},
skipping sentence pairs with missing labels and using the provided tokenization.
This results in a training set of 549,367 sentence pairs, a development set of
9,842 sentence pairs and a test set of 9,824 sentence pairs.
We report timing measurements in Table~\ref{table:snli_timing} and
visualizations of the produced attention weights in Figure~\ref{fig:supp_snli}.

\begin{table}[h]
\begin{minipage}[c]{0.30\textwidth}
\centering\small
    \begin{tabular}{r r@{ }l}
    \toprule
    attention & \multicolumn{2}{c}{time per epoch} \\
    \midrule
    softmax   &      1h 26m 40s & $\pm$ 51s \\
    sparsemax &      1h 24m 21s & $\pm$ 54s \\
    \spacerule
    fusedmax  &      1h 23m 58s & $\pm$ 50s \\
    oscarmax  &      1h 23m 19s & $\pm$ 50s \\
    \bottomrule
\end{tabular}
\end{minipage}\hfill
\begin{minipage}[c]{0.65\textwidth}
\mbox{}\\[-\baselineskip]
\caption{Timing results for training textual entailment on SNLI, using the
implementation and experimental setup from \cite{sparsemax}.  With this C++ CPU
implementation, fusedmax and oscarmax are as fast as sparsemax, and all three
sparse attention mechanisms are slightly faster than
softmax.\label{table:snli_timing}}
\end{minipage}
\end{table}

\subsection{Machine translation results}
\label{appendix:machine_translation}

{\bf Experimental setup.} Because our goal is to demonstrate that our attention
mechanisms can be drop-in replacements for existing ones, we focus on
OpenNMT-py with default settings for all of our sequence-to-sequence
experiments. These defaults are: an unidirectional LSTM, 500 dimensions for the
word vectors and for the LSTM hidden representations, drop-out probability
of 0.3, global attention, and input-feeding \cite{nmt_luong}.
Following the default, we train our models for 13 epochs with
stochastic gradient updates (batches of size 64 and initial learning
rate of 1, halved every epoch after the 8\textsuperscript{th}).  Weights
(including word embeddings) are initialized uniformly over $[-0.1, 0.1]$, and
gradients are normalized to have norm 5 if their norm exceeds this value.  For
test scores and visualizations, we use the model snapshot at the epoch with the
highest validation set accuracy.
All of the experiments in this section are performed on machines equiped with
Xeon E5 CPUs and Nvidia Tesla K80 GPUs.

{\bf Datasets.} We employ training and test datasets from multiple sources.
\begin{itemize}
    \item {\footnotesize BENCHMARK}: Training, validation, and test data from the
        NMT-Benchmark project (\url{http://scorer.nmt-benchmark.net/}). All
        languages have \textasciitilde 1M training sentence pairs, and equal
        validation and test sets of size 1K (French) and 2K (Italian, Dutch
        and Swedish).

    \item {\footnotesize BENCHMARK}$^+$: Training and validation data as above, but
        testing on all available {\em newstest} data. For Italian we use the
        2009 data (\textasciitilde 2.5K sentence pairs), and for French we
        concatenate 2009--2014 (\textasciitilde 11K sentence pairs).

    \item {\footnotesize WMT16,
        WMT17}: Translation tasks at the first and second ACL Conferences for
        Machine Translation, available at
        \url{http://www.statmt.org/wmt16/translation-task.html} and
        \url{http://www.statmt.org/wmt17/translation-task.html}.
        Training, validation, and test sizes are, approximately, for Romanian
        400K/2K/2K, for German 5.8M/6K/3K, for Finnish 2.6M/2K/2K, for Latvian
        4.5M/2K/2K, and for Turkish 207K/1K/3K. 
\end{itemize}

We use the preprocessing scripts from Moses \cite{moses}
for tokenization, and, where needed, SGML parsing. We limit source and target
vocabulary sizes to 50K lower-cased tokens and prune sentences longer than 50
tokens at training time and 100 tokens at test time.  We do not perform
recasing.

We report BLEU scores in Table~\ref{table:nmt} and showcase the enhanced
interpretability induced by our proposed attention mechanisms in
Figure~\ref{fig:nmt_supp}.
Timing measurements can be found in Table~\ref{table:nmt_time}.

\setlength{\tabcolsep}{0.2em}
\begin{table}[h]
    \caption{Neural machine translation results: tokenized BLEU scores on test data.\label{table:nmt}}
    \small
    \centering

    \begin{tabular}{r c c c c c c c c c c c c}
        \toprule
        & \multicolumn{4}{c}{\scriptsize BENCHMARK} &
        \multicolumn{2}{c}{\scriptsize BENCHMARK$^+$} &
        {\scriptsize WMT16} &
        \multicolumn{4}{c}{\scriptsize WMT17} \\ 
        \cmidrule(lr){2-5} \cmidrule(lr){6-7} \cmidrule(lr){8-8}
        \cmidrule(lr){9-12}
        & \lang{fr} & \lang{it} & \lang{nl} & \lang{sv} & \lang{fr} & \lang{it}
        & \lang{ro} & \lang{de} & \lang{fi} & \lang{lv} & \lang{tr} \\
        \midrule


\multicolumn{13}{l}{\bf from English} \\
\addlinespace[0.2em]

softmax      	&      36.94  &	     37.20  &	     36.12  &	     34.97  &	     27.13  &	{\bf 24.86} &	     17.71 &	     22.32  &	     14.54  &	     11.02  &	{\bf 11.95} &	\\
sparsemax    	&      37.03  &	     37.21  &	     36.12  &	{\bf 35.09} &	     26.99  &	     24.49  &	     17.61 &	{\bf 22.43} &	{\bf 14.85} &	     11.07  &	     11.66  &	\\
\spacerule                                                                                                                 
fusedmax     	&      37.08  &	     36.73  &	     36.04  &	     34.30  &	     26.89  &	     24.47  &	     17.19 &	     22.25  &	     14.28  &	{\bf 11.27} &	     11.32  &	\\
oscarmax     	&      36.66  &	     36.89  &	     35.96  &	     34.86  &	     27.02  &	     24.76  &	     17.26 &	     22.42  &	     14.02  &	     11.19  &	     11.63  &	\\
sq-pnorm-max 	& {\bf 37.16} &	{\bf 37.39} &	{\bf 36.21} &	     34.63  &	{\bf 27.25} &	     24.56  &	{\bf 17.80}&	     -----  &	     14.45  &	     -----  &	     11.58  &	\\

\addlinespace[1em]
\multicolumn{13}{l}{\bf to English} \\
\addlinespace[0.2em]

softmax      	&      36.79  &	     39.95  &	     40.06  &	     37.96  &	     25.72  &	     25.37  &	     17.86  &	{\bf 25.82} &	     15.11  &	     13.60  &	     11.78  &	\\
sparsemax    	& {\bf 36.91} &	     40.13  &	     40.25  &	     38.09  &	     25.97  &	     25.62  &	     17.46  &	     25.76  &	     14.95  &	     13.59  &	{\bf 12.04} &	\\
\spacerule
fusedmax     	&      36.64  &	     39.64  &	     39.87  &	     37.83  &	     25.72  &	     25.41  &	{\bf 18.29} &	     25.58  &	     15.08  &	     13.53  &	     11.91  &	\\
oscarmax     	&      36.90  &	     40.05  &	     40.17  &	{\bf 38.12} &	{\bf 26.13} &	     25.65  &	     17.89  &	     25.69  &	     14.94  &	{\bf 13.71} &	     11.70  &	\\
sq-pnorm-max 	&      36.84  &	{\bf 40.23} &	{\bf 40.48} &	{\bf 38.12} &	     25.72  &	{\bf 25.70} &	     17.44  &	     -----  &	{\bf 15.20} &	     -----  &	     11.93  &	\\

\bottomrule
    \end{tabular}
\end{table}
\vspace{1.5em} 

\begin{table}[h]
\begin{minipage}[c]{0.30\textwidth}
\centering
\small
\begin{tabular}{r l}
    \toprule
    attention & time per epoch \\
    \midrule
    softmax & 2h \\
    sparsemax & 2h 18m \\
    \spacerule
    fusedmax & 3h 5m \\
    oscarmax & 3h 25m \\
    sq-pnorm-max & 7h 5m \\
    \bottomrule
\end{tabular}
\end{minipage}\hfill
\begin{minipage}[c]{0.66\textwidth}
\mbox{}\\[-\baselineskip] \caption{\label{table:nmt_time} Timing results for
    French-to-English translation using OpenNMT-py (all standard errors are
    under 2 minutes).  For simplicity, all attention mechanisms, except softmax,
    are implemented on the CPU, thus incurring memory copies in both directions.
    (The rest of the pipeline runs on the GPU.) Even without special
    optimization, sparsemax, fusedmax, and oscarmax are practical, taking within
    1.75x the training time of a softmax model on the GPU.  }
\end{minipage}
\end{table}
\clearpage
\begin{SidewaysFigure}
    \begin{tabular}{c c c c c}
    fusedmax & oscarmax & sq-pnorm-max & softmax & sparsemax \\
    \midrule
    \includegraphics[width=0.3\textwidth]{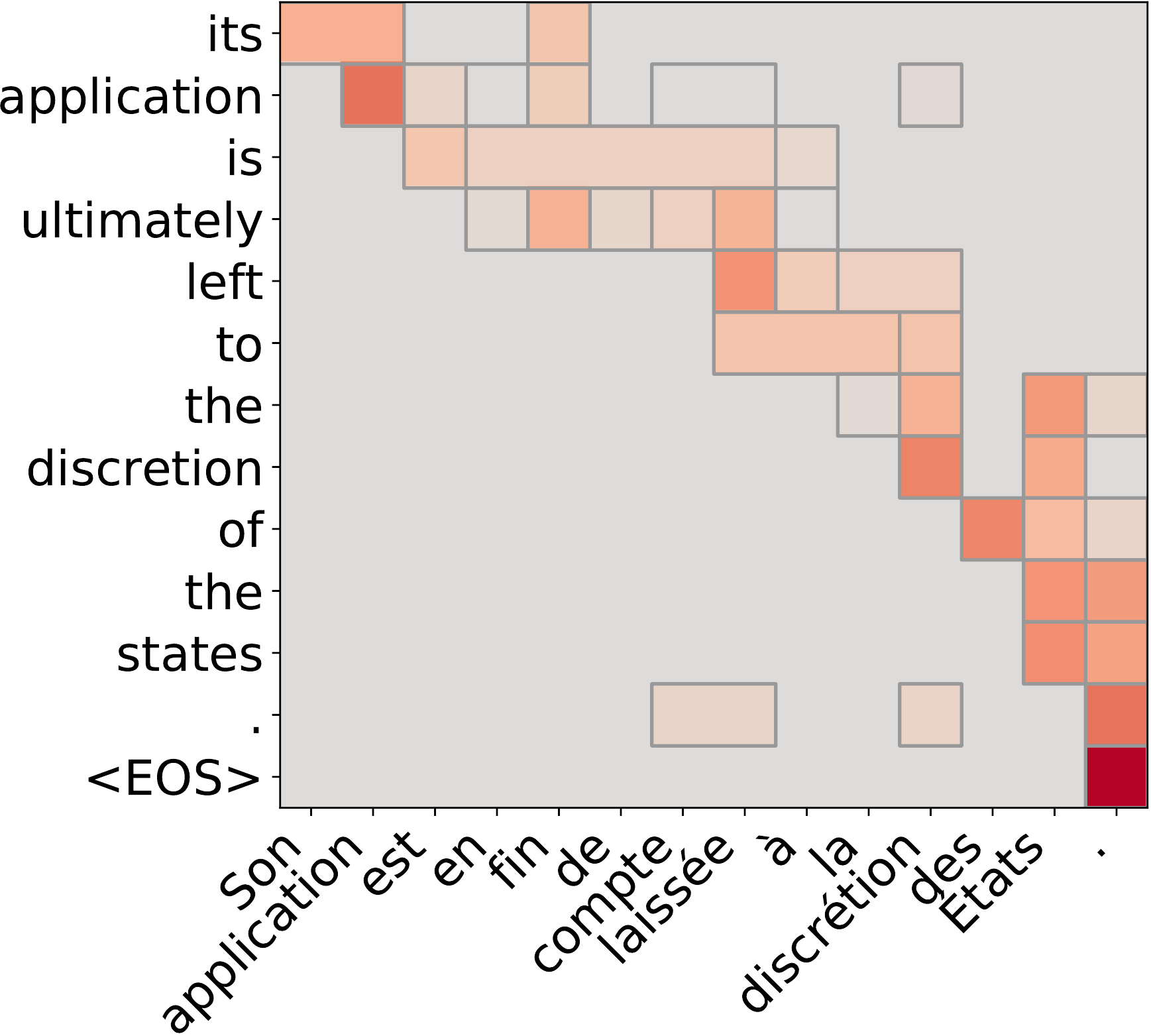} & 
    \includegraphics[width=0.3\textwidth]{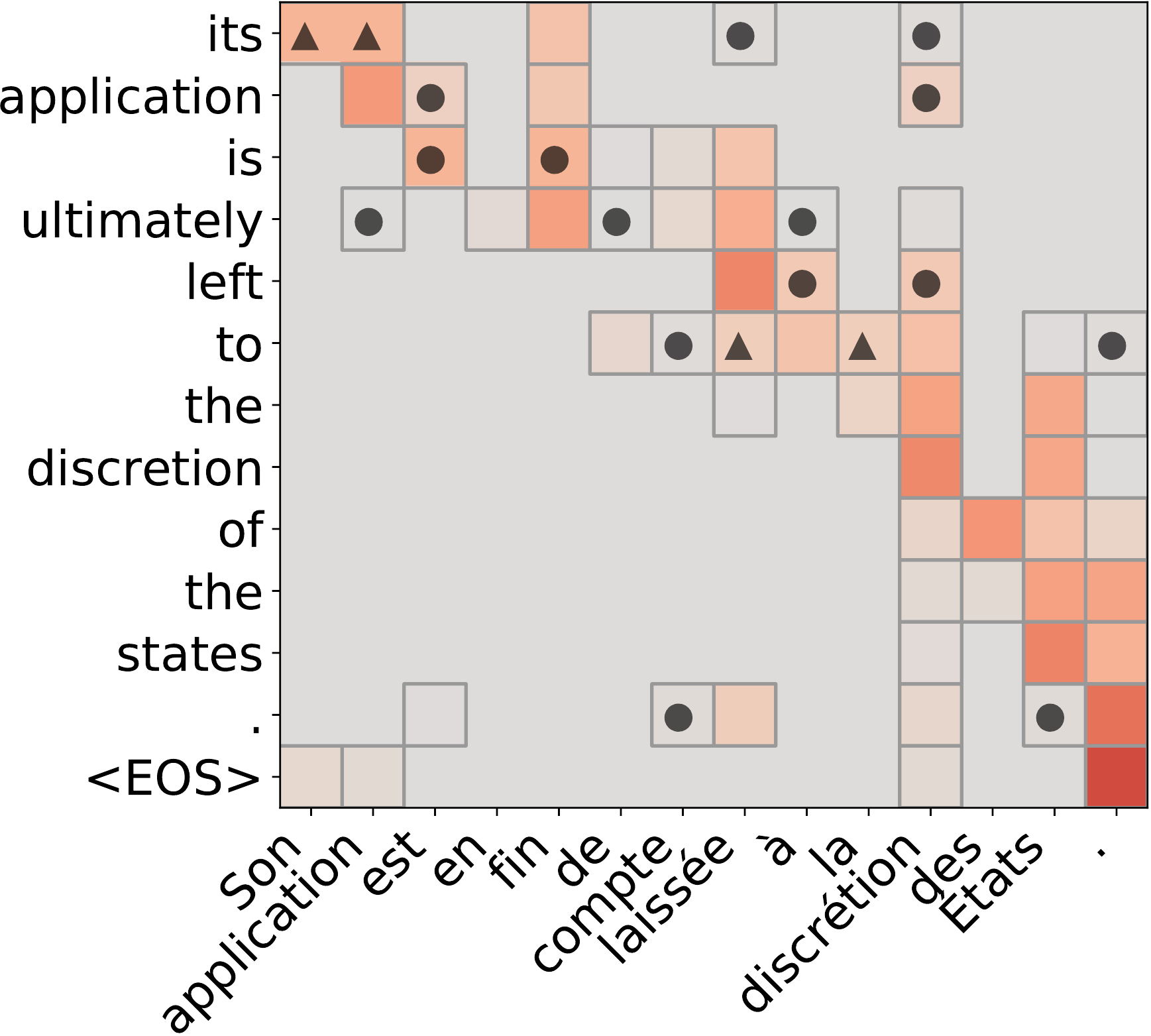} & 
    \includegraphics[width=0.3\textwidth]{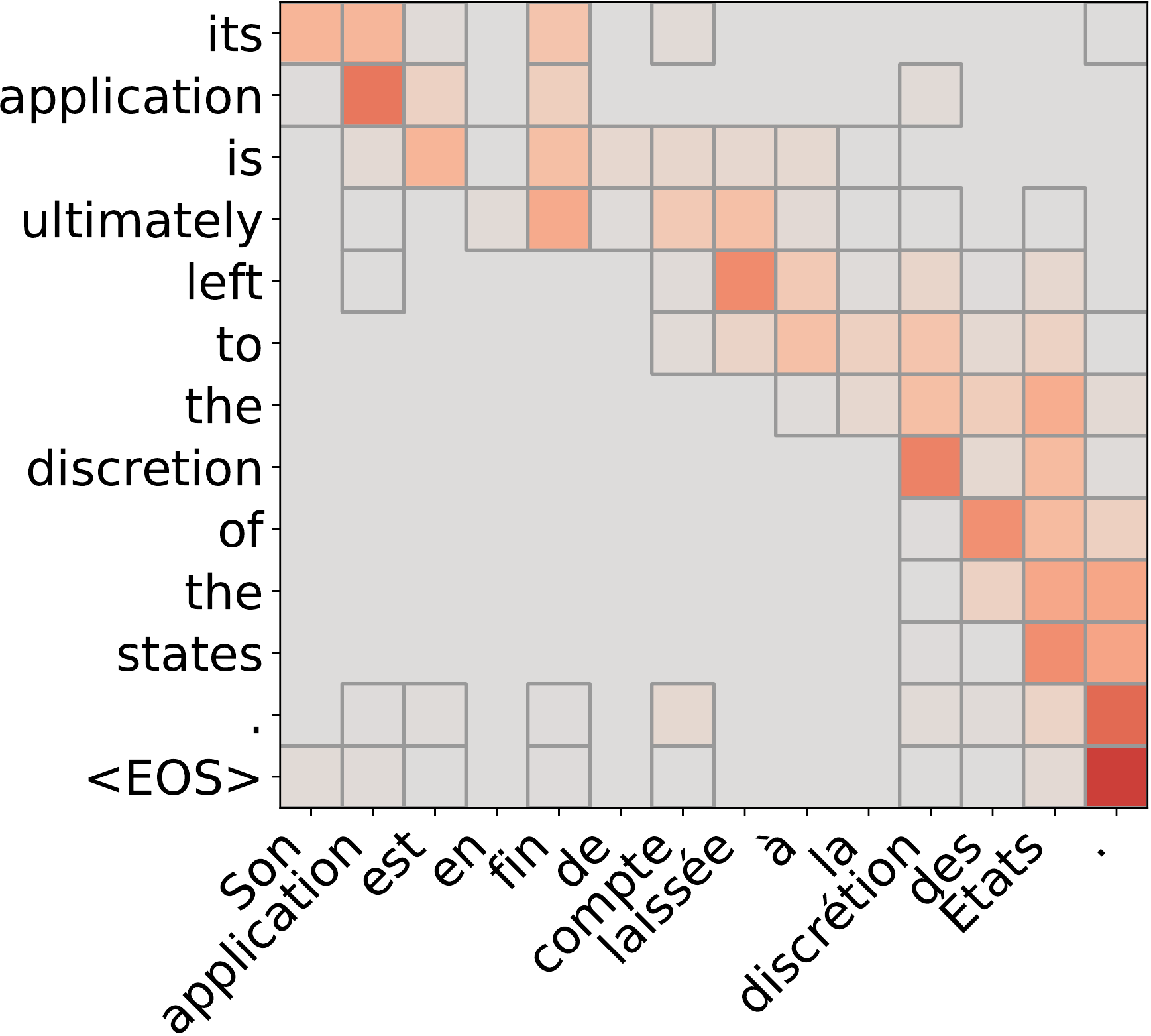} & 
    \includegraphics[width=0.3\textwidth]{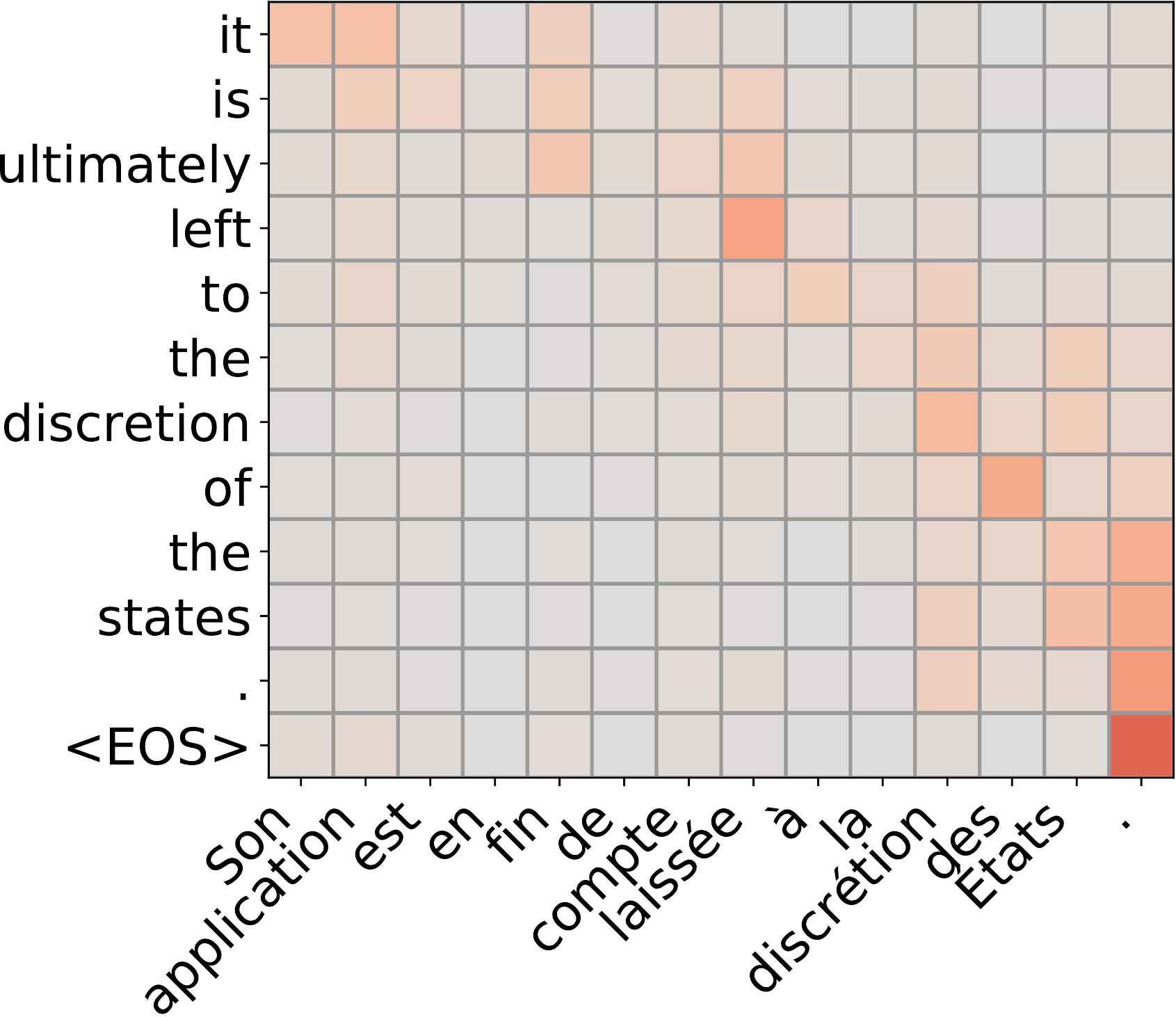} & 
    \includegraphics[width=0.3\textwidth]{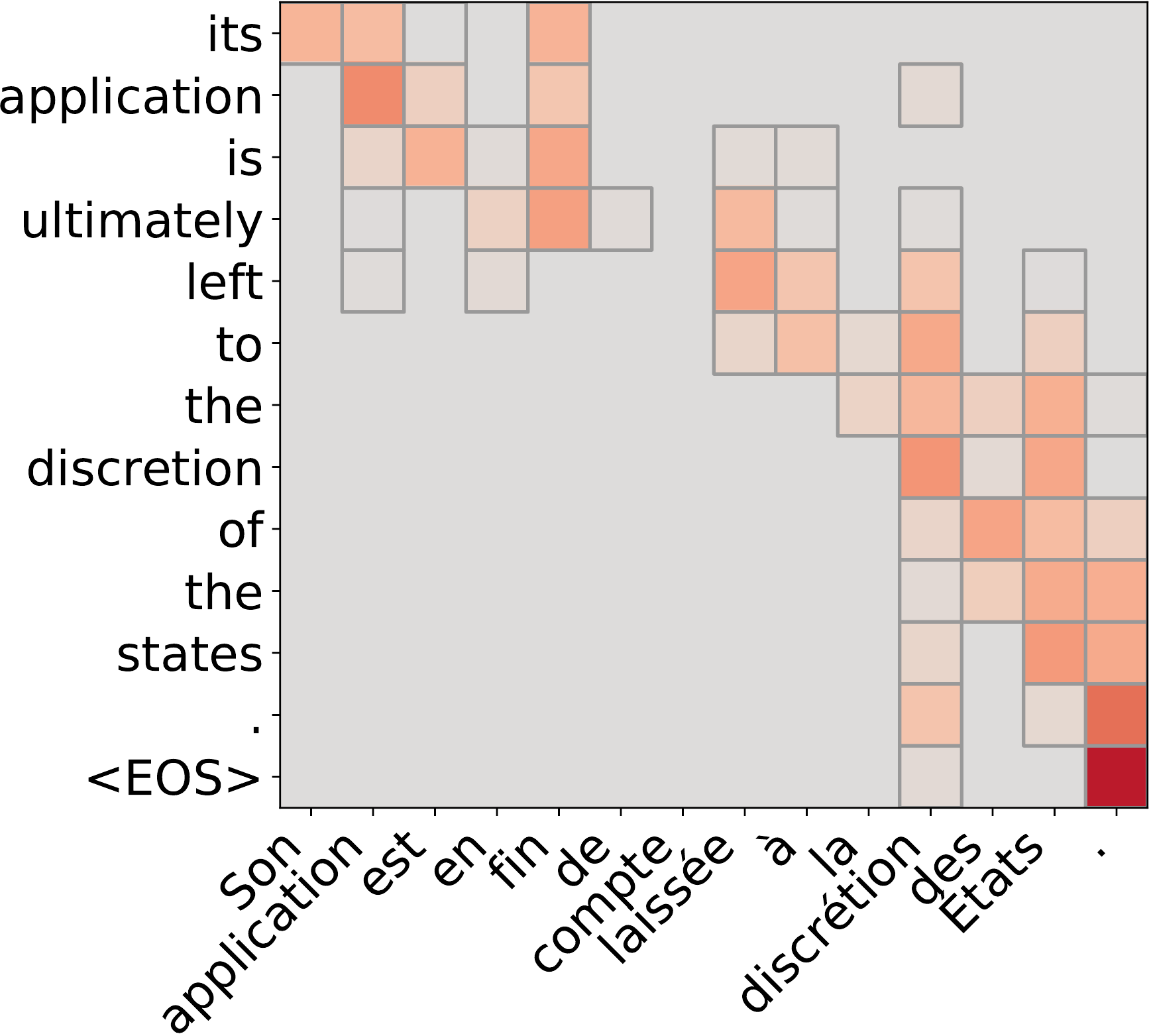}
    \\[0.7cm]
    \includegraphics[width=0.3\textwidth]{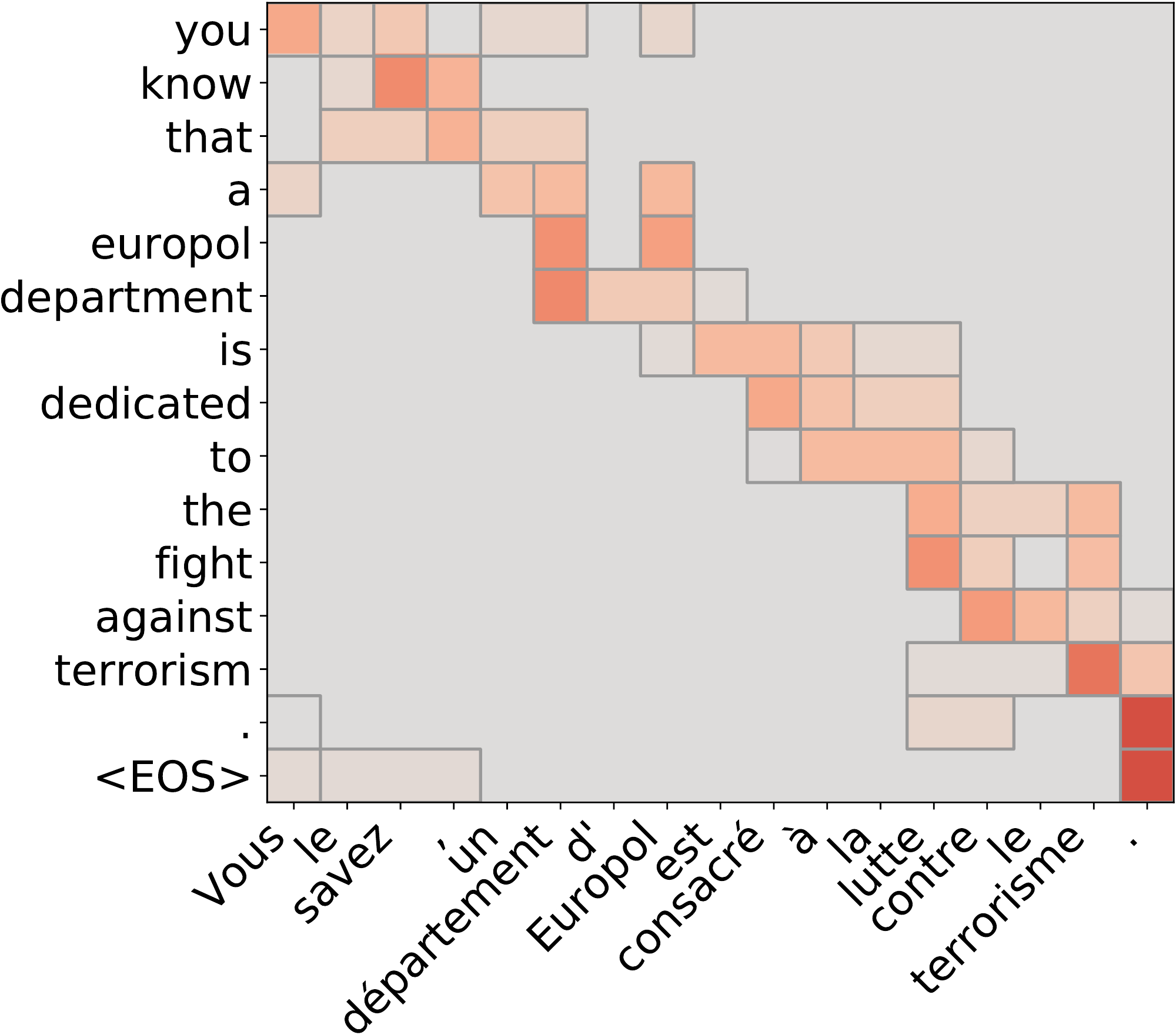}&
    \includegraphics[width=0.3\textwidth]{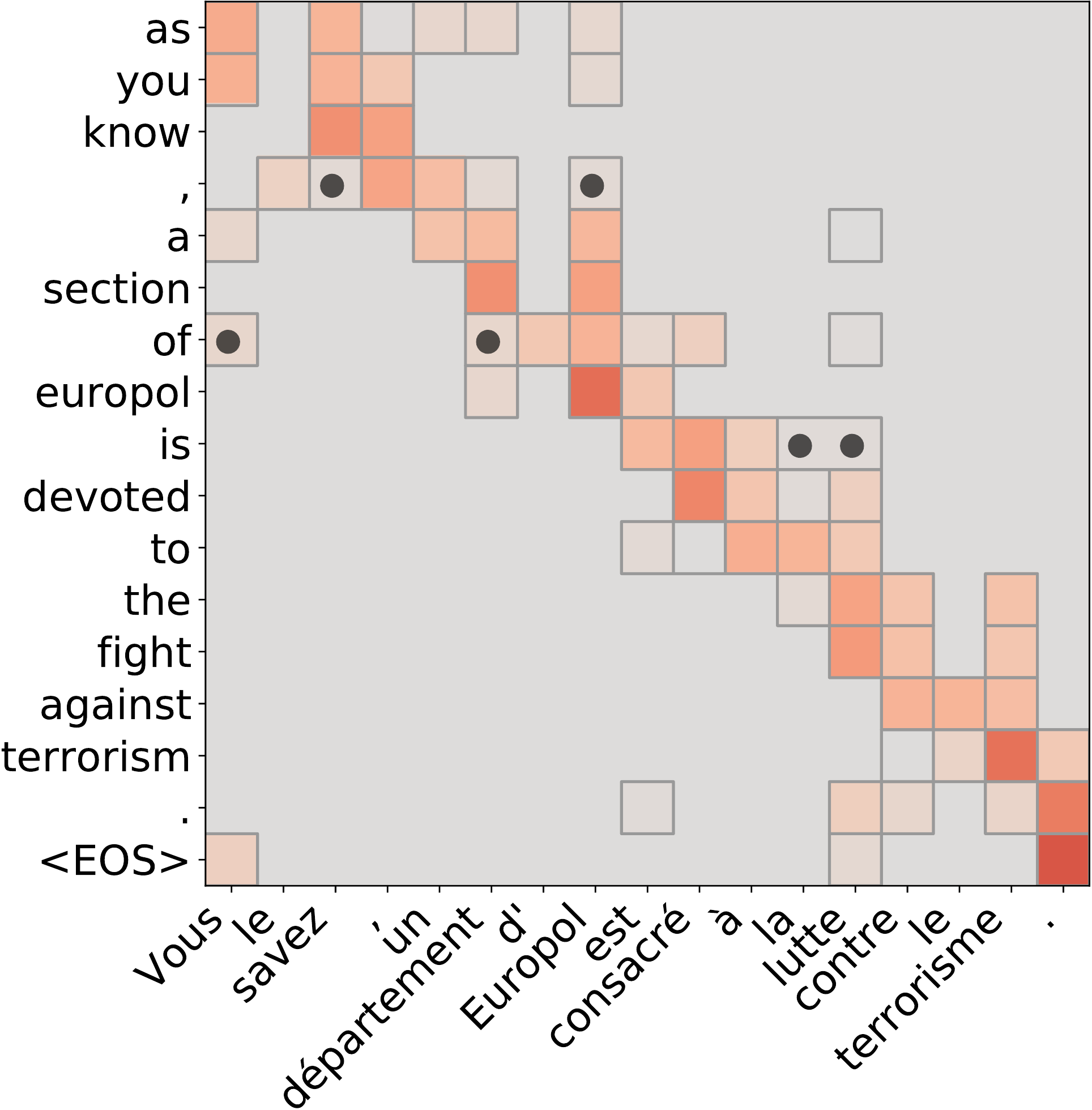}&
    \includegraphics[width=0.3\textwidth]{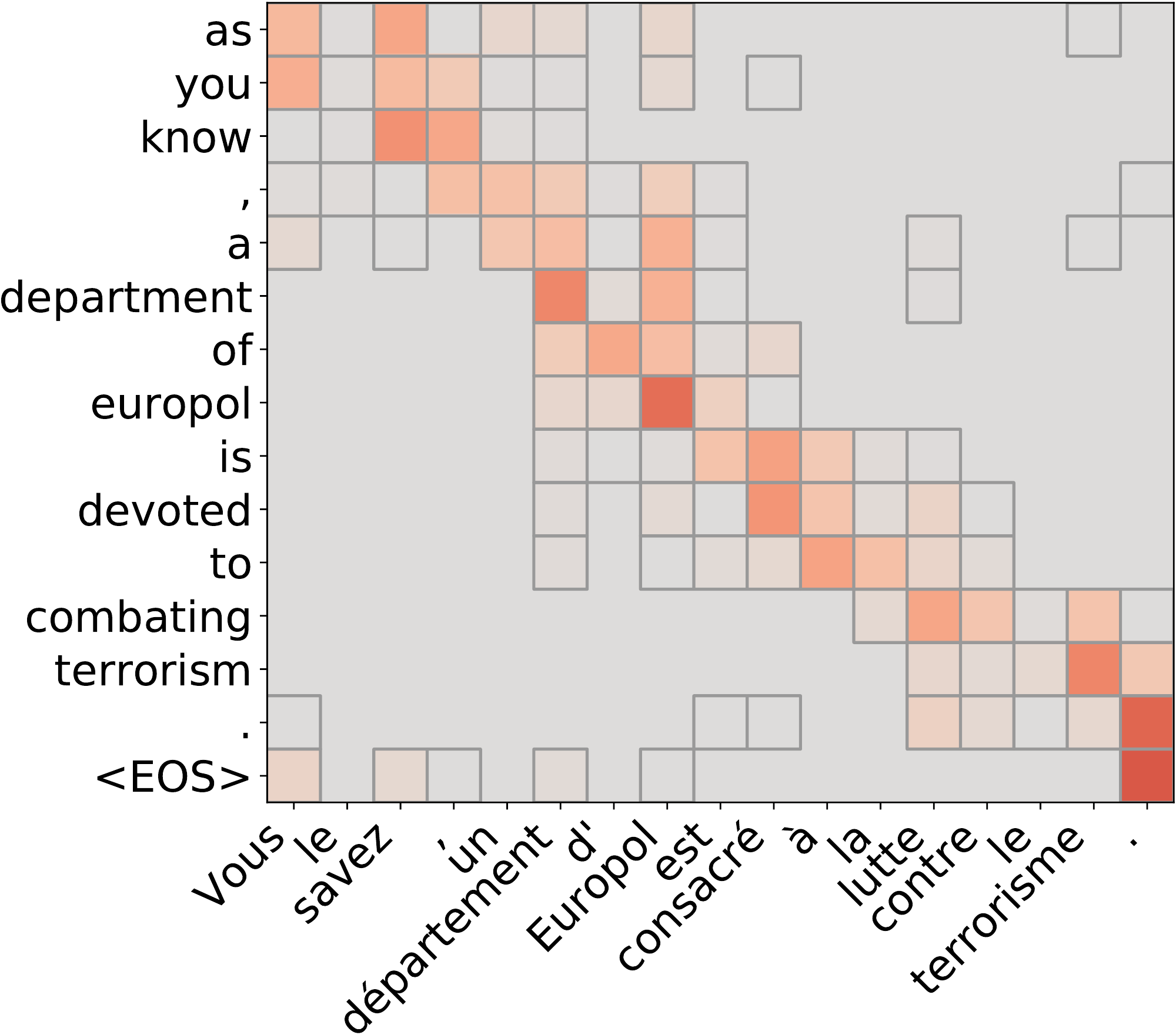}&
    \includegraphics[width=0.3\textwidth]{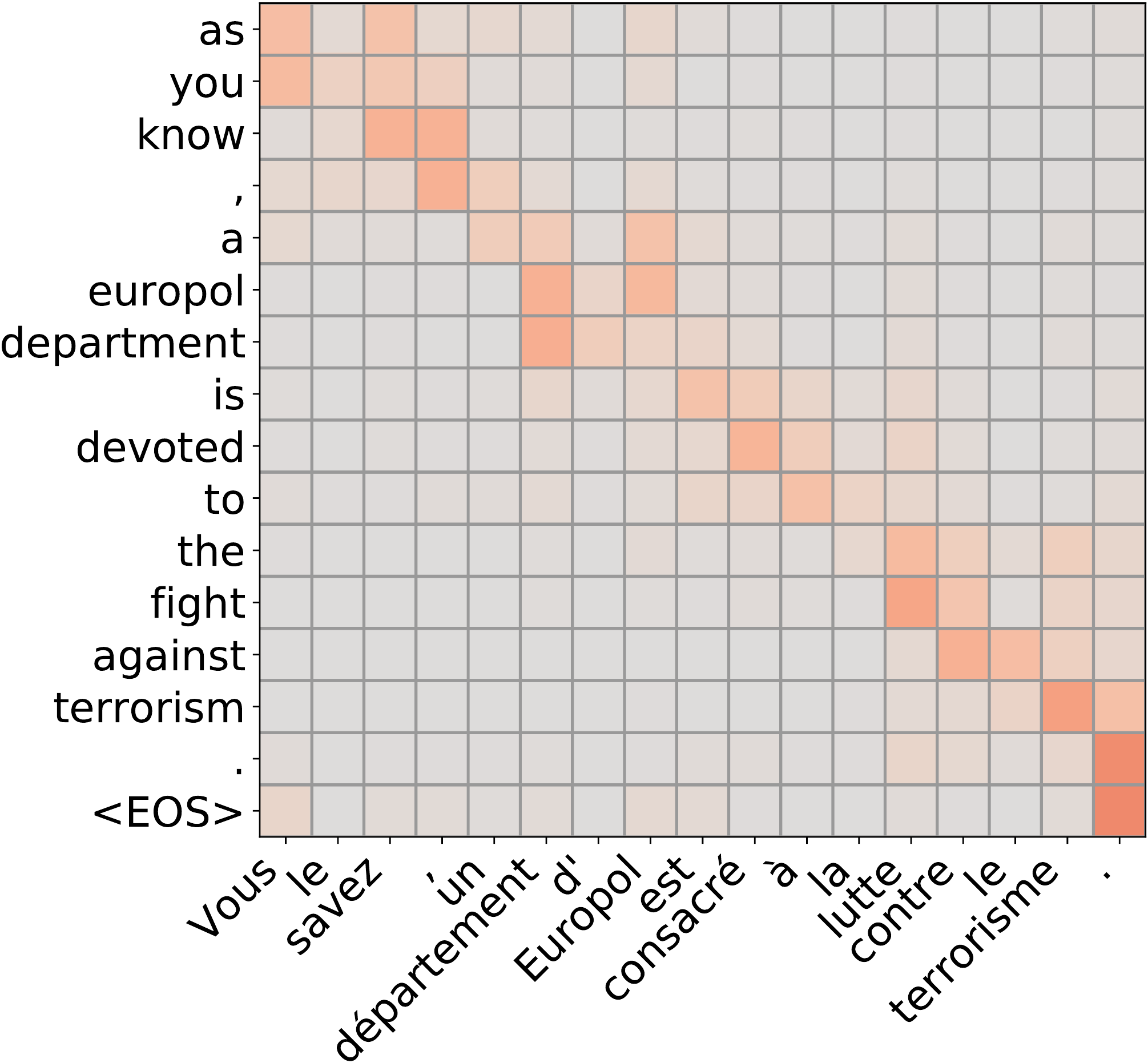}& 
    \includegraphics[width=0.3\textwidth]{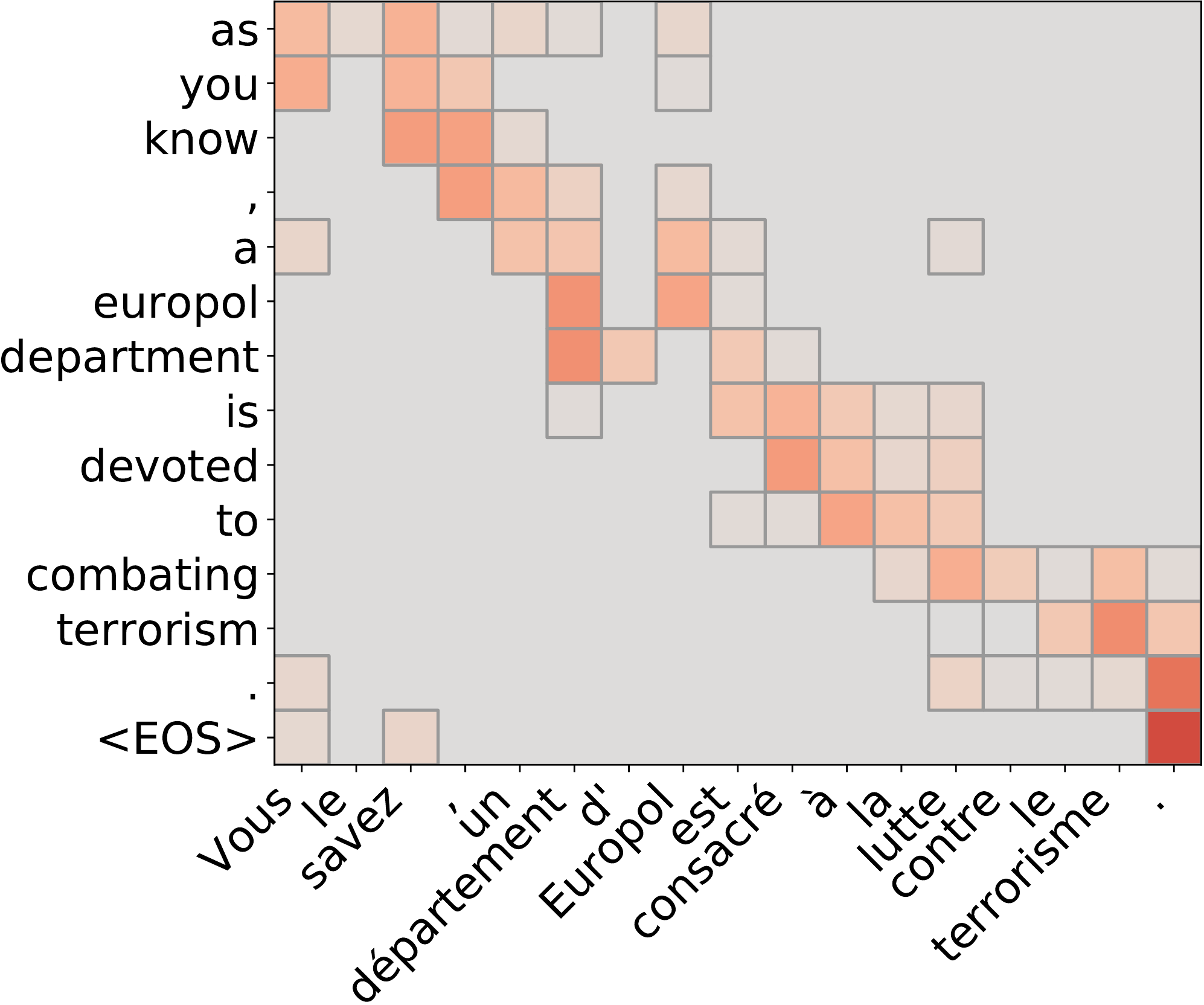} \\
    \\[0.7cm]
    \includegraphics[width=0.3\textwidth]{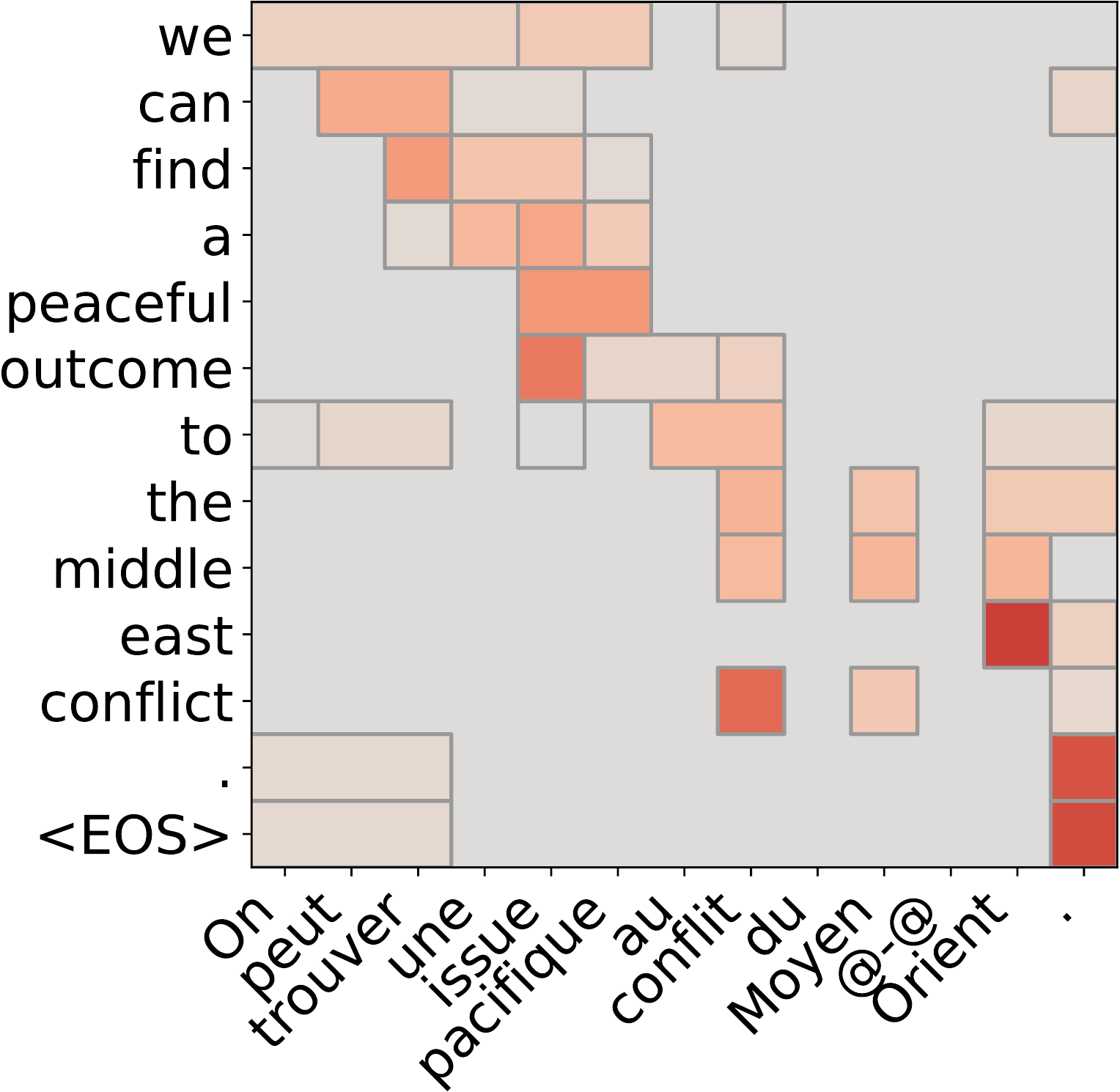} & 
    \includegraphics[width=0.3\textwidth]{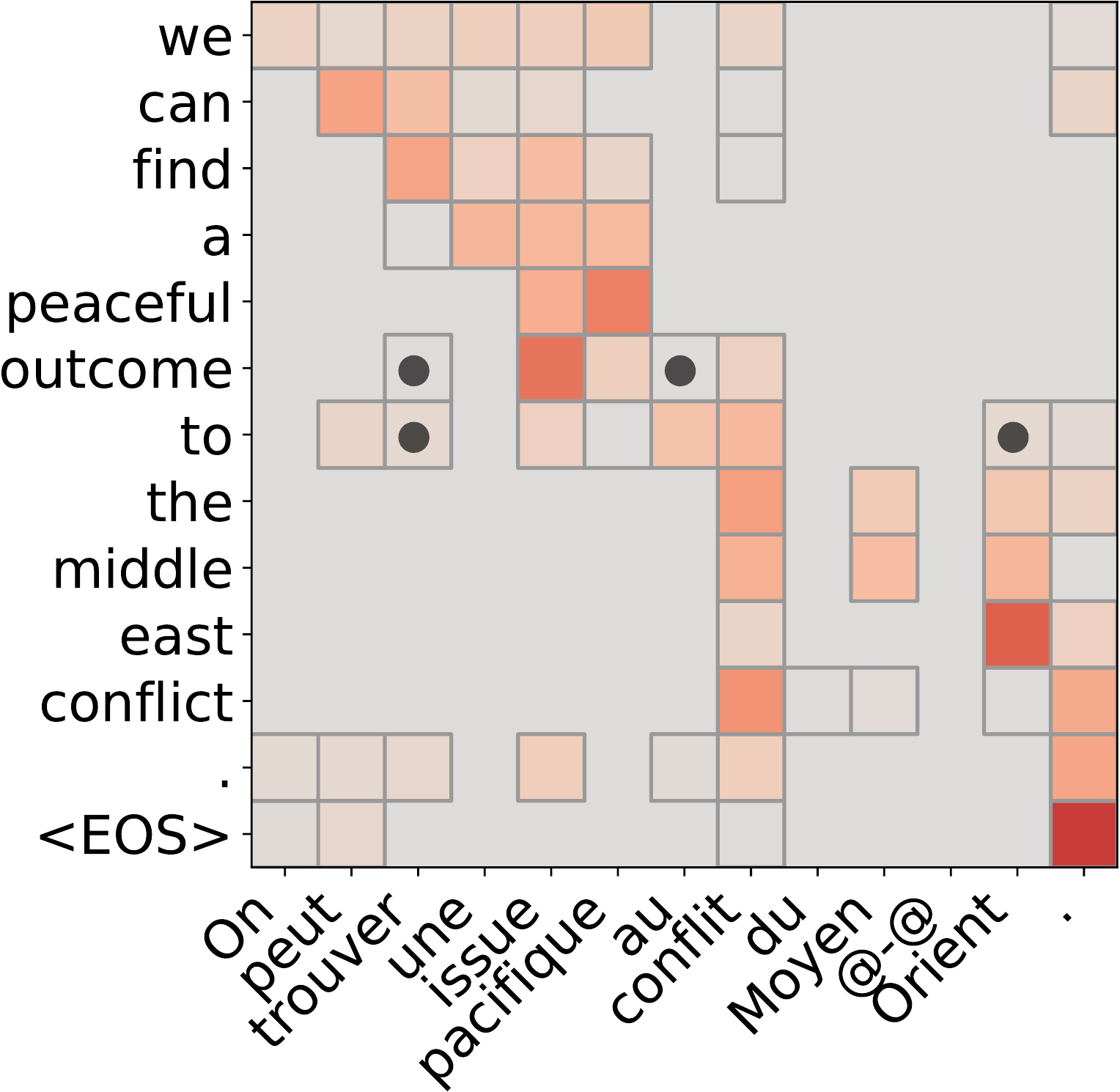} & 
    \includegraphics[width=0.3\textwidth]{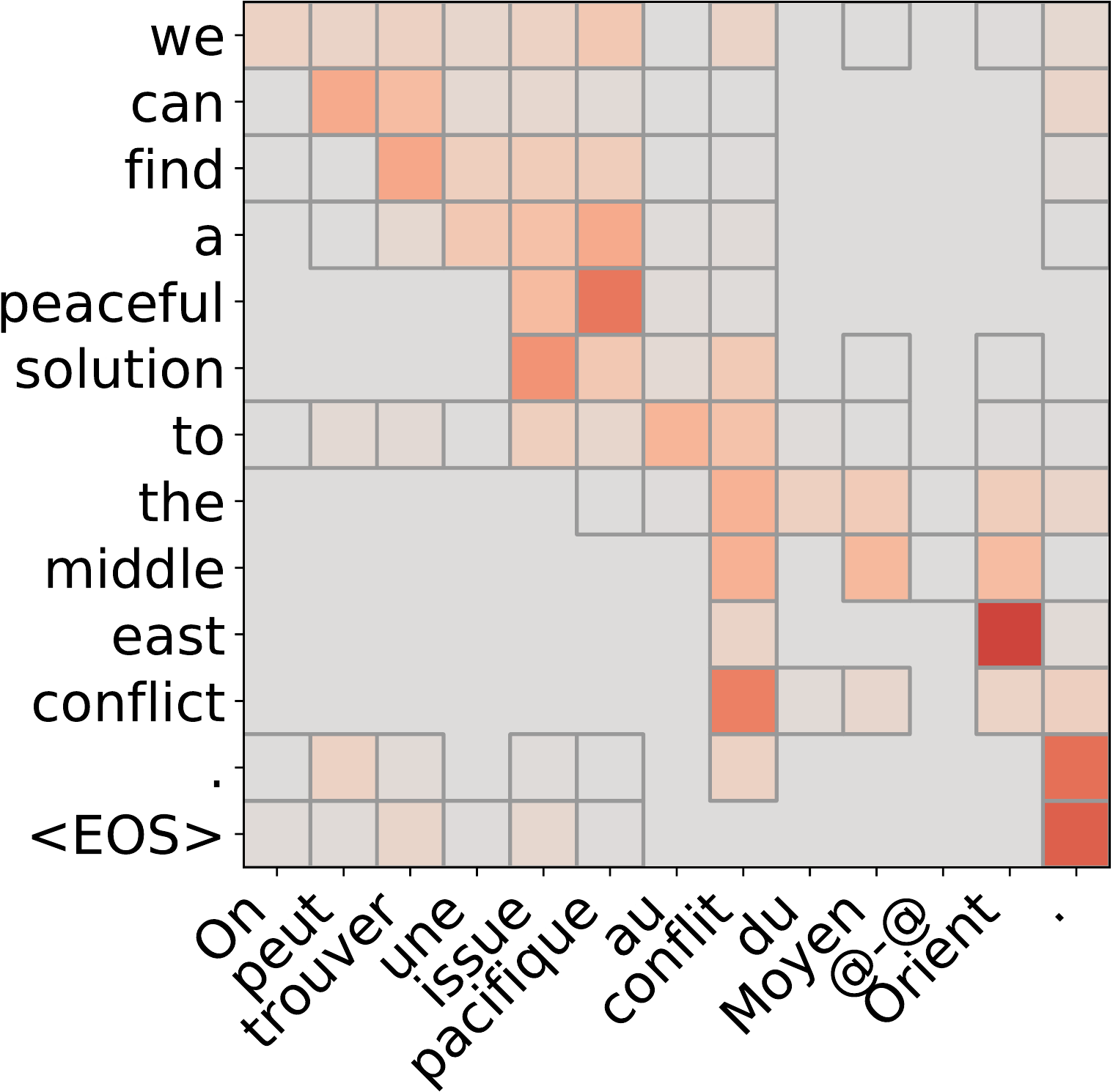} & 
    \includegraphics[width=0.3\textwidth]{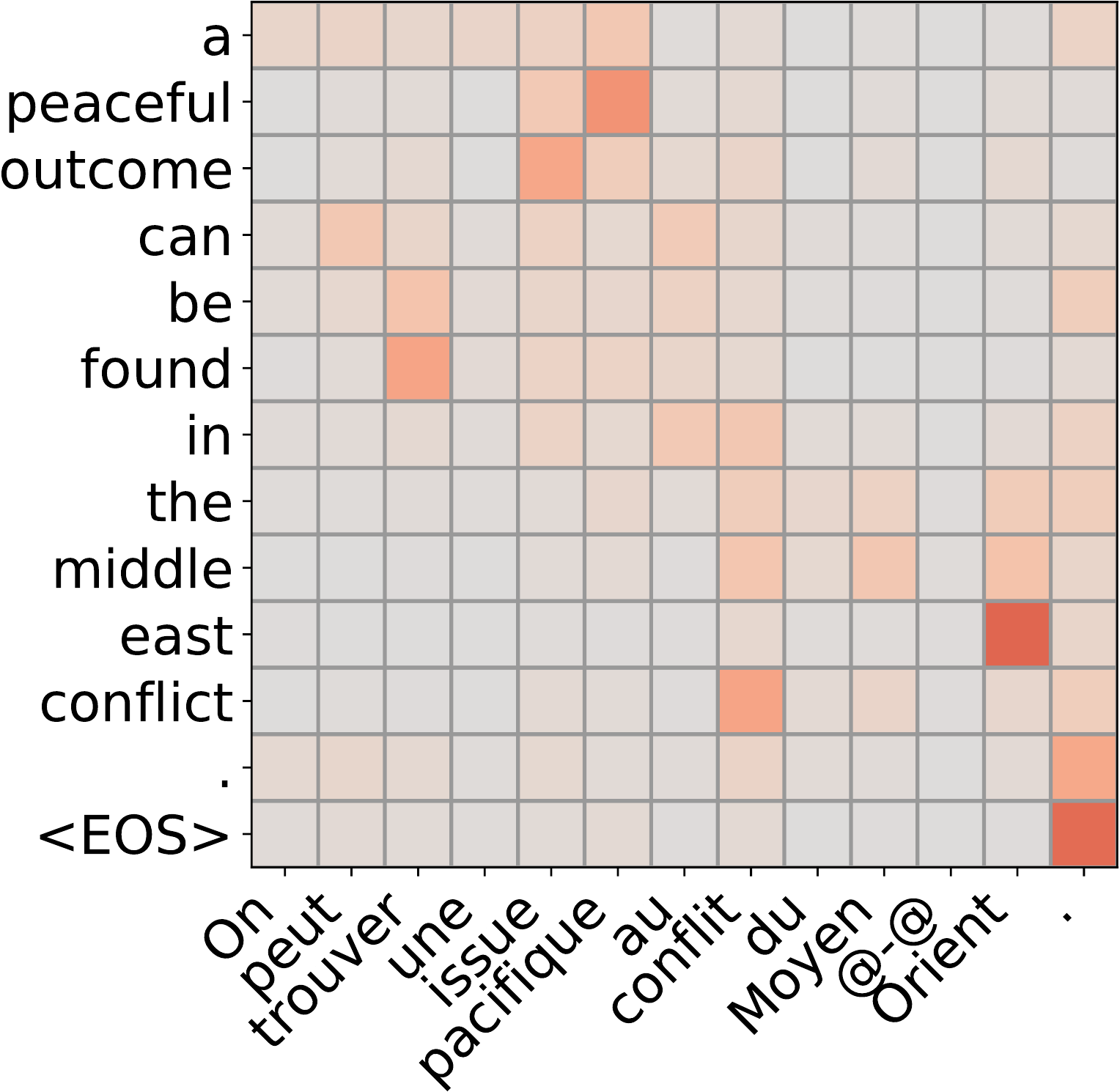} & 
    \includegraphics[width=0.3\textwidth]{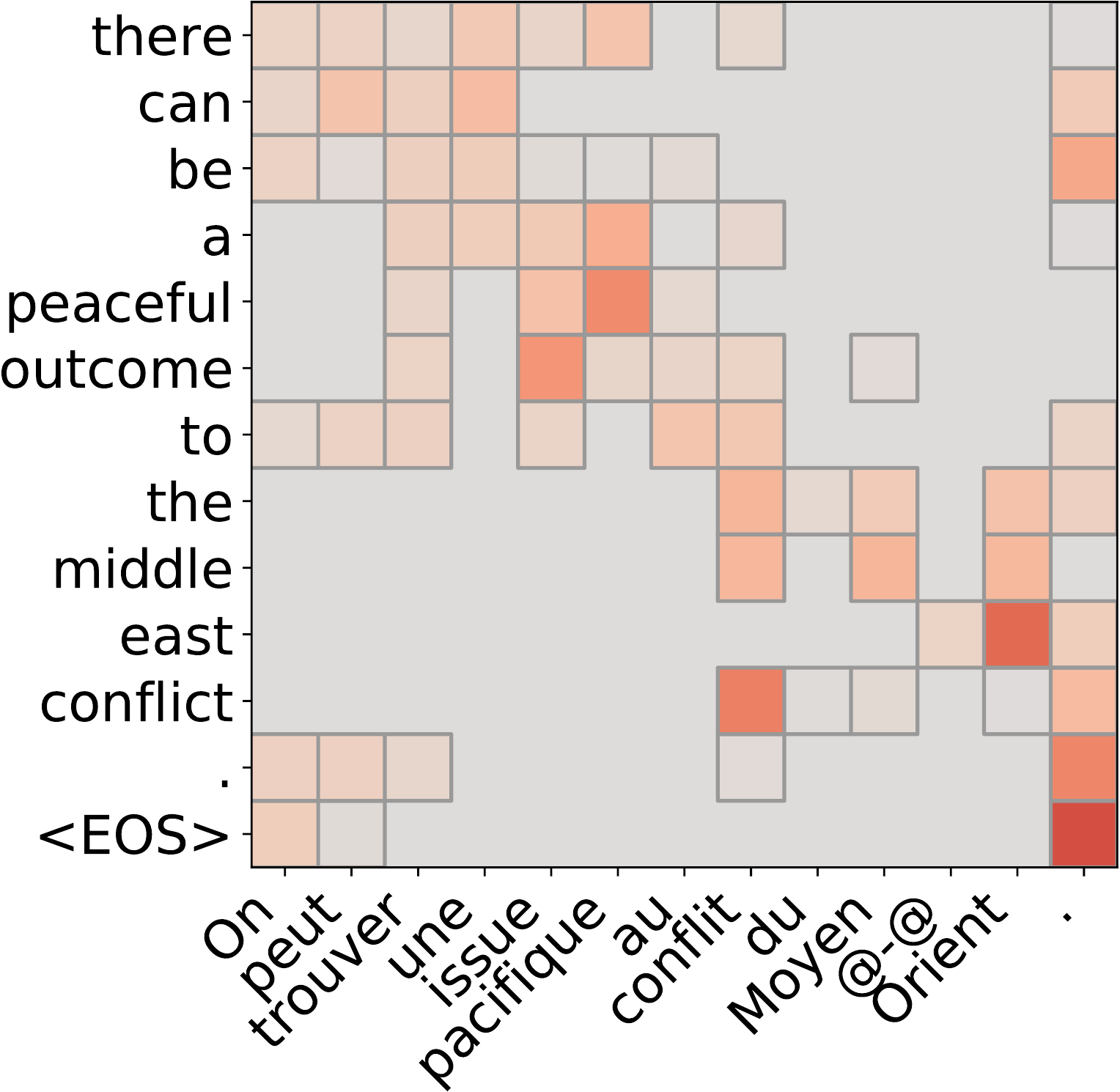}
    \end{tabular}
    \caption{Attention alignment examples for French-to-English translation,
        following the conventions of Figure~\ref{fig:intro_summ}.
        ``@-@" denotes a hyphen not separated by spaces.  When oscarmax induces
        multiple clusters, we denote them using different bullets (e.g.,
        $\bullet, \blacktriangle, \blacksquare$).
        Fusedmax often selects meaningful grammatical segments, such as ``est
        consacr\'e,'' as well as determiner-noun constructions.\label{fig:nmt_supp}}
\end{SidewaysFigure}
\begin{SidewaysContFigure}
    \captionsetup{list=off,format=cont}
    \begin{tabular}{c c c c c}
    fusedmax & oscarmax & sq-pnorm-max & softmax & sparsemax \\
    \midrule
    \includegraphics[width=0.3\textwidth]{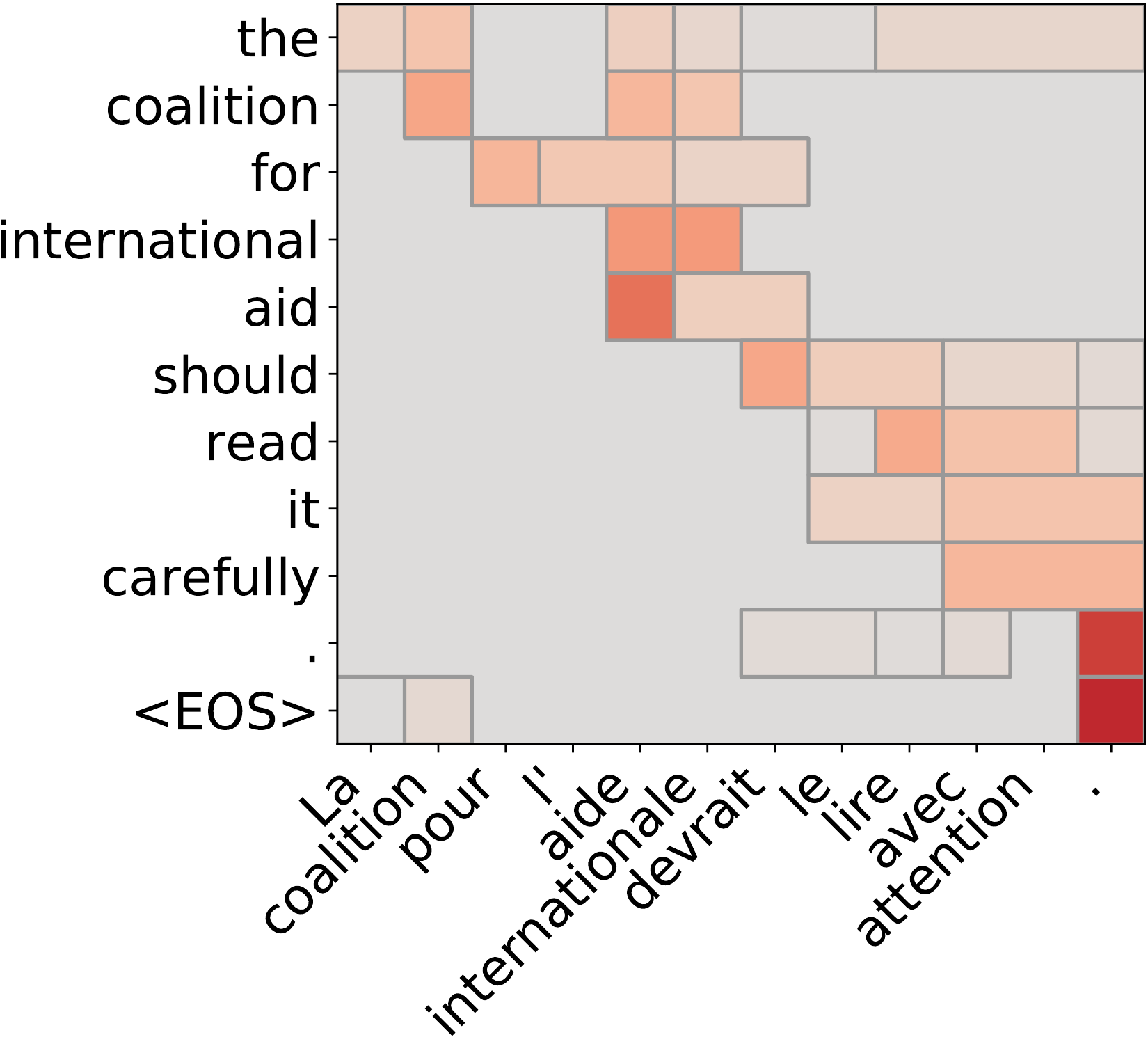} &
    \includegraphics[width=0.3\textwidth]{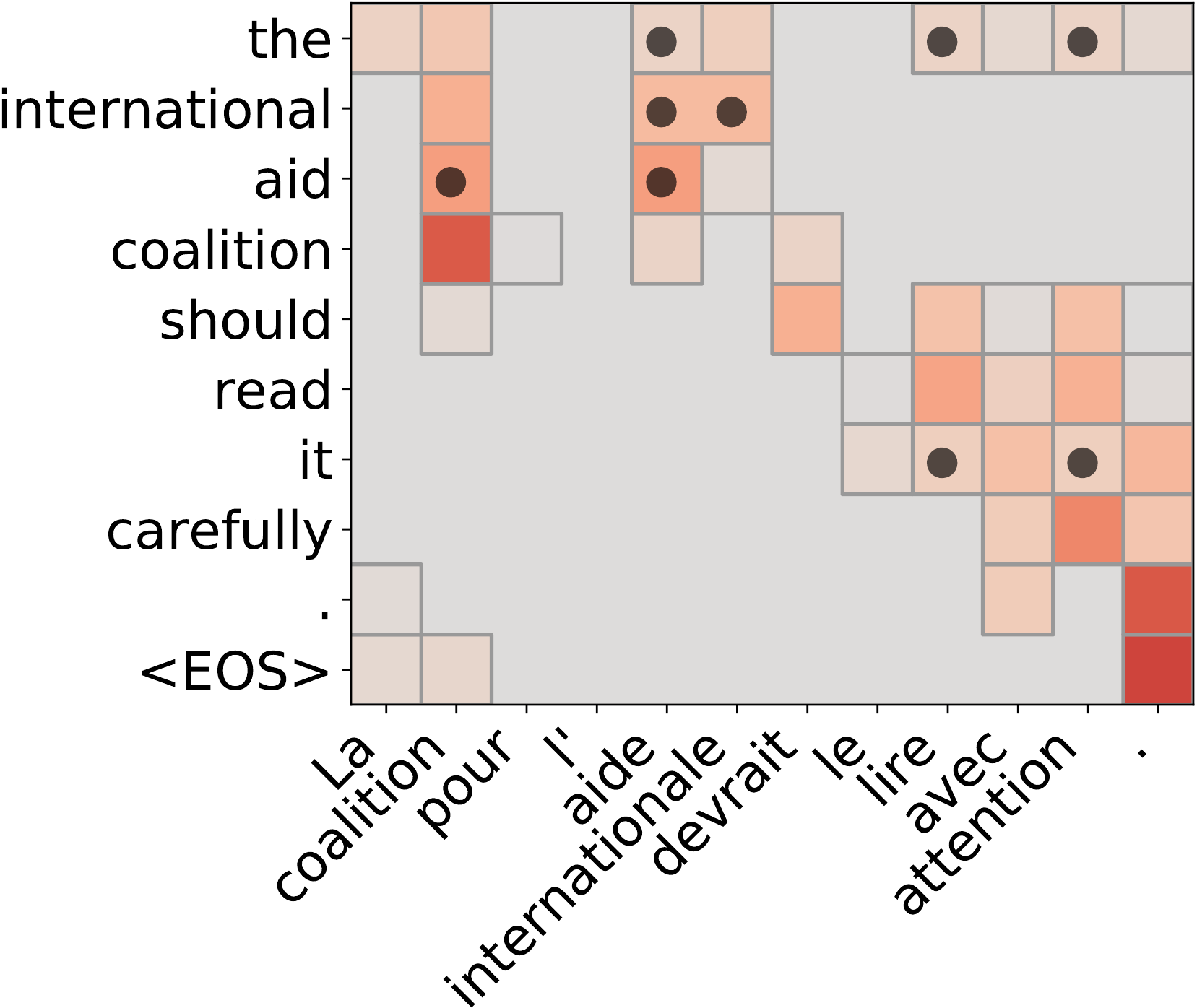} &
    \includegraphics[width=0.3\textwidth]{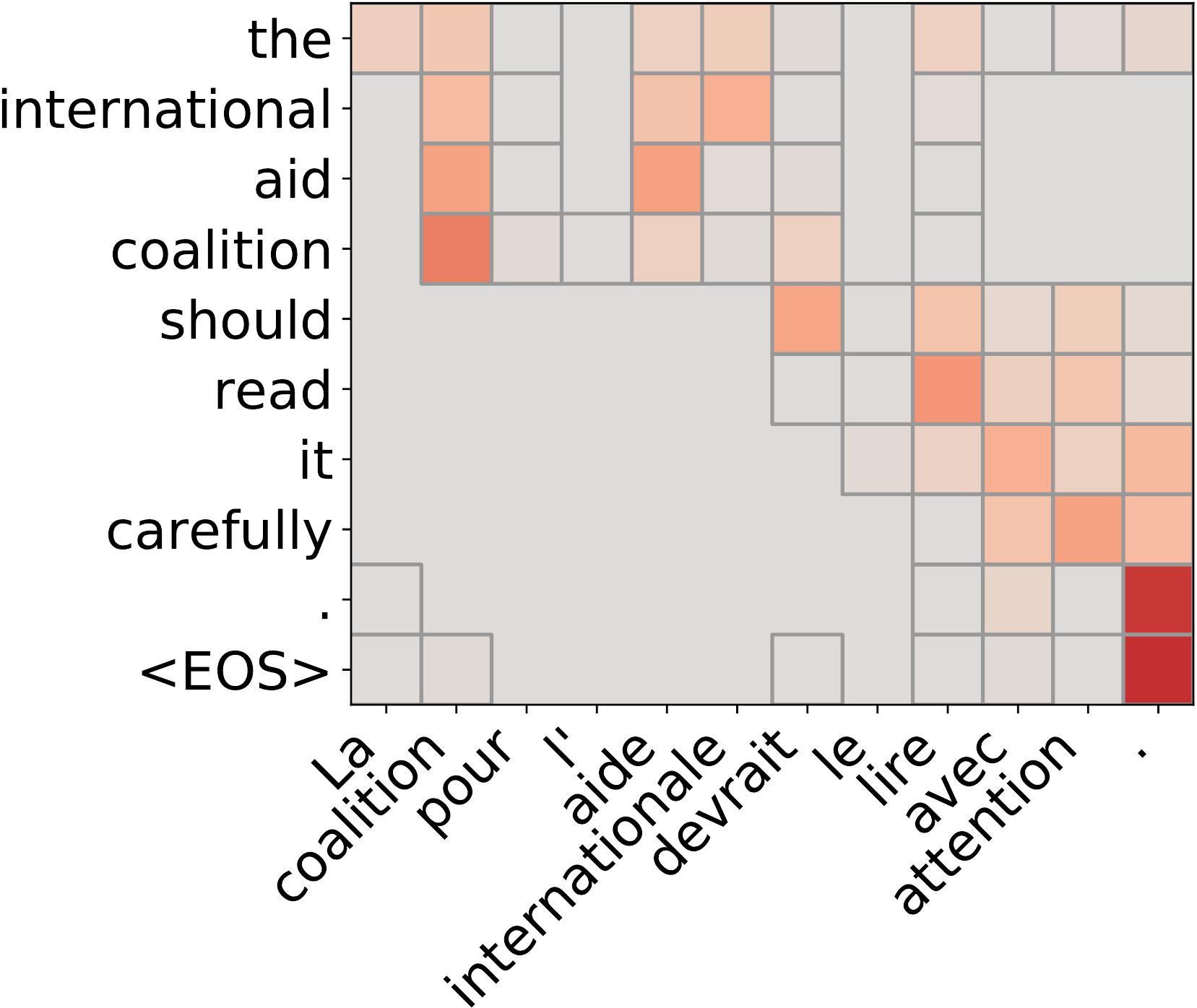} & 
    \includegraphics[width=0.3\textwidth]{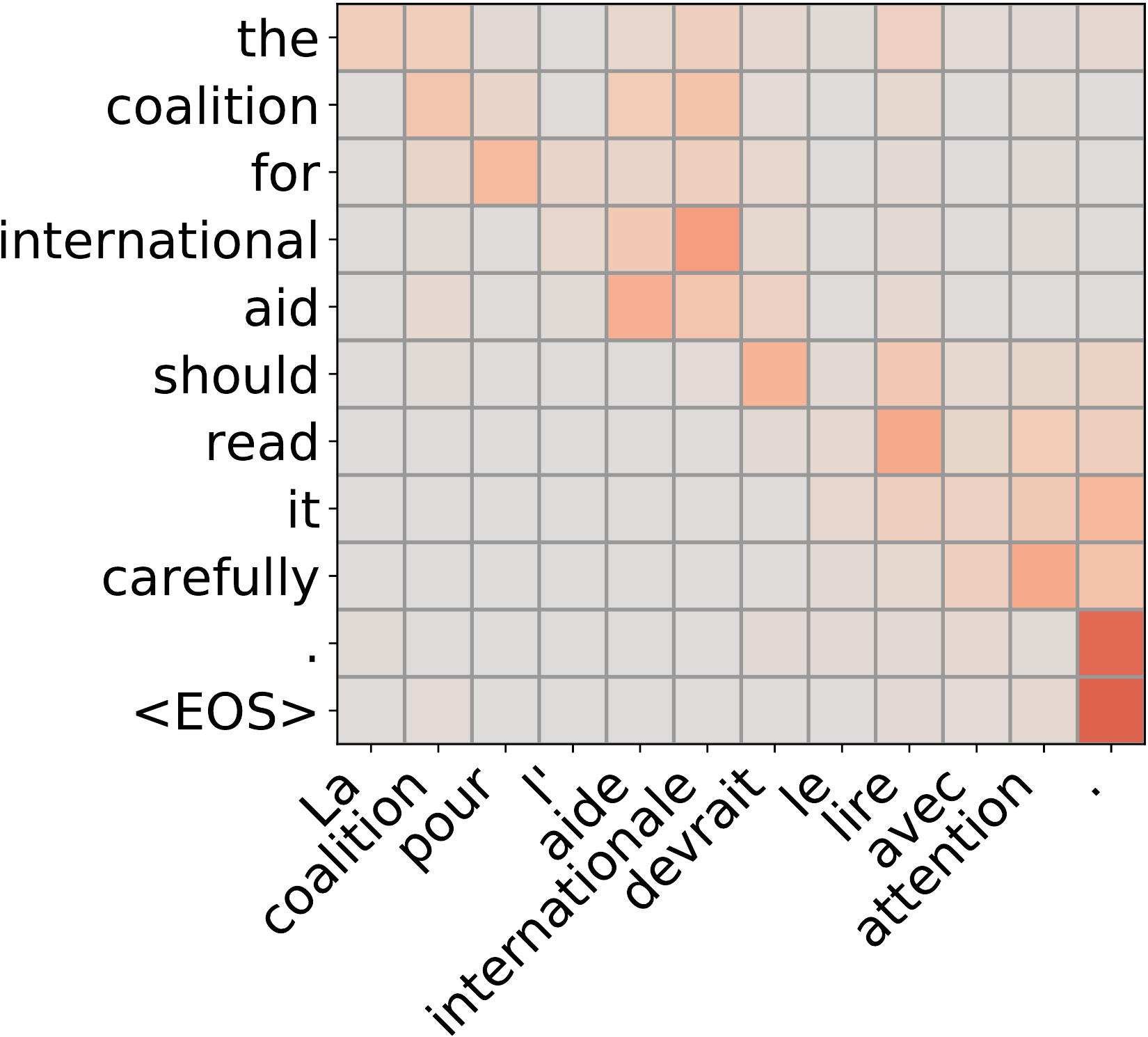} &
    \includegraphics[width=0.3\textwidth]{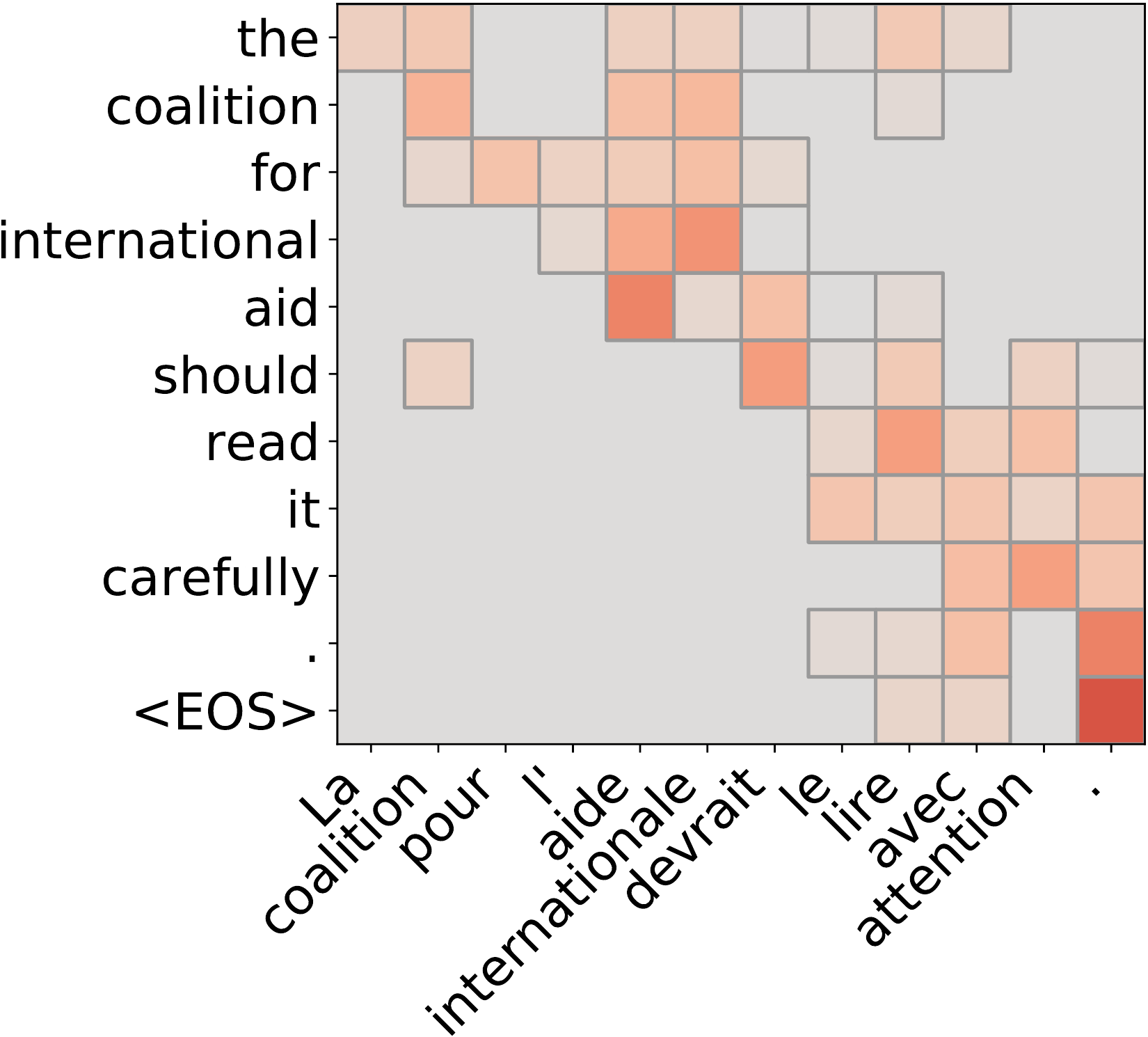}
    \\[0.7cm]
    \includegraphics[width=0.3\textwidth]{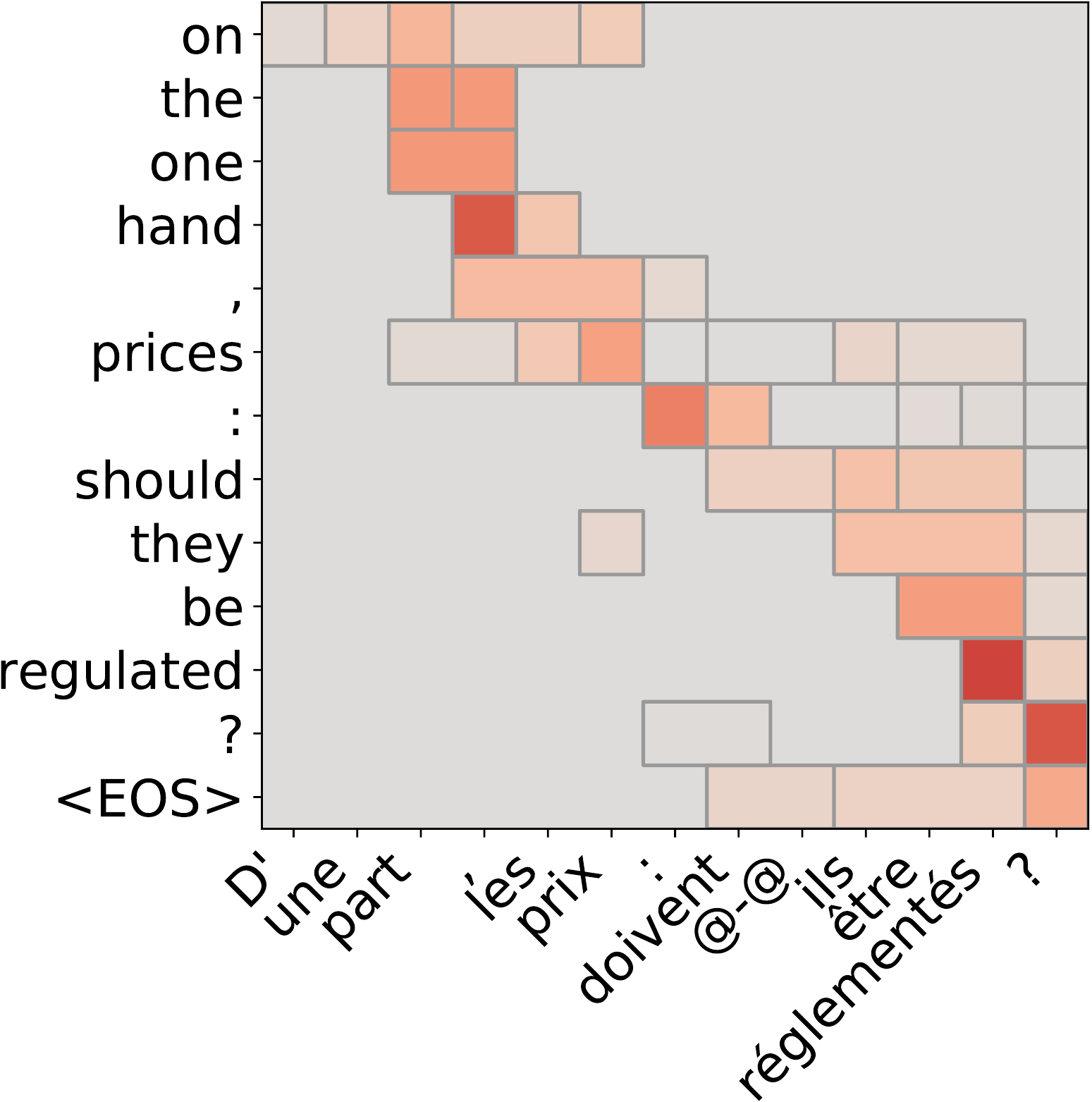}&
    \includegraphics[width=0.3\textwidth]{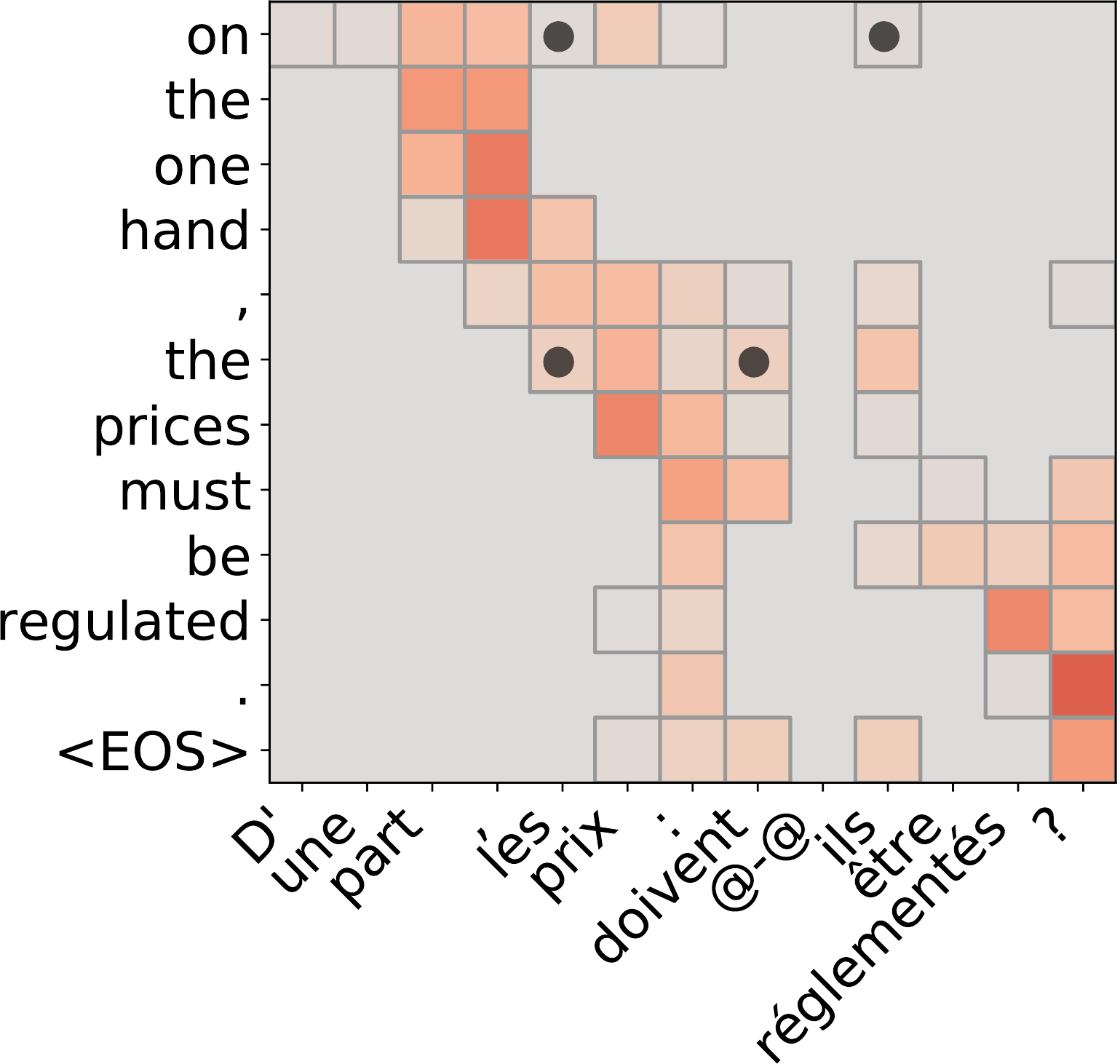}&
    \includegraphics[width=0.3\textwidth]{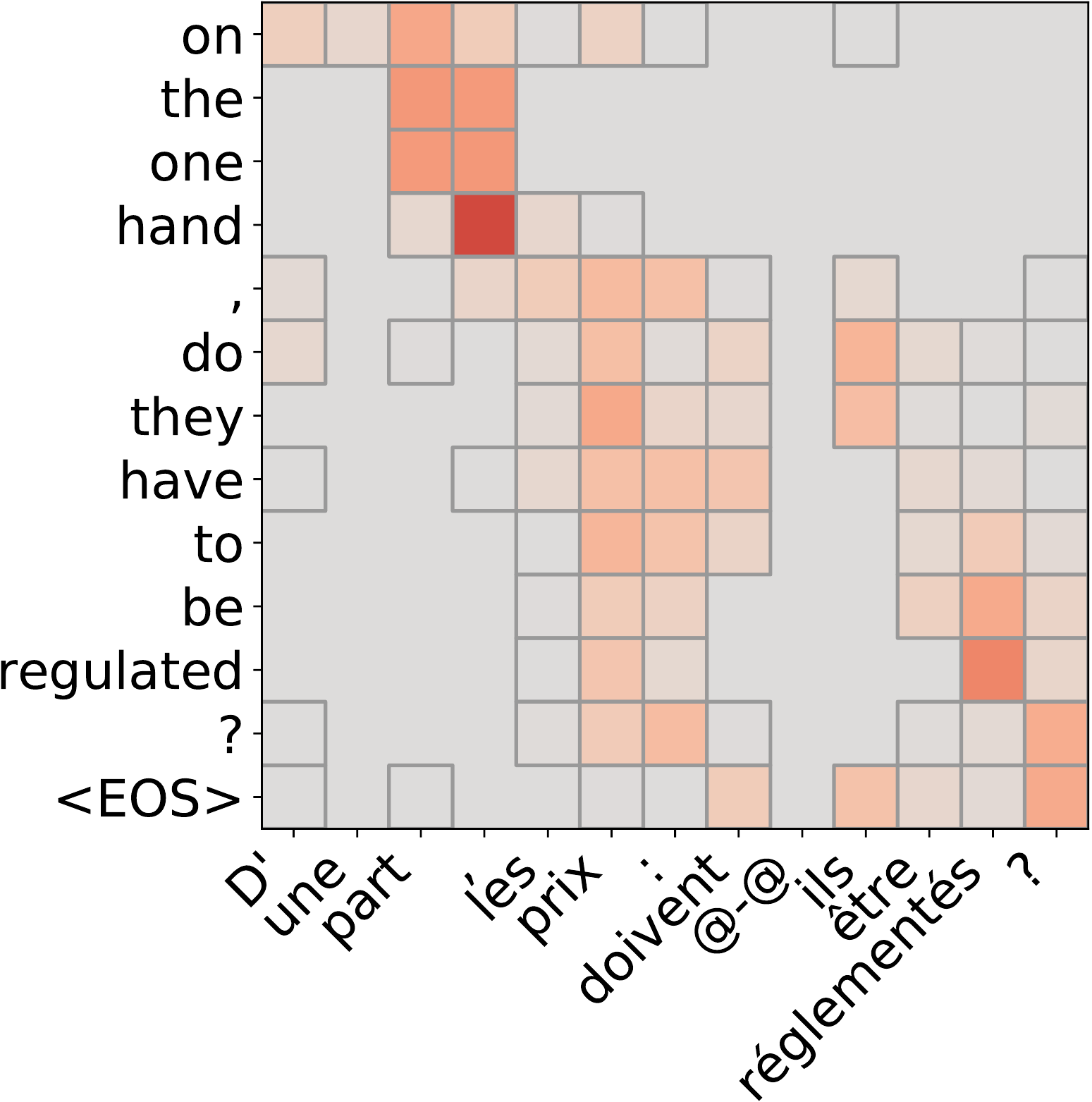}&
    \includegraphics[width=0.3\textwidth]{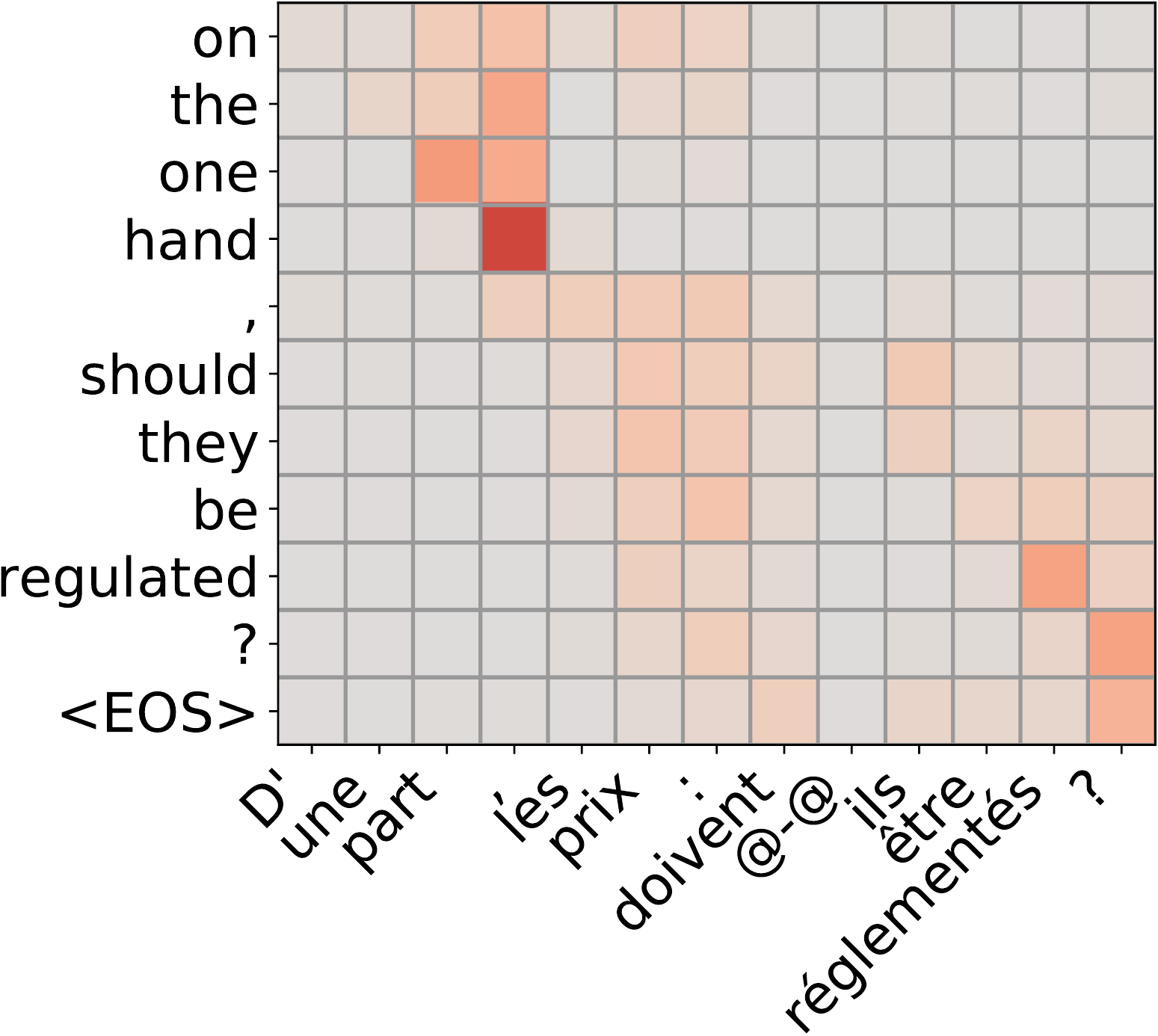}& 
    \includegraphics[width=0.3\textwidth]{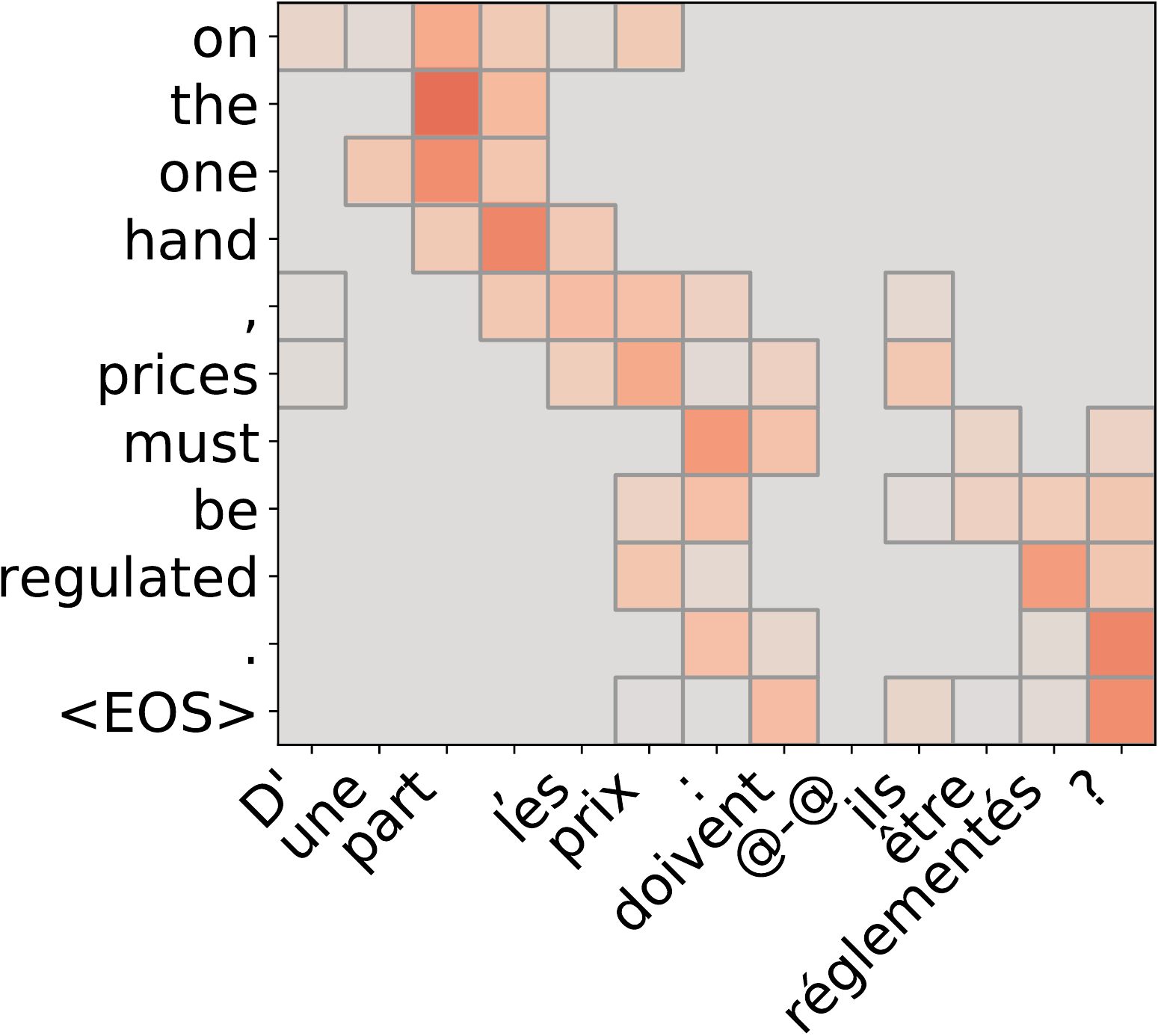} \\
    \\[0.4cm]
    \includegraphics[width=0.3\textwidth]{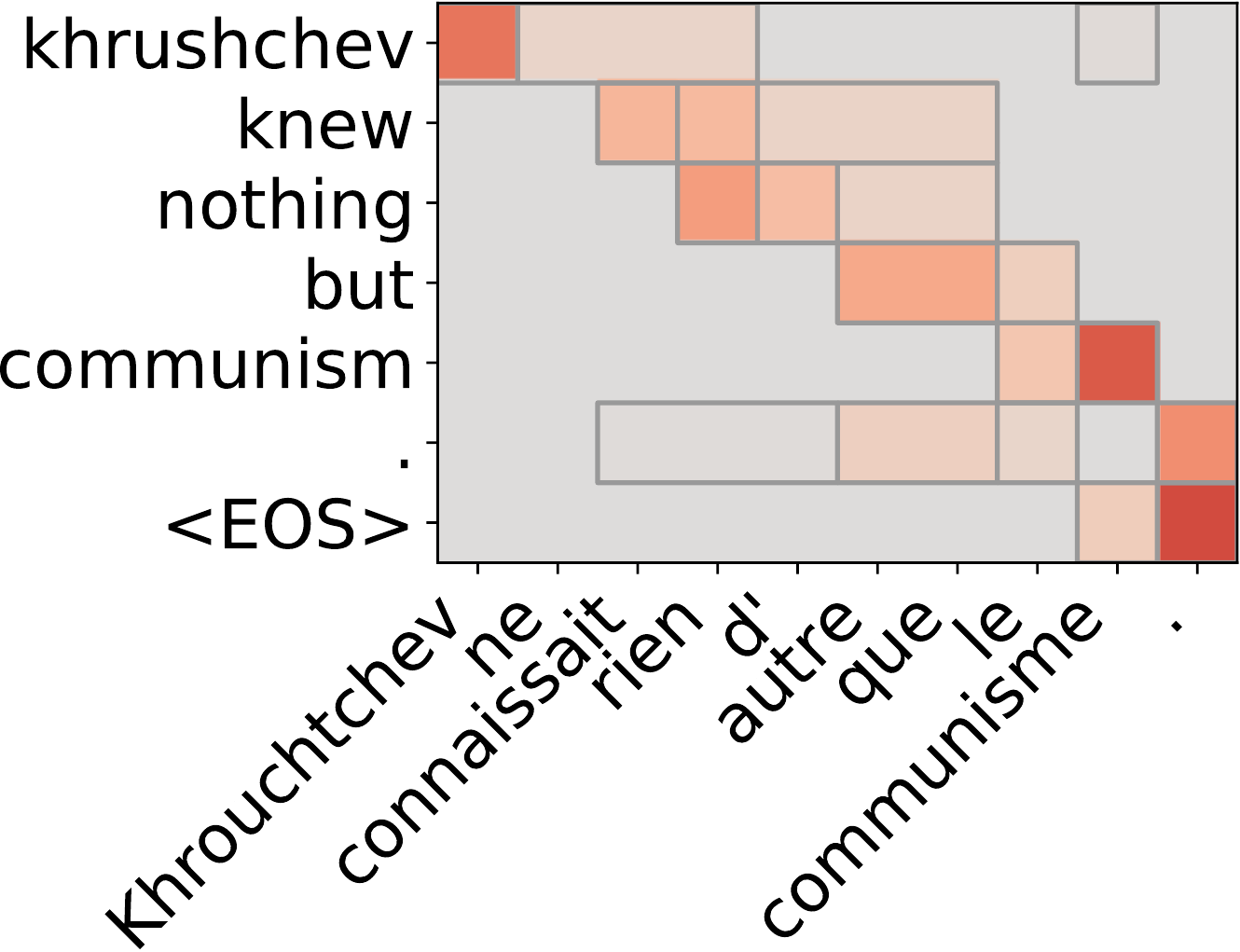} &
    \includegraphics[width=0.3\textwidth]{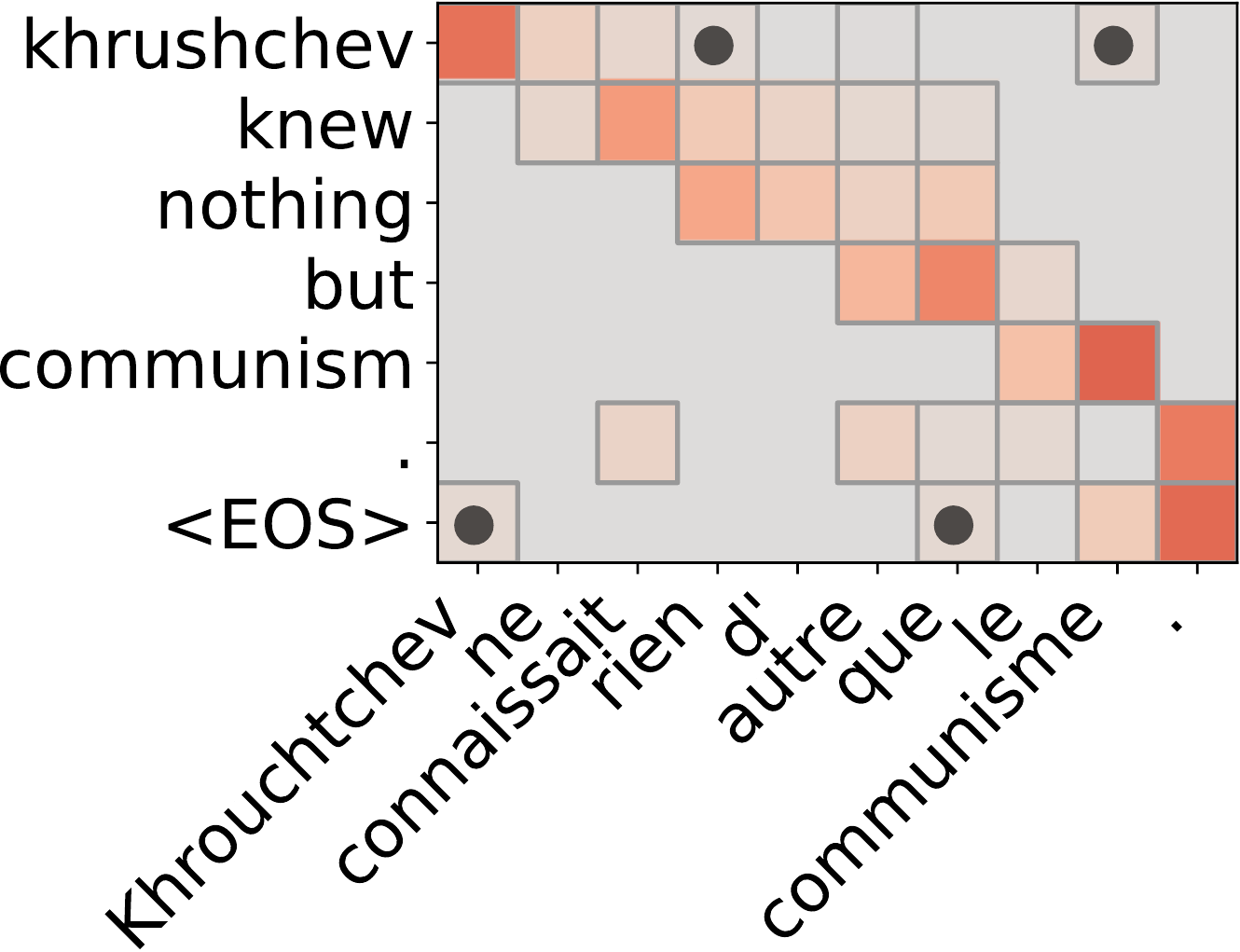} &
    \includegraphics[width=0.3\textwidth]{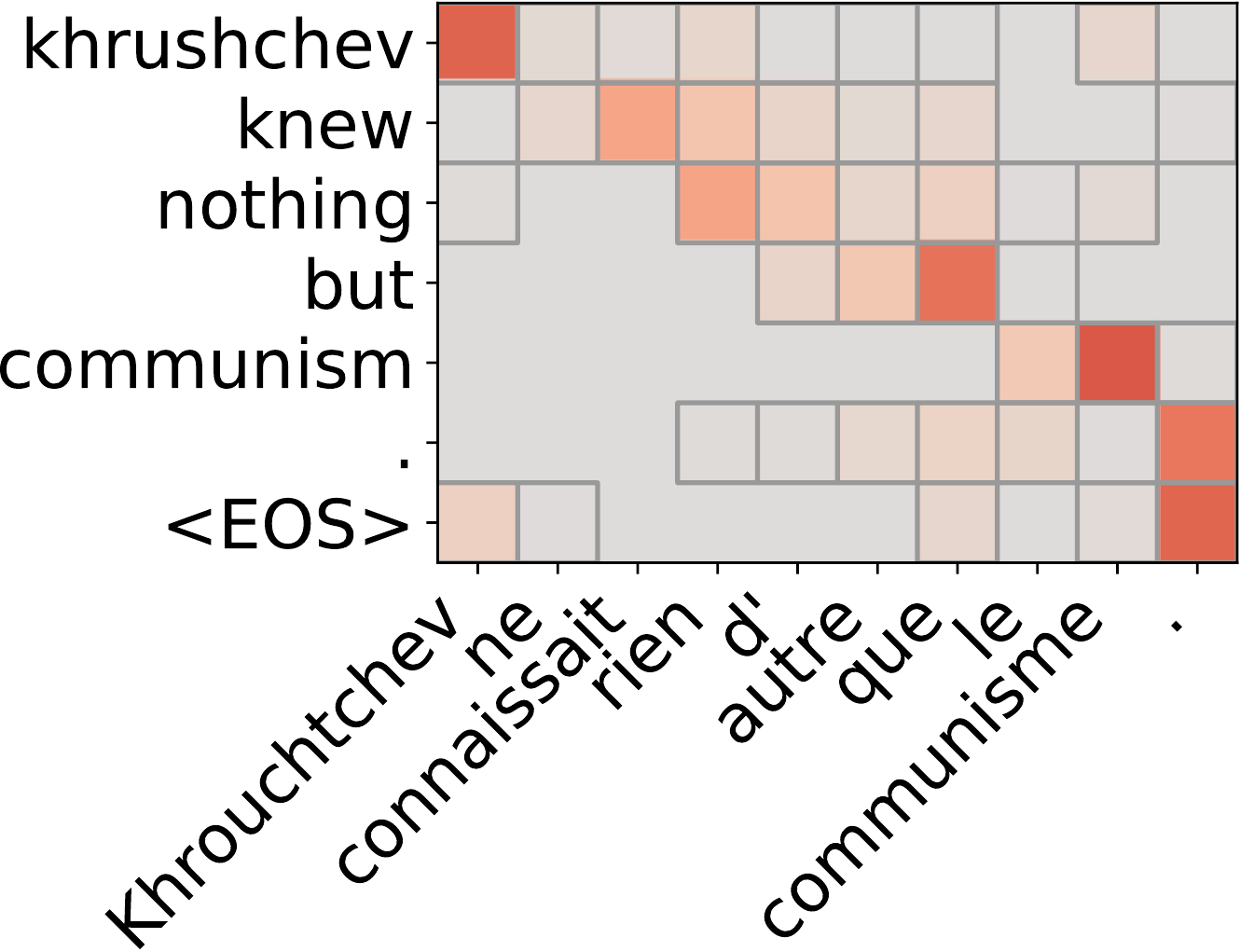} & 
    \includegraphics[width=0.3\textwidth]{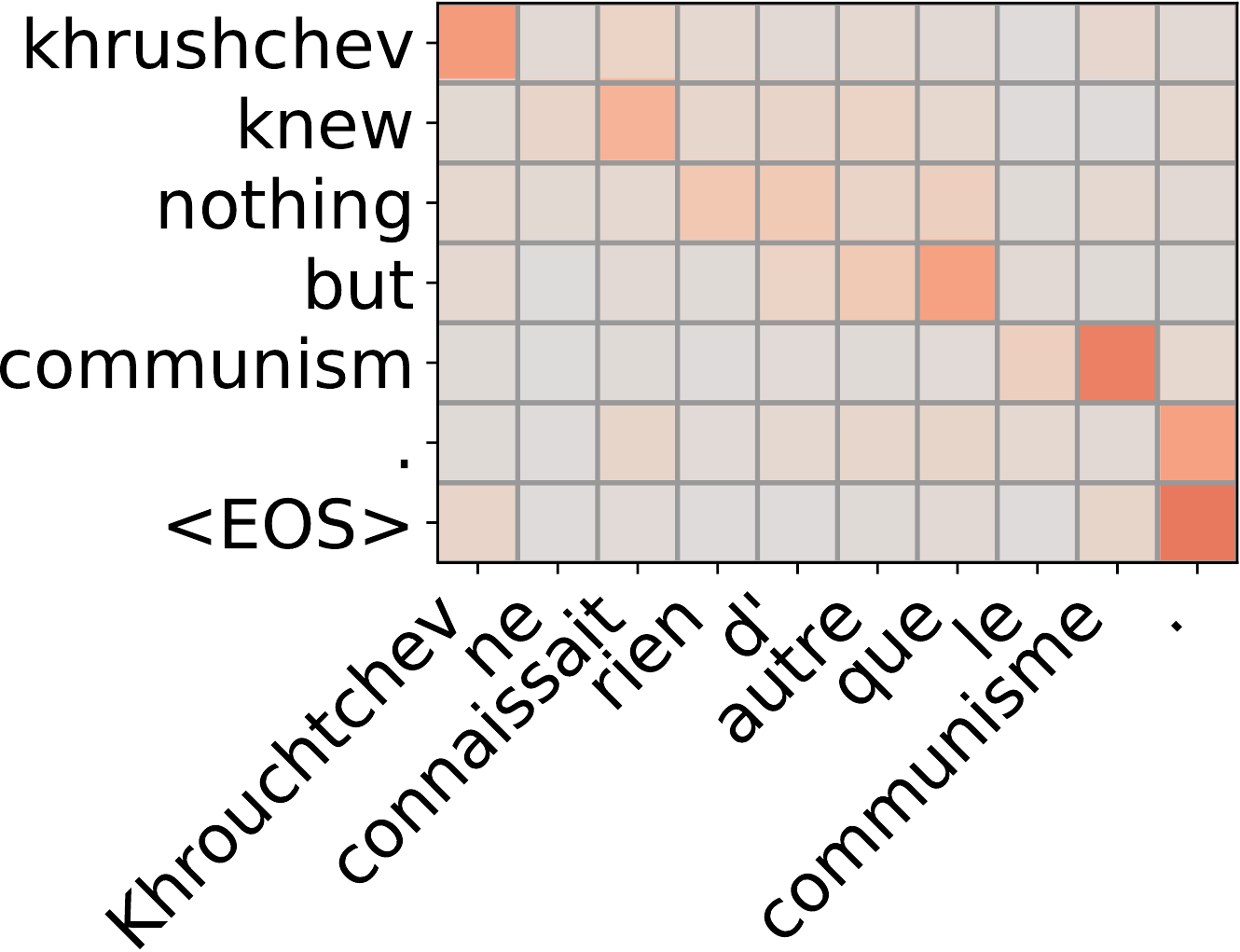} &
    \includegraphics[width=0.3\textwidth]{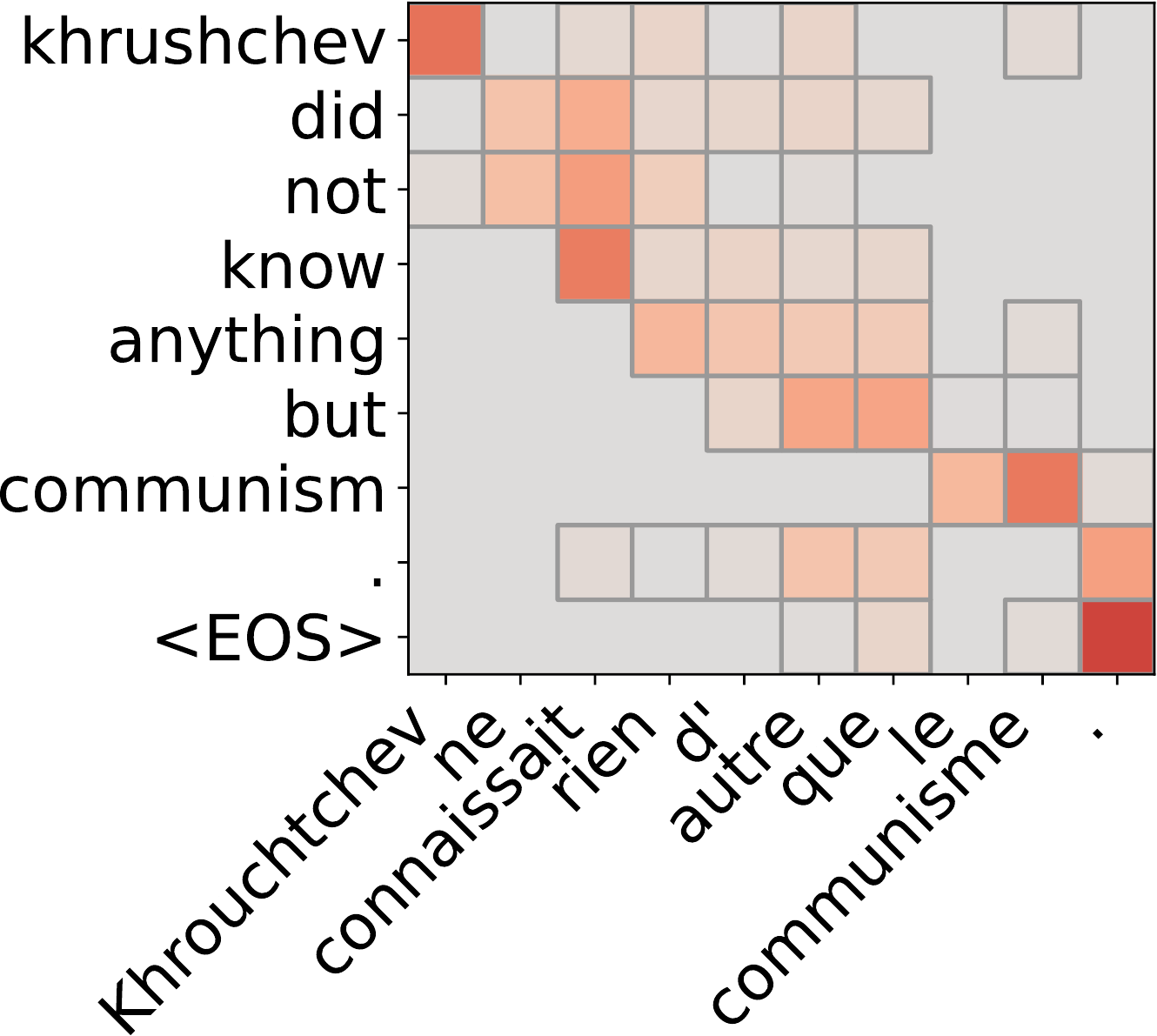}
    \end{tabular}
    \caption{Further translation examples from French to English.}
\end{SidewaysContFigure}

\clearpage
\subsection{Sentence summarization results}
\label{appendix:summarization}

{\bf Experimental setup and data.} We use the exact same experimental setup and
preprocessing as for machine translation, described in
Appendix~\ref{appendix:machine_translation}.  We use the preprocessed Gigaword
sentence summarization dataset, made available by the authors of
\cite{rush_summary} at \url{https://github.com/harvardnlp/sent-summary}. Since,
unlike \cite{rush_summary}, we do not perform any tuning on DUC-2003, we can 
report results on this dataset, as well. We observe that the simple
sequence-to-sequence model is able to keep summaries short without any explicit
constraints, informed only through training data statistics; therefore, in this
section, we also report results without output truncation at 75 bytes
(Table~\ref{table:summarization_detailed}). We also provide precision and recall
scores for {\footnotesize ROUGE-L} in Table~\ref{table:summarization_prf}.
Finally, we provide attention weights plots for all studied attention mechanisms
and a number of validation set examples in Figure~\ref{fig:supp_summ}.

\begin{table}[h]
    \caption{Sentence summarization $F_1$ scores for several {\footnotesize
    ROUGE} variations.\label{table:summarization_detailed}}
    \centering \small

\begin{tabular}{r r r r r r r r r}
    \toprule
                  & \multicolumn{4}{c}{Truncated} & \multicolumn{4}{c}{Not
truncated}\\
        \cmidrule(lr){2-5} \cmidrule(lr){6-9} 
        attention &
        \makebox[\widthof{\scriptsize xUGE-W$_{1.2}$}][r]{\scriptsize ROUGE-1}&
        \makebox[\widthof{\scriptsize xUGE-W$_{1.2}$}][r]{\scriptsize ROUGE-2}&
        \makebox[\widthof{\scriptsize xUGE-W$_{1.2}$}][r]{\scriptsize ROUGE-L}&
        {\scriptsize ROUGE-W$_{1.2}$} &
        \makebox[\widthof{\scriptsize xUGE-W$_{1.2}$}][r]{\scriptsize ROUGE-1}&
        \makebox[\widthof{\scriptsize xUGE-W$_{1.2}$}][r]{\scriptsize ROUGE-2}&
        \makebox[\widthof{\scriptsize xUGE-W$_{1.2}$}][r]{\scriptsize ROUGE-L}&
        {\scriptsize ROUGE-W$_{1.2}$} \\ 
        \midrule
    
\multicolumn{9}{l}{\bf DUC 2003} \\
softmax    & 26.63 & 8.72 & 23.87 & 16.95 & 27.06 & 8.86 & 24.23 & 17.02 \\
sparsemax  & 26.54 & 8.78 & 23.89 & 16.93 & 26.95 & 8.94 & 24.21 & 16.99 \\
    \spacerule
fusedmax   & {\bf 27.12} & 8.93 & {\bf 24.39} & {\bf 17.28} & {\bf 27.48} & 9.04
           & {\bf 24.66} & {\bf 17.30} \\
oscarmax   & 26.72 & {\bf 9.08} & 24.02 & 17.06 & 27.11 & {\bf 9.23} & 24.32 & 17.10 \\
sq-pnorm-max & 26.55 & 8.77 & 23.78 & 16.87 & 26.92 & 8.89 & 24.07 & 16.92 \\

\addlinespace[0.6em]
\multicolumn{9}{l}{\bf DUC 2004} \\

softmax    & 27.16 & 9.48 & 24.47 & 17.14 & 27.25 & 9.52 & 24.55 & 17.20 \\
sparsemax  & 27.69 & 9.55 & 24.96 & 17.44 & 27.77 & 9.61 & 25.02 & 17.48 \\
    \spacerule
fusedmax   & {\bf 28.42} & {\bf 9.96} & {\bf 25.55} & {\bf 17.78} & {\bf 28.43}
           & {\bf 9.96} & {\bf 25.55} & {\bf 17.79} \\
oscarmax   & 27.84 & 9.46 & 25.14 & 17.55 & 27.88 & 9.47 & 25.17 & 17.57 \\
sq-pnorm-max      & 27.94 & 9.28 & 25.08 & 17.49 & 28.01 & 9.30 & 25.13 & 17.52 \\

\addlinespace[0.6em]
\multicolumn{9}{l}{\bf Gigaword} \\

softmax    & 35.13 & 17.15 & 32.92 & 24.17 & 35.01 & 17.10 & 32.77 & 24.00 \\
sparsemax  & 36.04 & {\bf 17.78} & 33.64 & 24.69 & 35.97 & {\bf 17.75} &
33.54 & {\bf 24.55} \\
    \spacerule
fusedmax   & {\bf 36.09} & 17.62 & {\bf 33.69} & 24.69 & {\bf 35.98} &
17.60 & {\bf 33.59} & 24.54 \\
oscarmax   & 35.36 & 17.23 & 33.03 & 24.25 & 35.26 & 17.20 & 32.92 & 24.10 \\
    sq-pnorm-max & 35.94 & 17.75 & 33.66 & {\bf 24.71} & 35.86 & 17.73 & 33.54 & {\bf
24.55} \\
\bottomrule
\end{tabular}
\end{table}

\begin{table}[h]

    \caption{Sentence summarization: ROUGE-L 
    precision, recall and F-scores.\label{table:summarization_prf}}
    \centering \small

    \begin{tabular}{r@{\quad}r@{~~~}r@{~~~}r@{\quad}r@{~~~}r@{~~~}r}
    \toprule
          & \multicolumn{3}{c}{Truncated} & \multicolumn{3}{c}{Not
truncated}\\
        \cmidrule(lr){2-4} \cmidrule(lr){5-7} 
        attention & $P$ & $R$ & $F_1$ & $P$ & $R$ & $F_1$ \\
        \midrule
    
\multicolumn{7}{l}{\bf DUC 2003} \\

softmax      & 29.57 & 20.67 & 23.87 & 30.40 & 20.80 & 24.23 \\
sparsemax    & 29.59 & 20.58 & 23.89 & 30.37 & 20.68 & 24.21 \\
fusedmax     & {\bf 30.02} & {\bf 21.11} & {\bf 24.39} & {\bf 30.75} & {\bf 21.15} & {\bf 24.66} \\
oscarmax     & 29.64 & 20.78 & 24.02 & 30.40 & 20.87 & 24.32 \\
sq-pnorm-max & 29.45 & 20.50 & 23.78 & 30.23 & 20.56 & 24.07 \\

\addlinespace[0.6em]
\multicolumn{7}{l}{\bf DUC 2004} \\

softmax      & 30.54 & 21.00 & 24.47 & 30.59 & 21.13 & 24.55 \\
sparsemax    & 30.99 & 21.57 & 24.96 & 31.03 & 21.64 & 25.02 \\
fusedmax     & {\bf 32.19} & {\bf 21.80} & {\bf 25.55} & {\bf 32.19} & {\bf 21.81}
             & {\bf 25.55} \\
oscarmax     & 31.89 & 21.46 & 25.14 & 31.91 & 21.51 & 25.17 \\
sq-pnorm-max & 31.42 & 21.55 & 25.08 & 31.46 & 21.63 & 25.13 \\

\addlinespace[0.6em]
\multicolumn{7}{l}{\bf Gigaword} \\

softmax      & 36.43 & 31.67 & 32.92 & 36.61 & 31.54 & 32.77 \\
sparsemax    & 37.32 & 32.18 & 33.64 & 37.54 & 32.07 & 33.54 \\
fusedmax     & {\bf 37.44} & 32.15 & {\bf 33.69} & {\bf 37.68} & 32.01 & {\bf
33.59} \\
oscarmax     & 36.40 & 31.78 & 33.03 & 36.61 & 31.67 & 32.92 \\
sq-pnorm-max & 37.12 & {\bf 32.37} & 33.66 & 37.31 & {\bf 32.26} & 33.54 \\

\bottomrule
\end{tabular}
\end{table}
\clearpage  
\begin{figure}[p]
    \centering
    \includegraphics[width=0.43\textwidth]{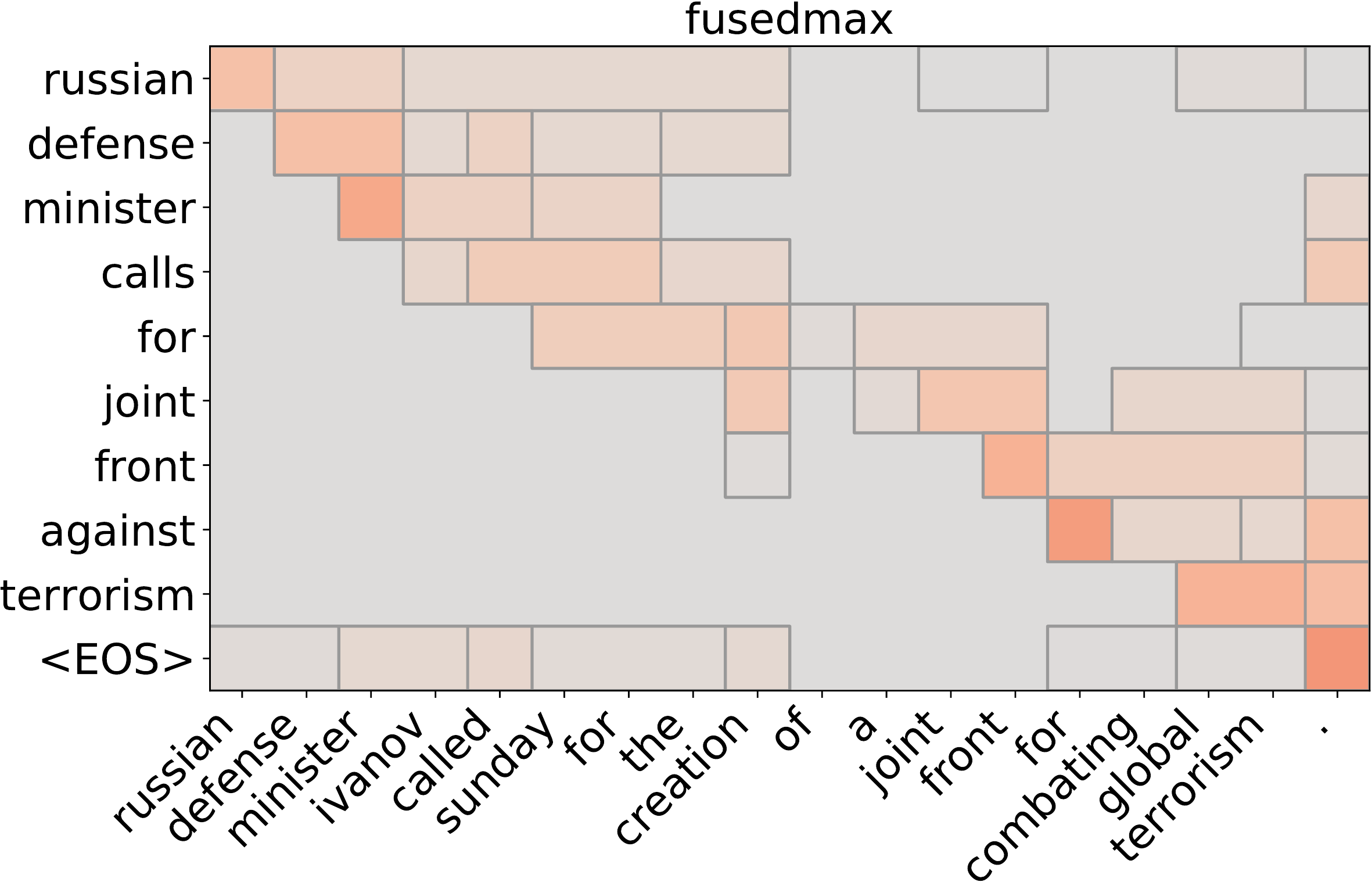}\\ 
    \includegraphics[width=0.43\textwidth]{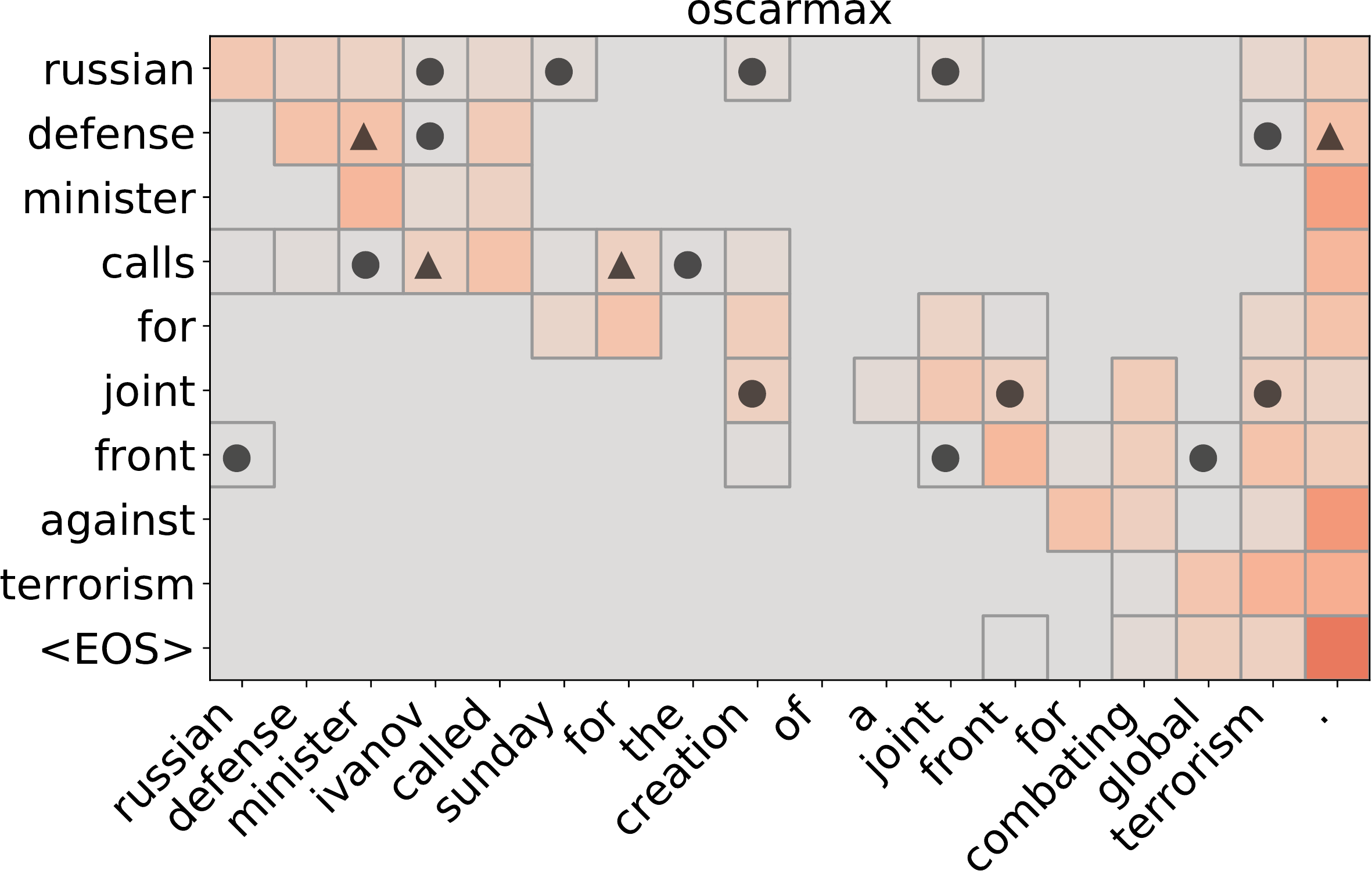} 
    \includegraphics[width=0.43\textwidth]{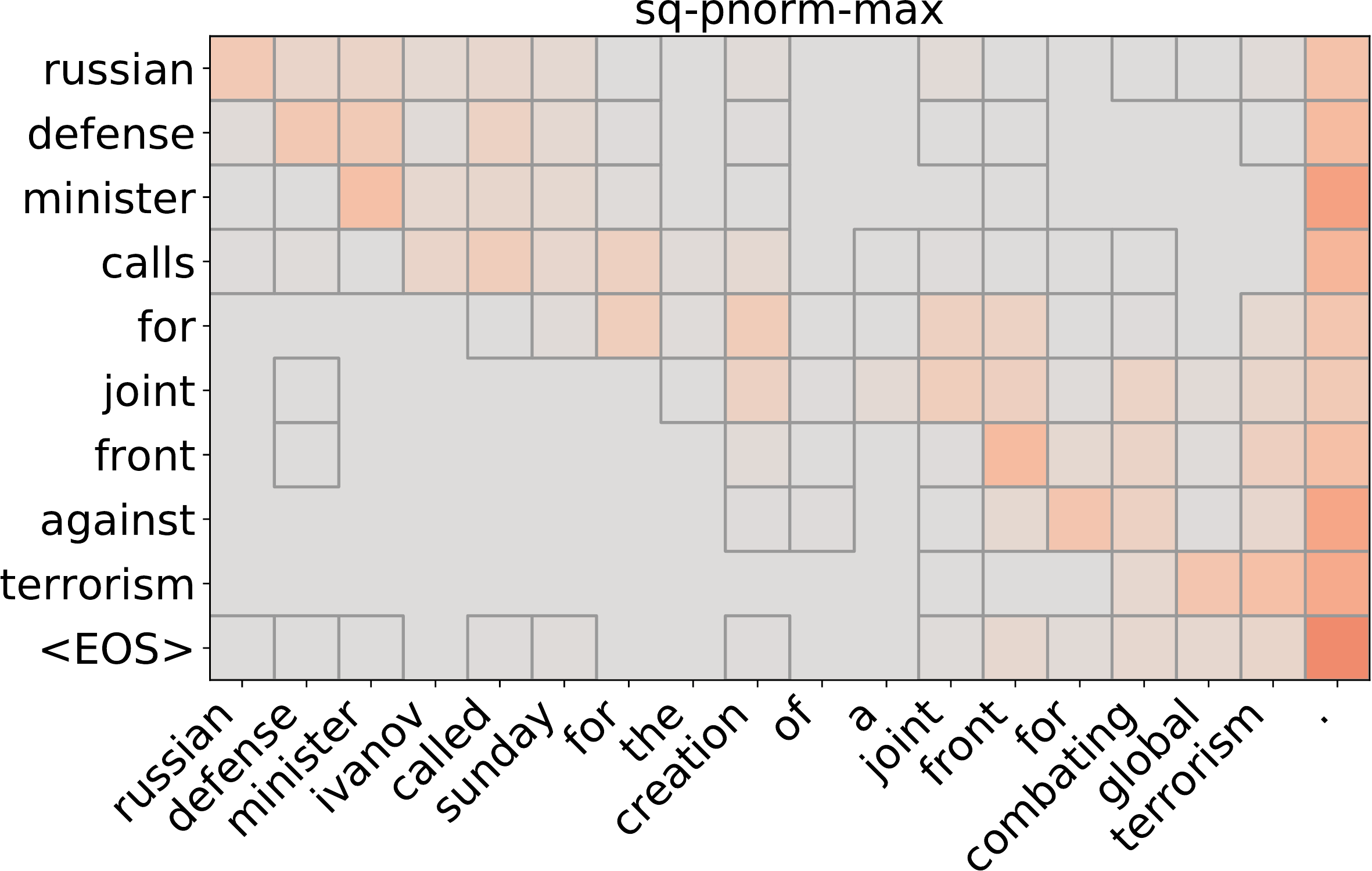}  
    \includegraphics[width=0.43\textwidth]{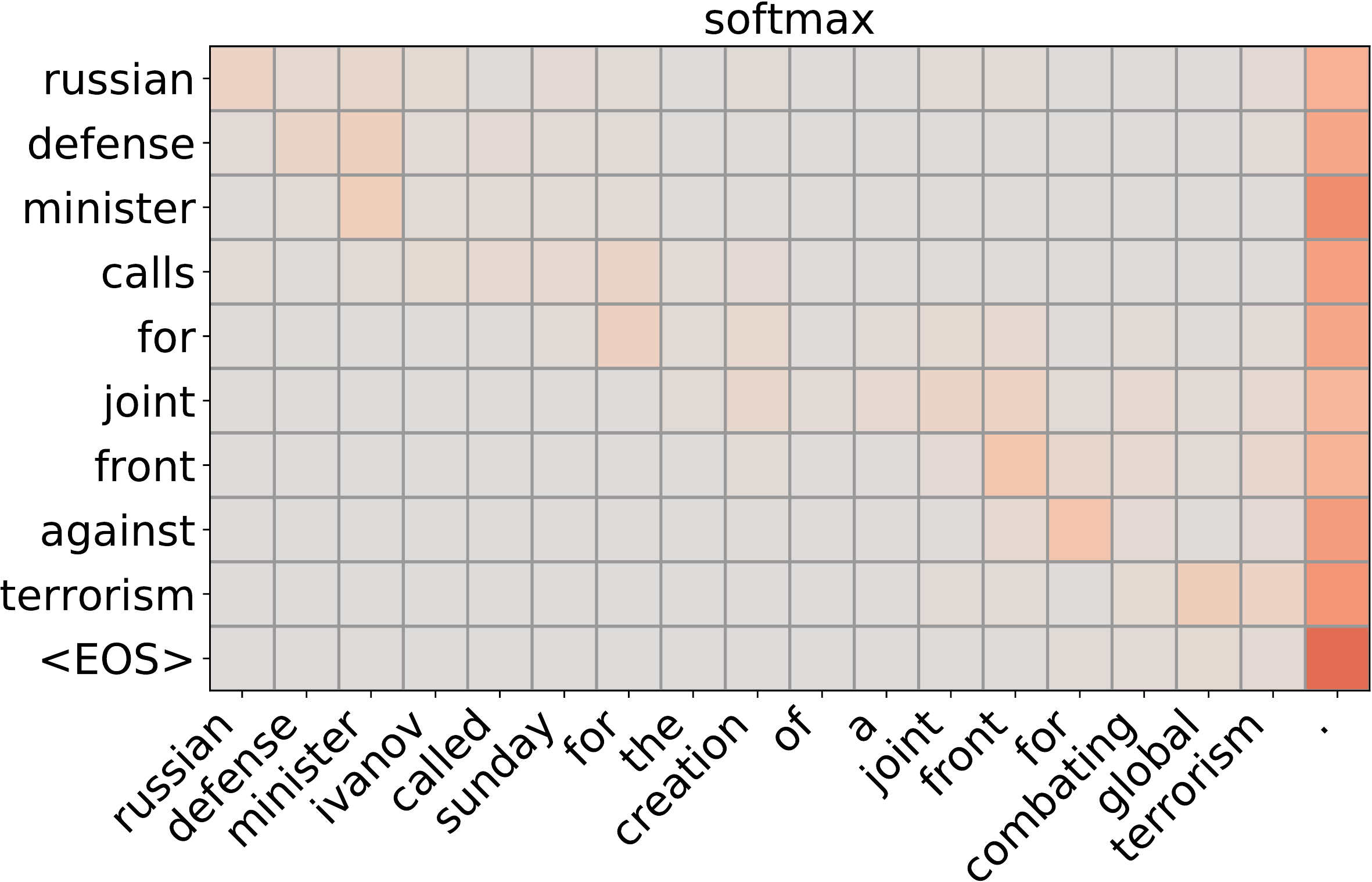} 
    \includegraphics[width=0.43\textwidth]{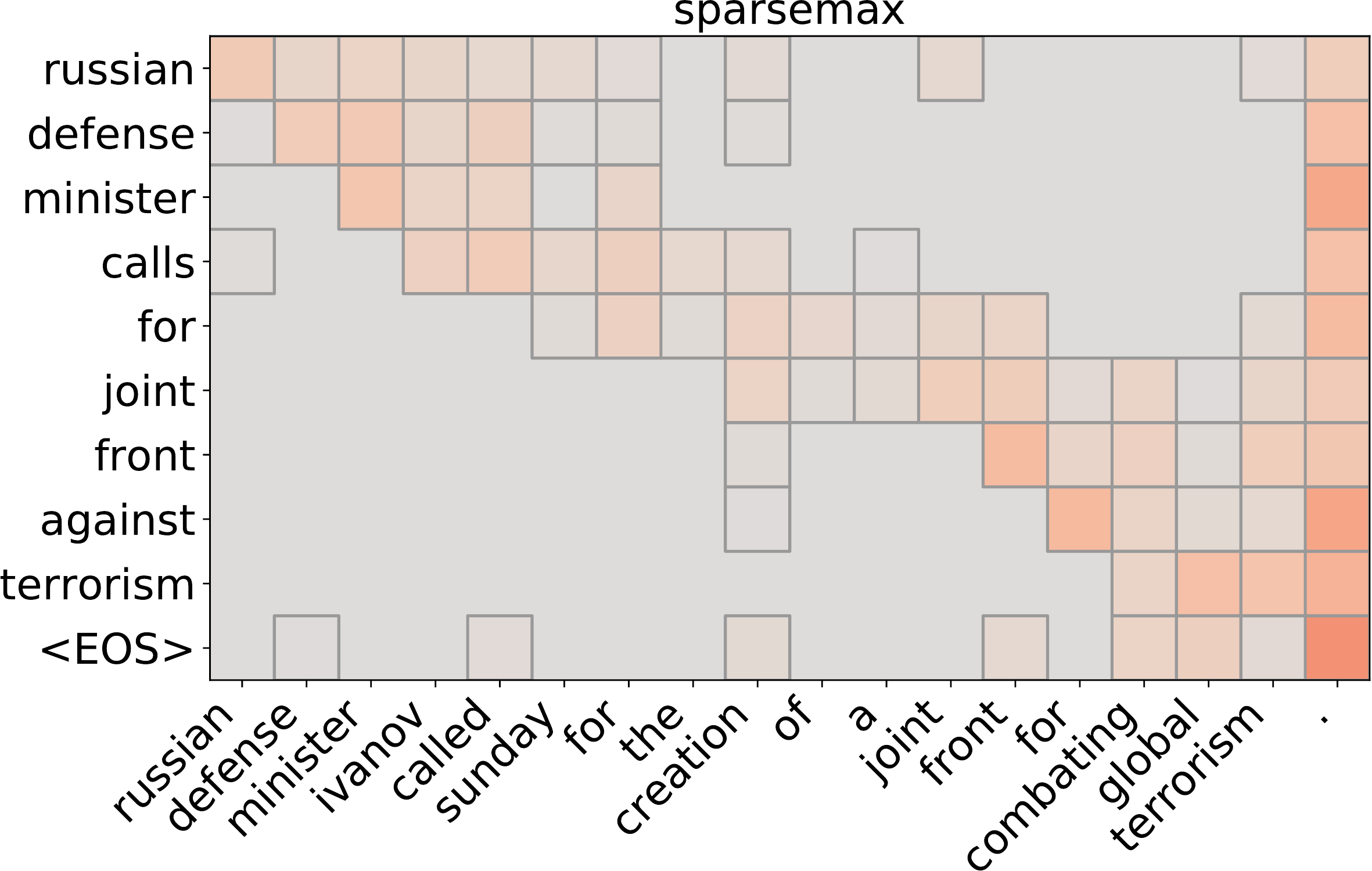}
    \\[0.66cm]
    \includegraphics[width=0.43\textwidth]{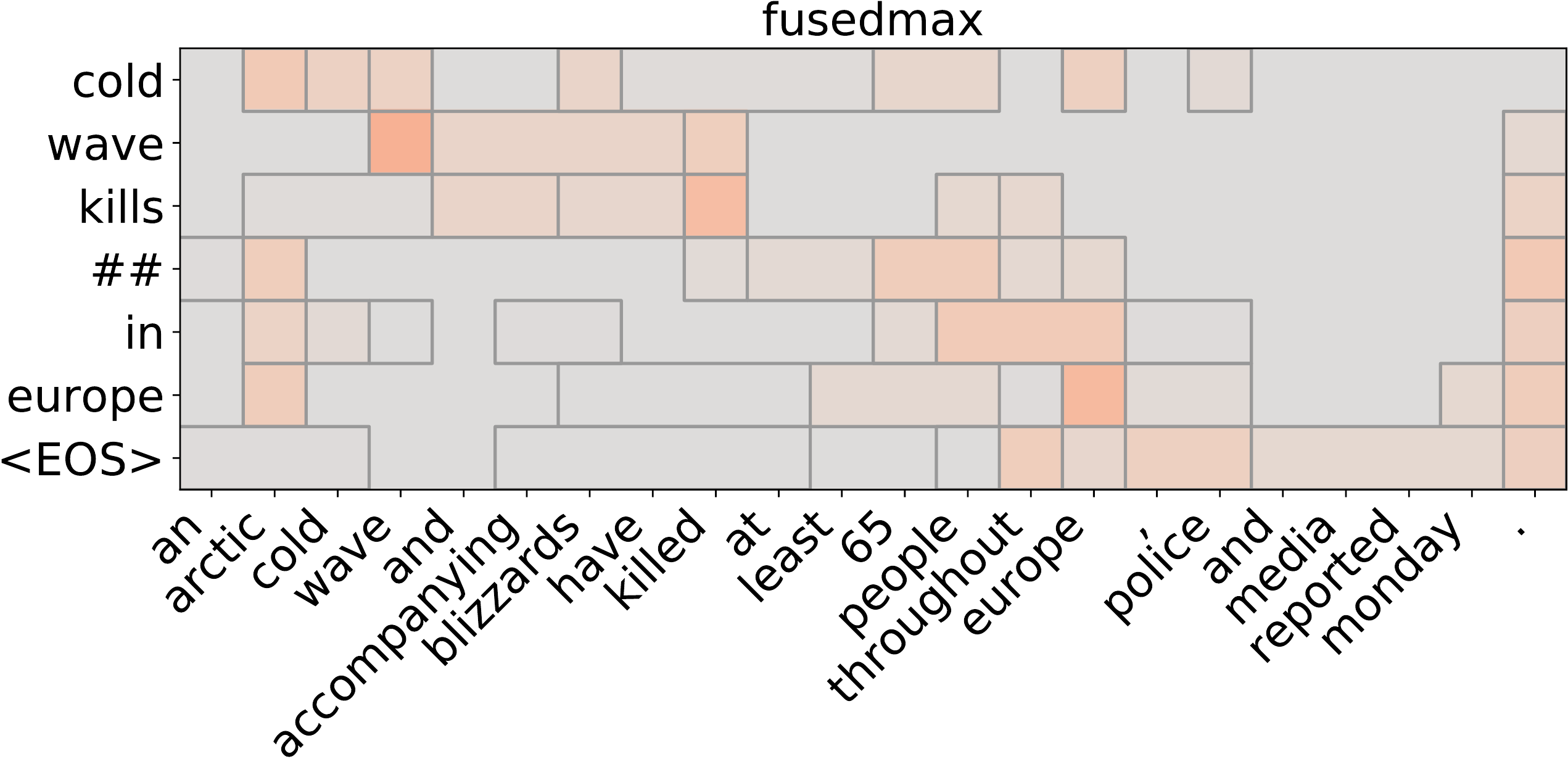}\\
    \includegraphics[width=0.43\textwidth]{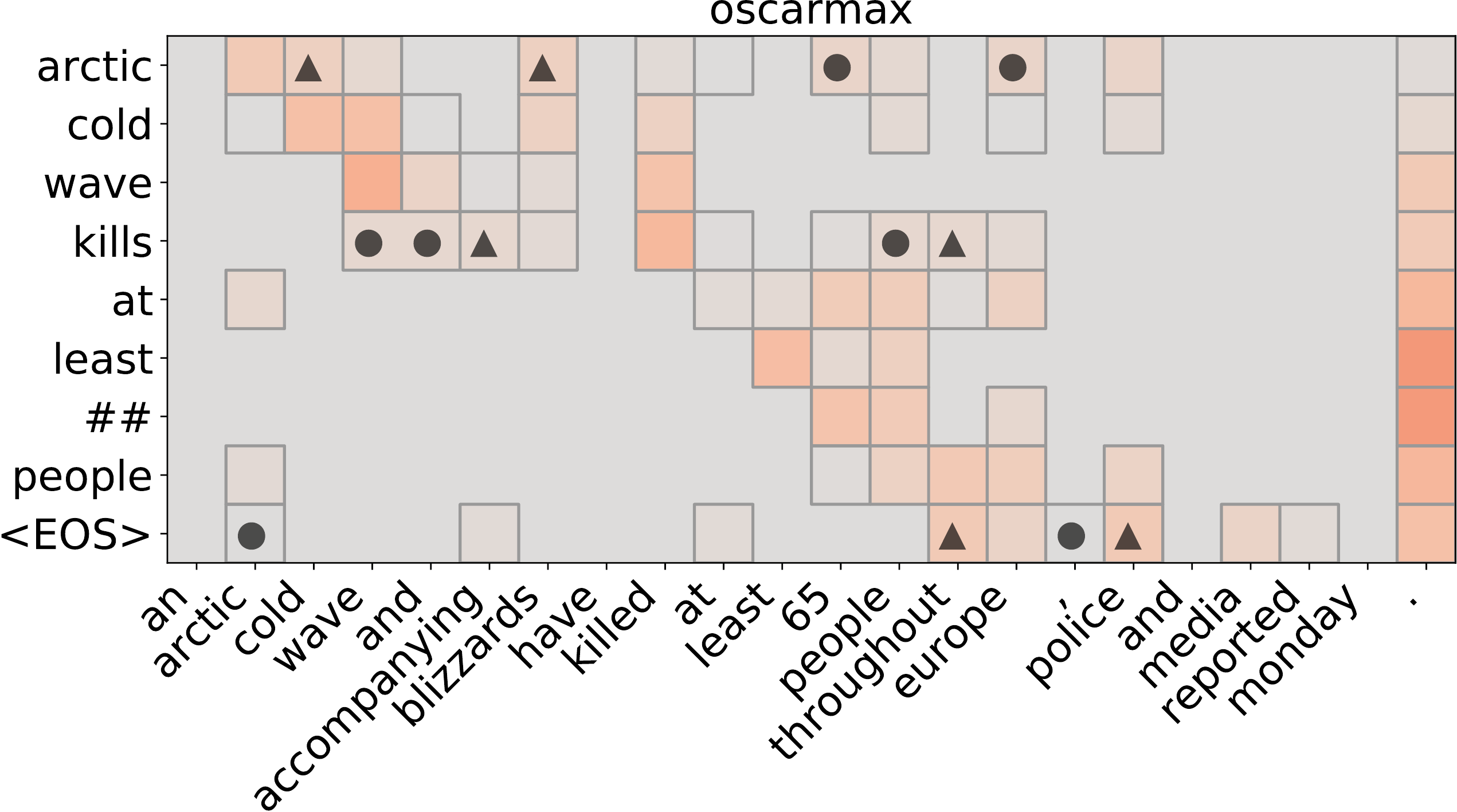}  
    \includegraphics[width=0.43\textwidth]{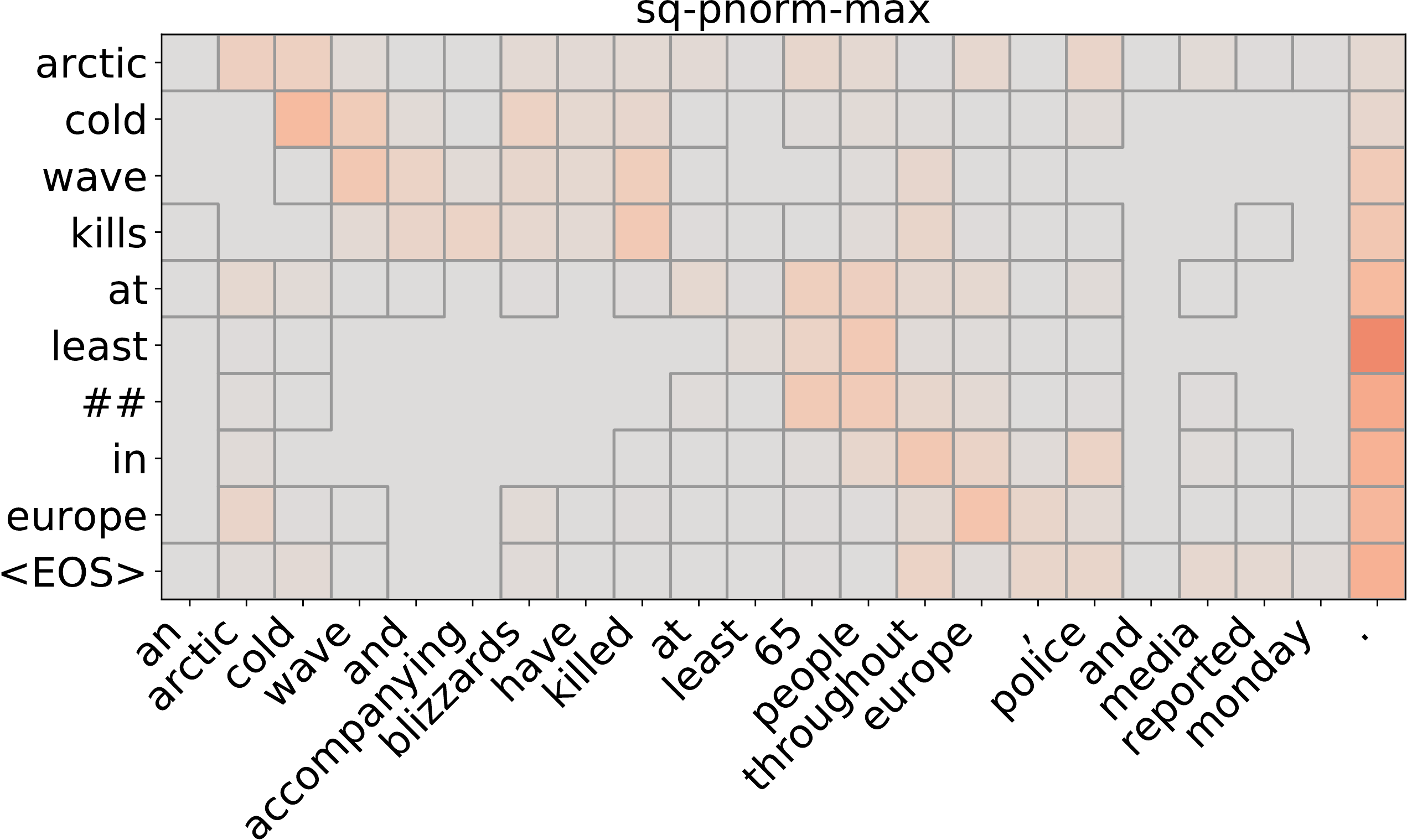}  
    \includegraphics[width=0.43\textwidth]{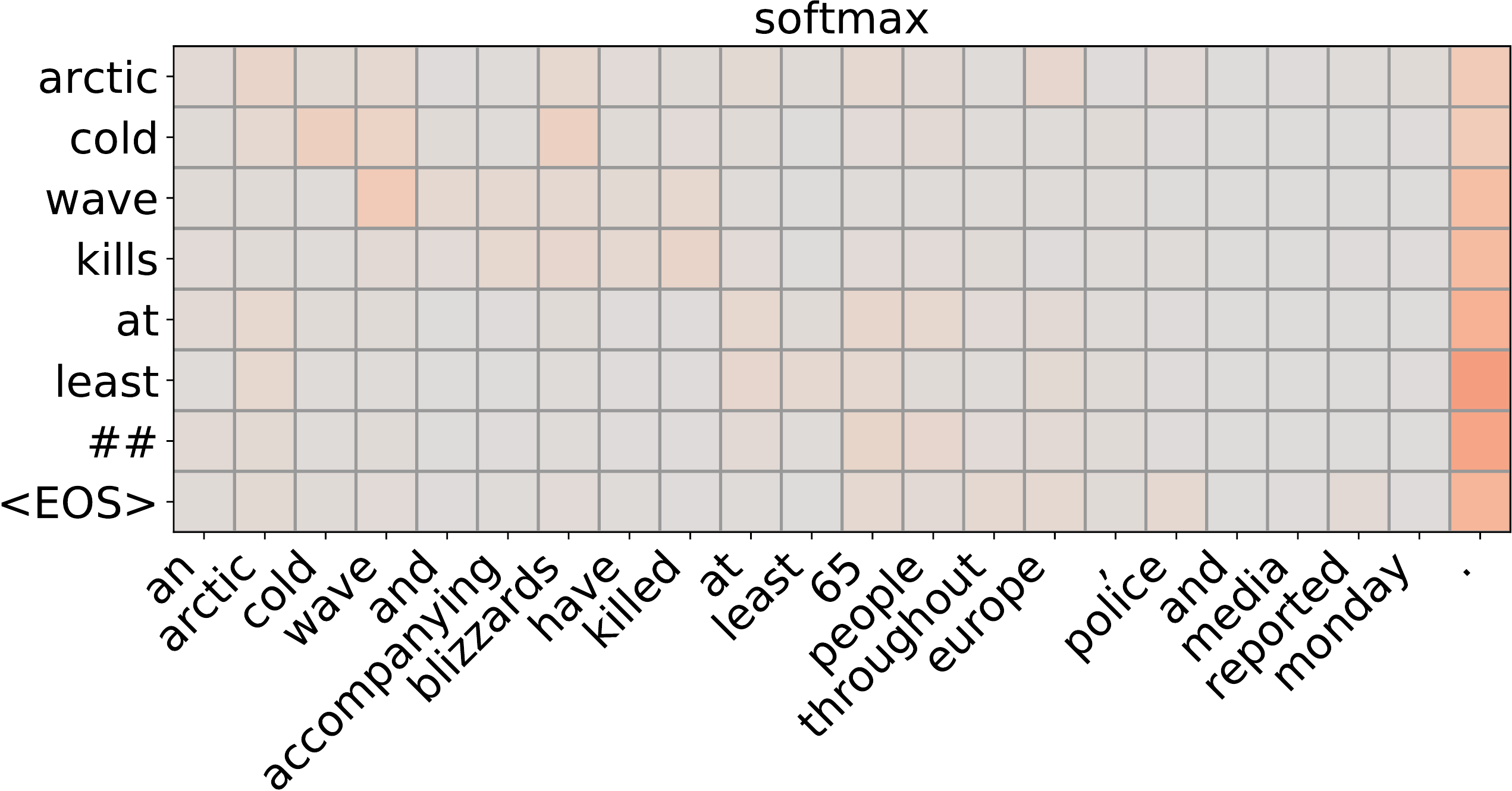} 
    \includegraphics[width=0.43\textwidth]{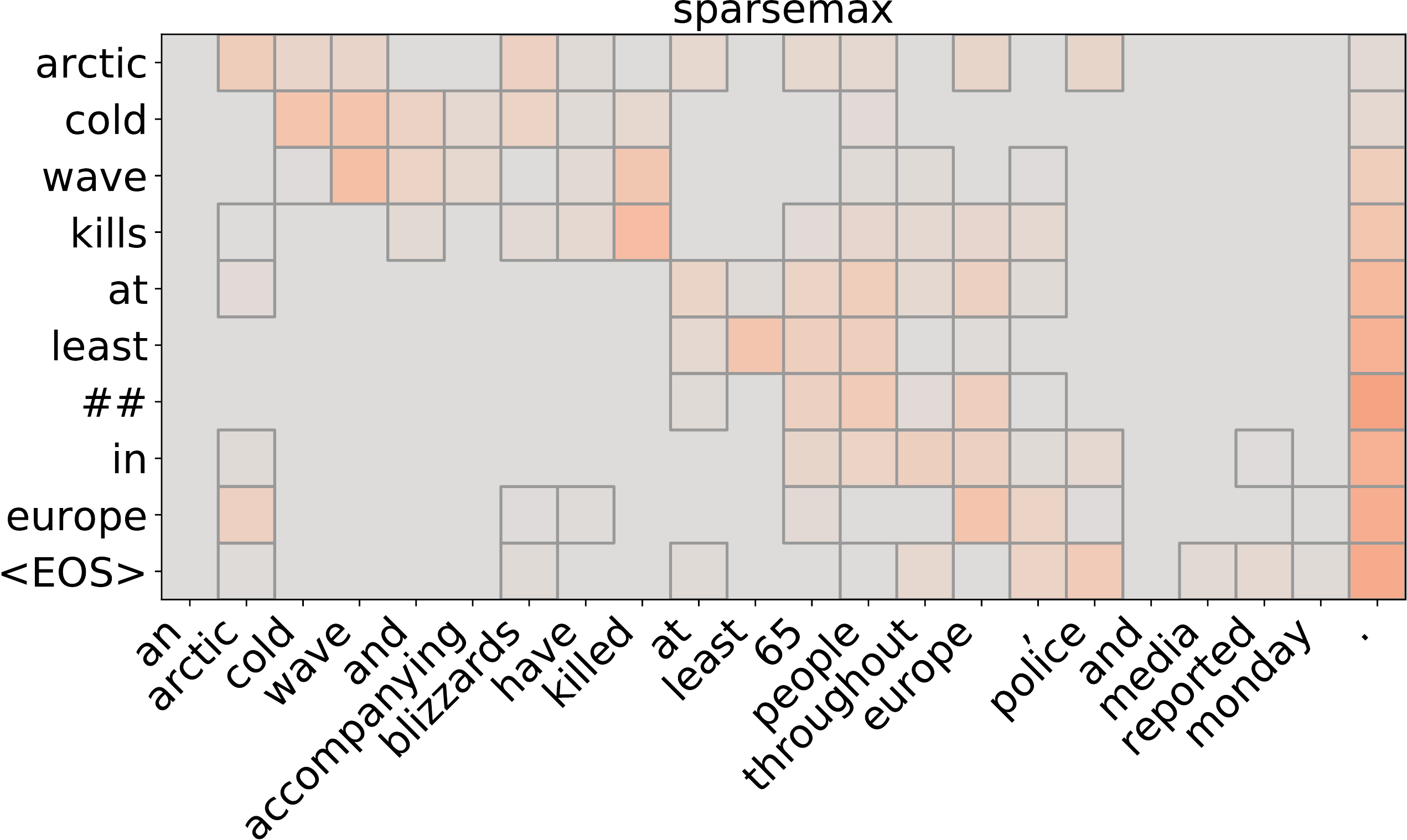}
    \caption{Summarization attention examples. 
    The 1-d TV prior of fusedmax
    captures well the intuition of aligning long input spans with single expressive
    words, as supported by ROUGE scores.\label{fig:supp_summ}}
\end{figure}
\begin{figure}[p]
    \ContinuedFloat \captionsetup{list=off,format=cont} \centering
    \includegraphics[width=0.9\textwidth]{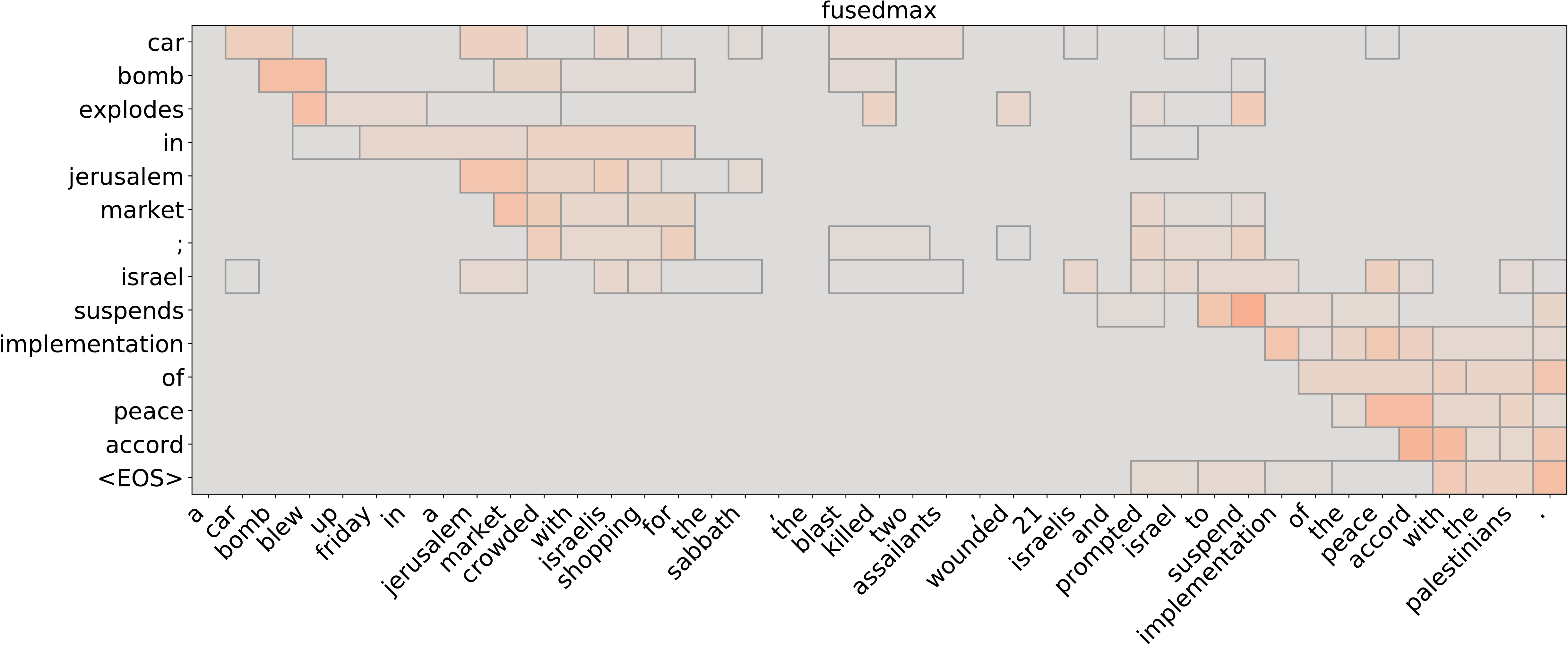}\\
    \includegraphics[width=0.9\textwidth]{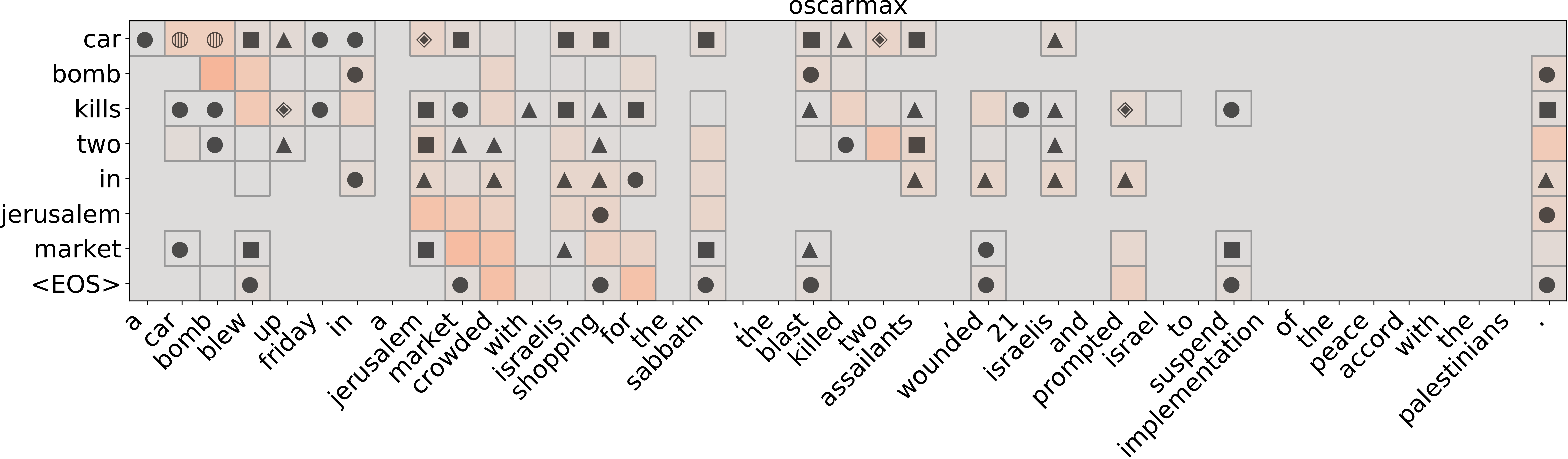}
    \includegraphics[width=0.9\textwidth]{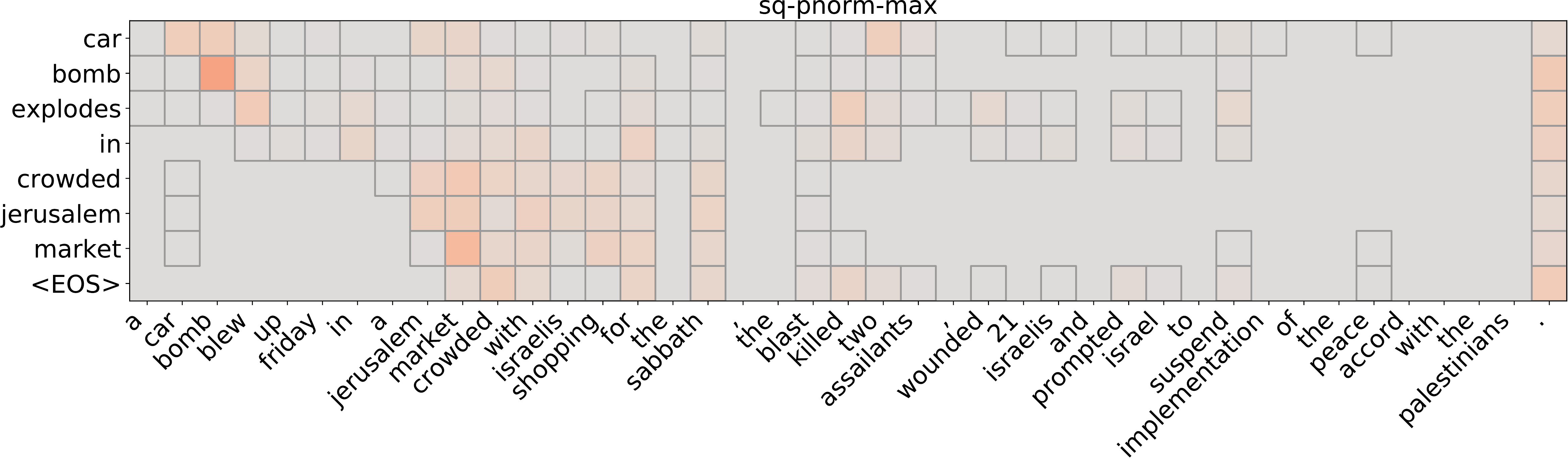}
    \includegraphics[width=0.9\textwidth]{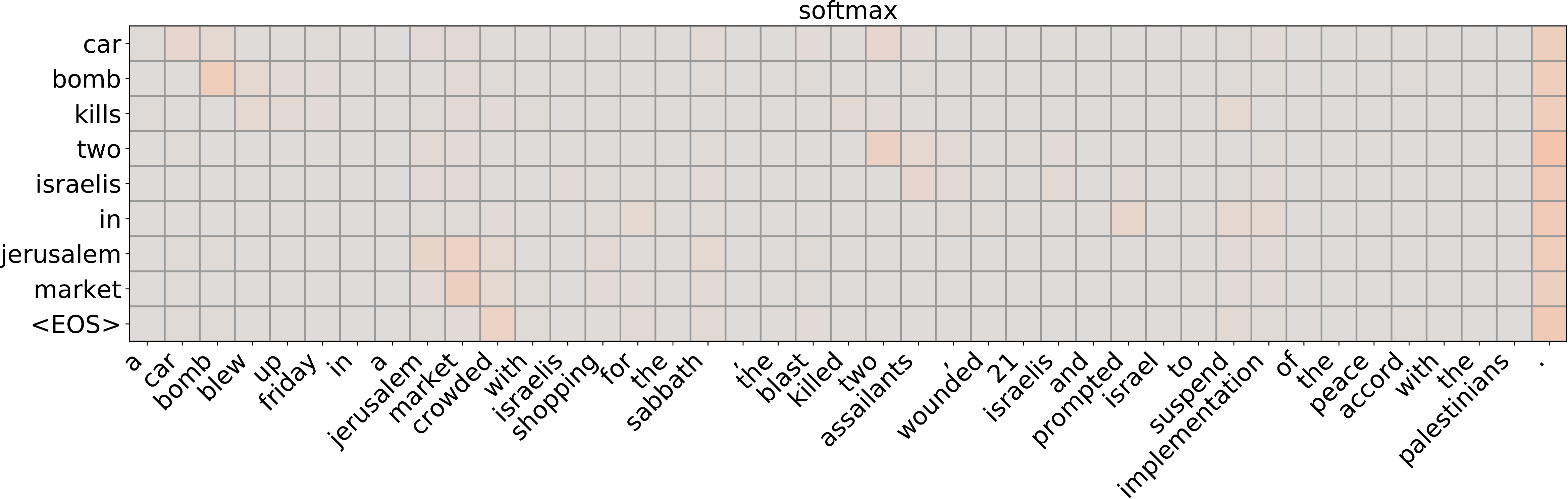}
    \includegraphics[width=0.9\textwidth]{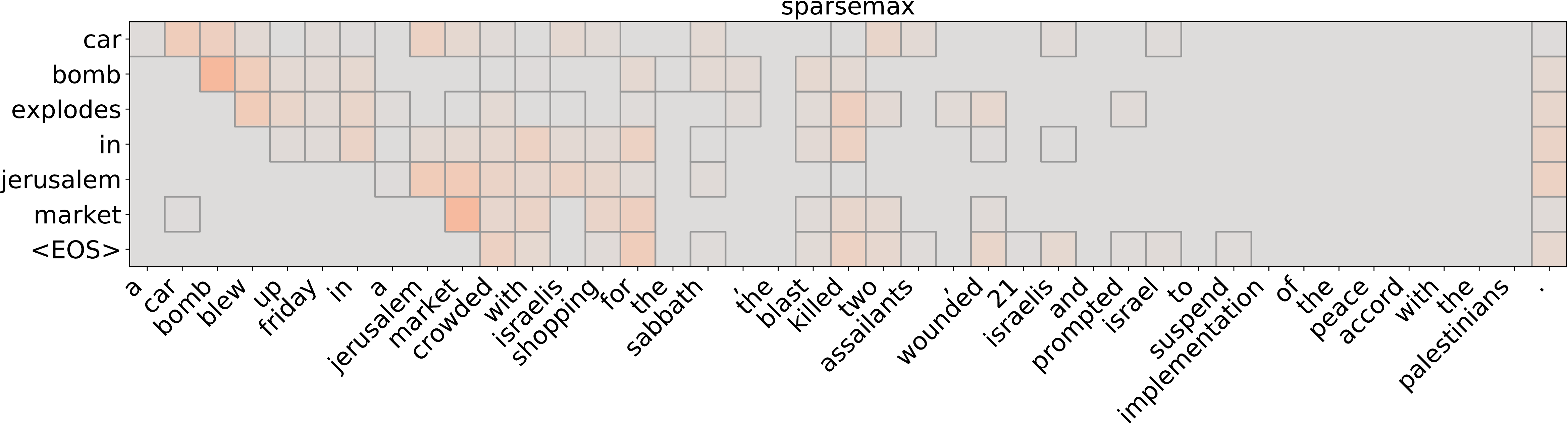}
    \caption{Summarization attention examples. Here, fusedmax recovers a longer
    but arguably better summary, identifying a separate but important part of
    the input sentence.} 
\end{figure}
\begin{figure}[p]
    \ContinuedFloat
    \captionsetup{list=off,format=cont}
    \centering
    \includegraphics[width=0.9\textwidth]{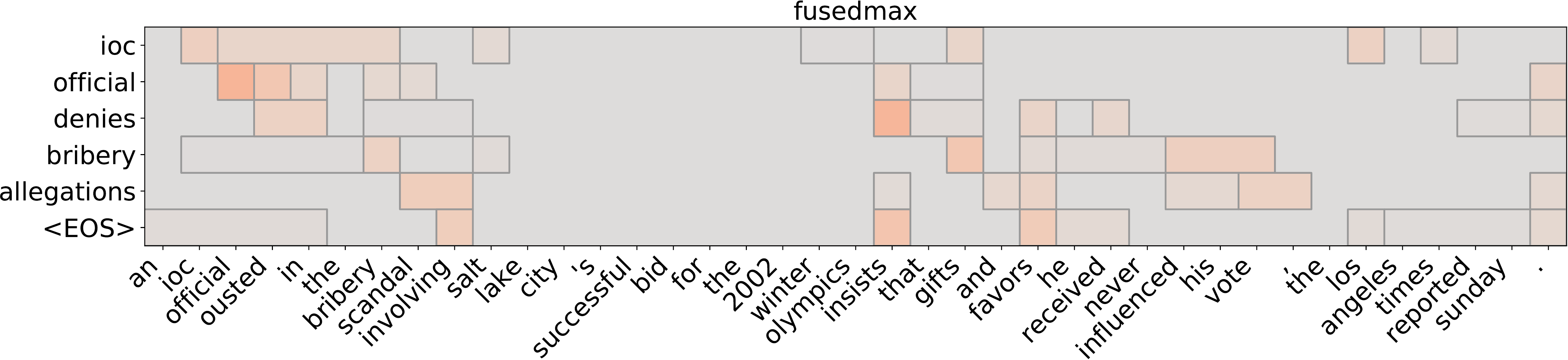}\\
    \includegraphics[width=0.9\textwidth]{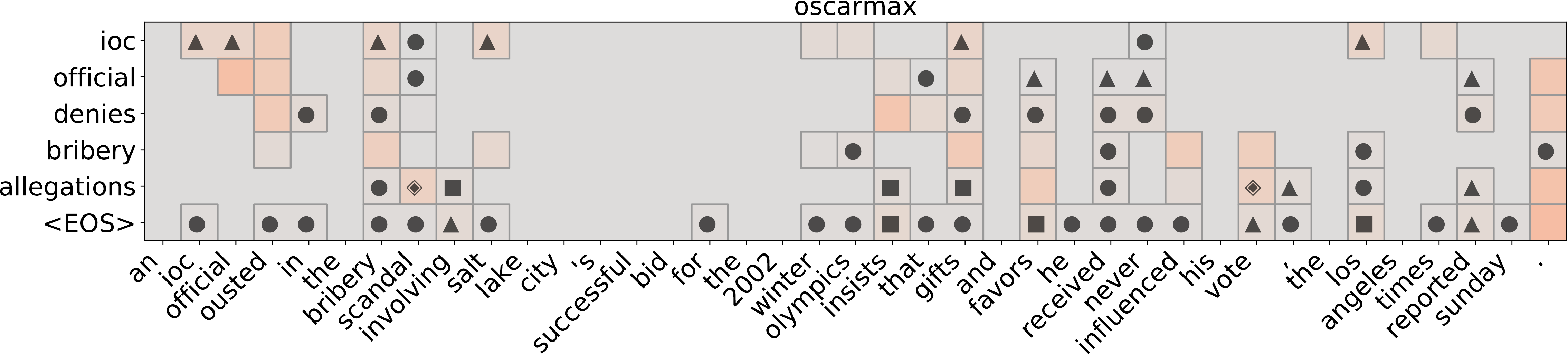}  
    \includegraphics[width=0.9\textwidth]{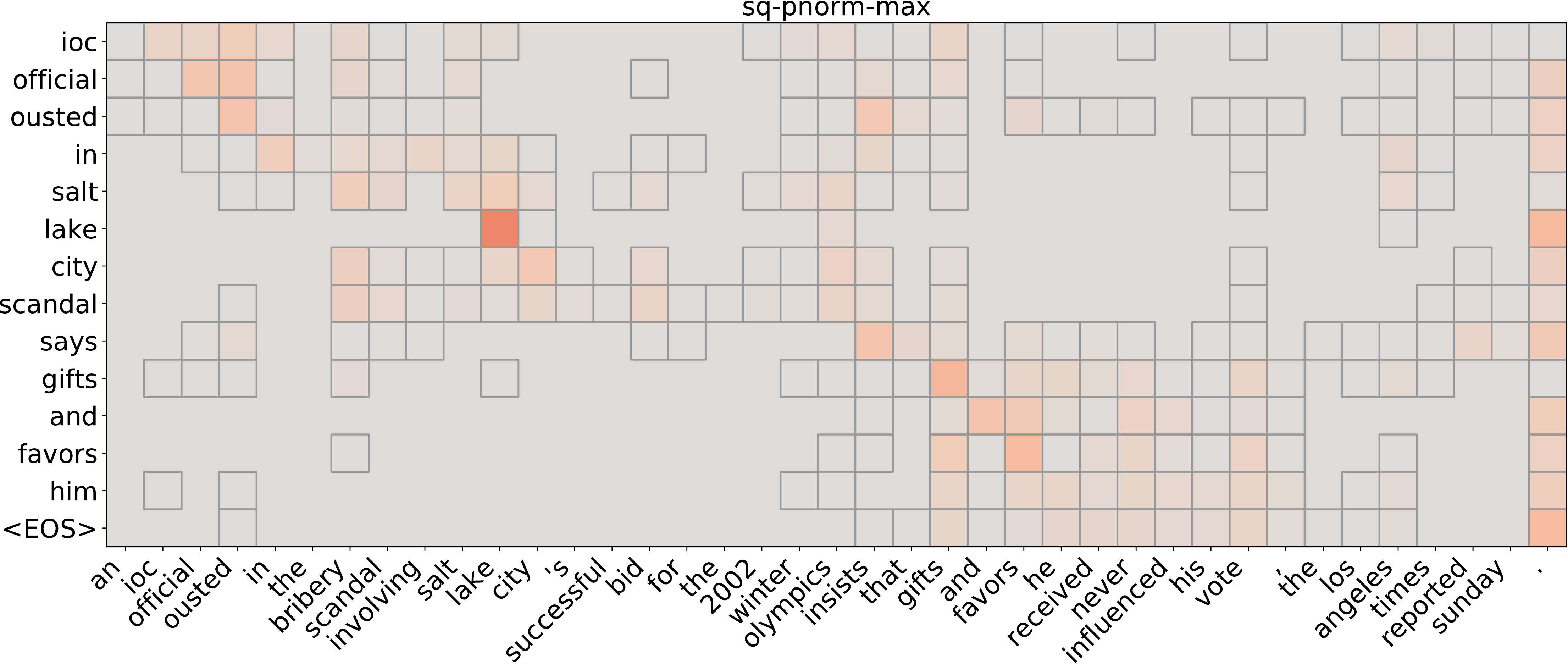}  
    \includegraphics[width=0.9\textwidth]{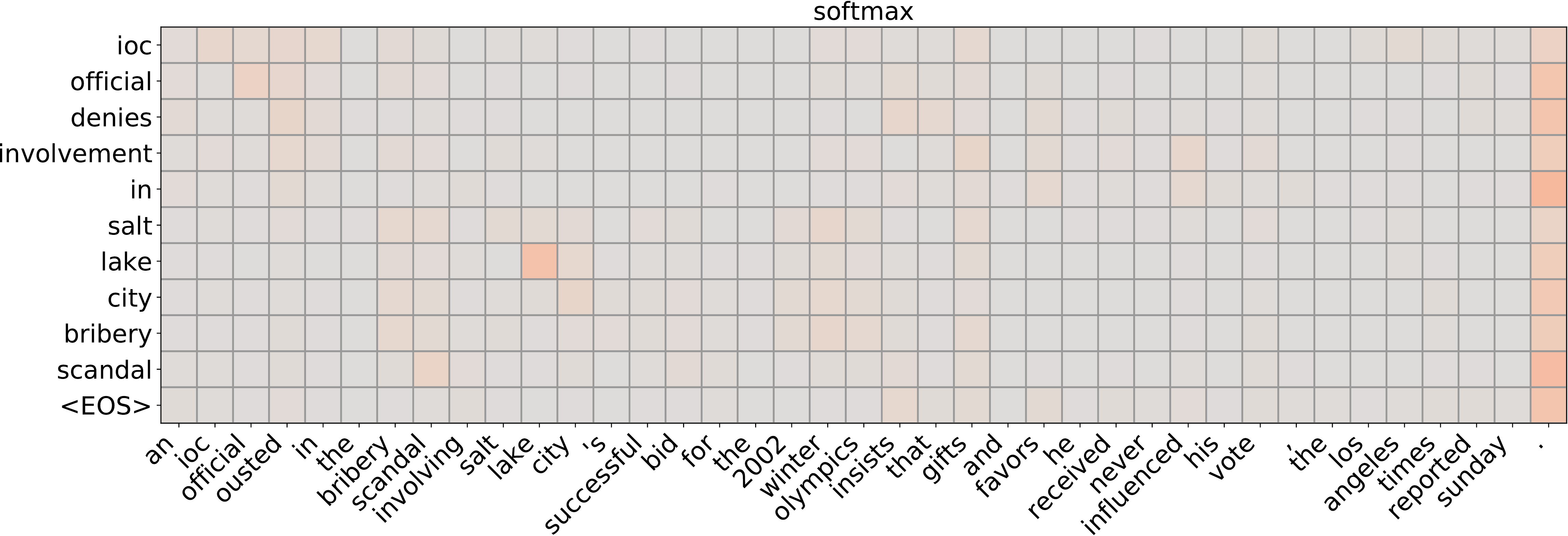} 
    \includegraphics[width=0.9\textwidth]{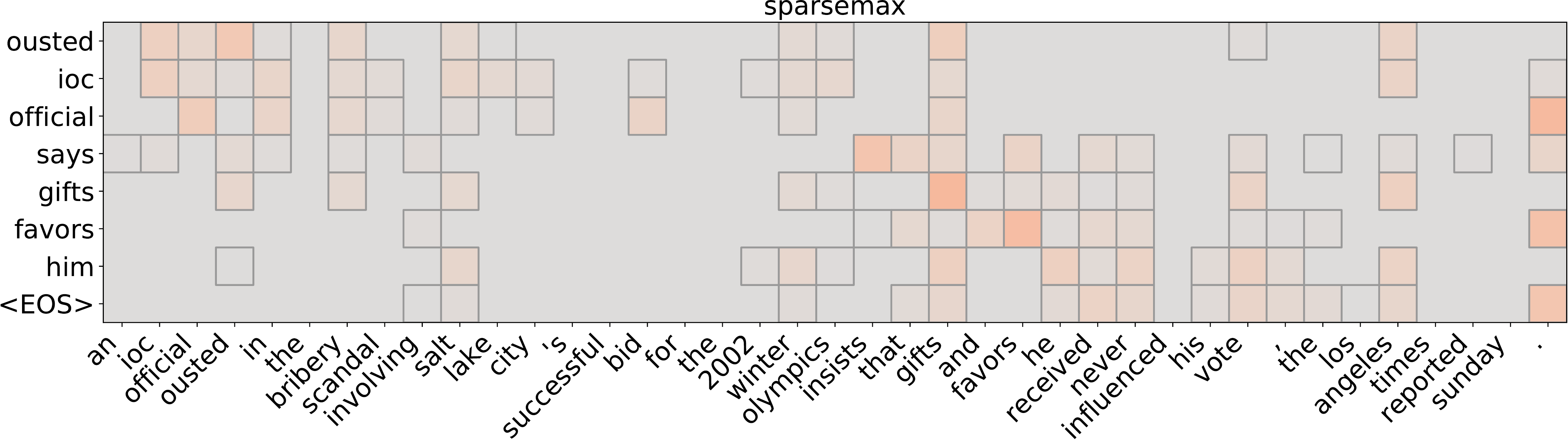}
    \caption{Summarization attention examples. Here, fusedmax and oscarmax
    produce a considerably shorter summary.}
\end{figure}
\end{document}